\documentclass[10pt,twocolumn,letterpaper]{article}

\usepackage{cvpr}              

\usepackage{arydshln}
\usepackage{graphicx}
\usepackage{amsmath}
\usepackage{amssymb}
\usepackage{booktabs}
\usepackage{multirow}
\usepackage{multicol}
\usepackage{enumitem}
\usepackage{color, colortbl}
\usepackage[dvipsnames,x11names]{xcolor}
\usepackage{etoc}
\usepackage{float}
\makeatletter
\@namedef{ver@everyshi.sty}{}
\makeatother
\usepackage{tikz}

\usepackage[pagebackref,breaklinks,colorlinks]{hyperref}

\usepackage[capitalize]{cleveref}
\crefname{section}{Sec.}{Secs.}
\Crefname{section}{Section}{Sections}
\Crefname{table}{Table}{Tables}
\crefname{table}{Tab.}{Tabs.}

\def\cvprPaperID{969} 
\def\confName{CVPR}
\def\confYear{2023}

\newcommand{\reward}{r}
\newcommand{\discountRate}{\gamma}

\newcommand{\mdp}{\mathcal{M}}

\newcommand{\supmat}{supp.~material\xspace}

\newcommand{\gc}{\cellcolor[HTML]{E4E4E4}}

\newcommand{\update}[1]{{\color{black}{#1}}}

\newcommand{\Fig}[1]{Fig.~\ref{fig:#1}}

\newcommand{\Tab}[1]{Tab.~\ref{tab:#1}}

\newcommand{\Eq}[1]{Eq.~\ref{eq:#1}}

\newcommand{\Sec}[1]{Sec.~\ref{sec:#1}}

\DeclareMathOperator*{\argmax}{arg\,max}

\definecolor{Gray}{gray}{0.9}

\newif\ifshowcomments

\makeatletter
\def\and{
  \end{tabular}%
  \hskip 2.85em \@plus.17fil%
  \begin{tabular}[t]{c}}
\makeatother

\ifshowcomments
        \newcommand{\note}[3]{{\textcolor{#2}{[#1: #3]}}}
        \newcommand{\OH}[1]{\note{Otmar}{green}{#1}}
		\newcommand{\SC}[1]{\note{Sammy}{teal}{#1}}
        \newcommand{\wei}[1]{{\color{blue} [WEI: #1] }}
        \newcommand{\ywchao}[1]{{\color{red} [Yu-Wei: #1] }}
        \newcommand{\claudia}[1]{{\color{magenta} [Claudia: #1] }}
        \newcommand{\dieter}[1]{{\color{olive} [Dieter: #1] }}
        \newcommand{\TL}[1]{{\color{cyan} [TL: #1] }}

        \usepackage{color}

\else
        \newcommand{\note}[3]{\unskip}
        \newcommand{\OH}[1]{\unskip}
		\newcommand{\SC}[1]{\unskip}
		\newcommand{\ywchao}[1]{\unskip}
		\newcommand{\wei}[1]{\unskip}
		\newcommand{\claudia}[1]{\unskip}
		\newcommand{\dieter}[1]{\unskip}
        \newcommand{\TL}[1]{{\unskip}}
\fi

\newcommand{\goalvec}{\mathbf{g}}

\newcommand{\replay}{\mathbf{\mathcal{D}}}

\newcommand{\actions}{\mathbf{a}}
\newcommand{\expertvec}{\mathbf{e}}

\newcommand{\pcvec}{\mathbf{p}}
\newcommand{\rewardvec}{\mathbf{r}}
\newcommand{\statevec}{\mathbf{s}}
\newcommand{\transitionvec}{\mathbf{d}}

\newcommand{\policy}{\boldsymbol{\pi}}
\newcommand{\grasppred}{\boldsymbol{\sigma}}
\newcommand{\policypre}{\boldsymbol{\pi_{\text{pre}}}}
\newcommand{\criticpre}{\boldsymbol{Q_{\text{pre}}}}
\newcommand{\policyft}{\boldsymbol{\pi_*}}
\newcommand{\criticft}{\boldsymbol{Q_*}}
\newcommand{\policyexp}{\boldsymbol{\pi_{\text{exp}}}}

\usepackage{etoolbox}

\renewcommand{\thefootnote}{\fnsymbol{footnote}}

\newcolumntype{C}[1]{>{\centering\let\newline\\\arraybackslash\hspace{0pt}}m{#1}}
\newcolumntype{L}[1]{>{\raggedright\let\newline\\\arraybackslash\hspace{0pt}}m{#1}}
\newcolumntype{R}[1]{>{\raggedleft\let\newline\\\arraybackslash\hspace{0pt}}m{#1}}

\newcommand{\greencheck}{{\color{Green4} \checkmark}}
\newcommand{\redcross}{{\color{red}$\times$}}

\begin{document}


\title{Learning Human-to-Robot Handovers from Point Clouds}

\author{%
  Sammy Christen$^{1,2*}$ \and Wei Yang$^{2}$ \and Claudia Pérez-D'Arpino$^2$ \and Otmar Hilliges$^1$ \and Dieter Fox$^{2,3}$ \and Yu-Wei Chao$^2$
  \vspace{1mm}
  \and
  $^1$ETH Zurich \quad $^2$NVIDIA \quad $^3$University of Washington \vspace{-1mm} \\ \vspace{-3mm}
  \makebox[0cm]{\tt\footnotesize \{sammy.christen, otmar.hilliges\}@inf.ethz.ch \{weiy, claudiap, dieterf, ychao\}@nvidia.com} \vspace{-2mm}
  }

\twocolumn[{%
\renewcommand\twocolumn[1][]{#1}%
\maketitle
\begin{center}
    \centering
    \captionsetup{type=figure}
    \includegraphics[width=0.94\textwidth]{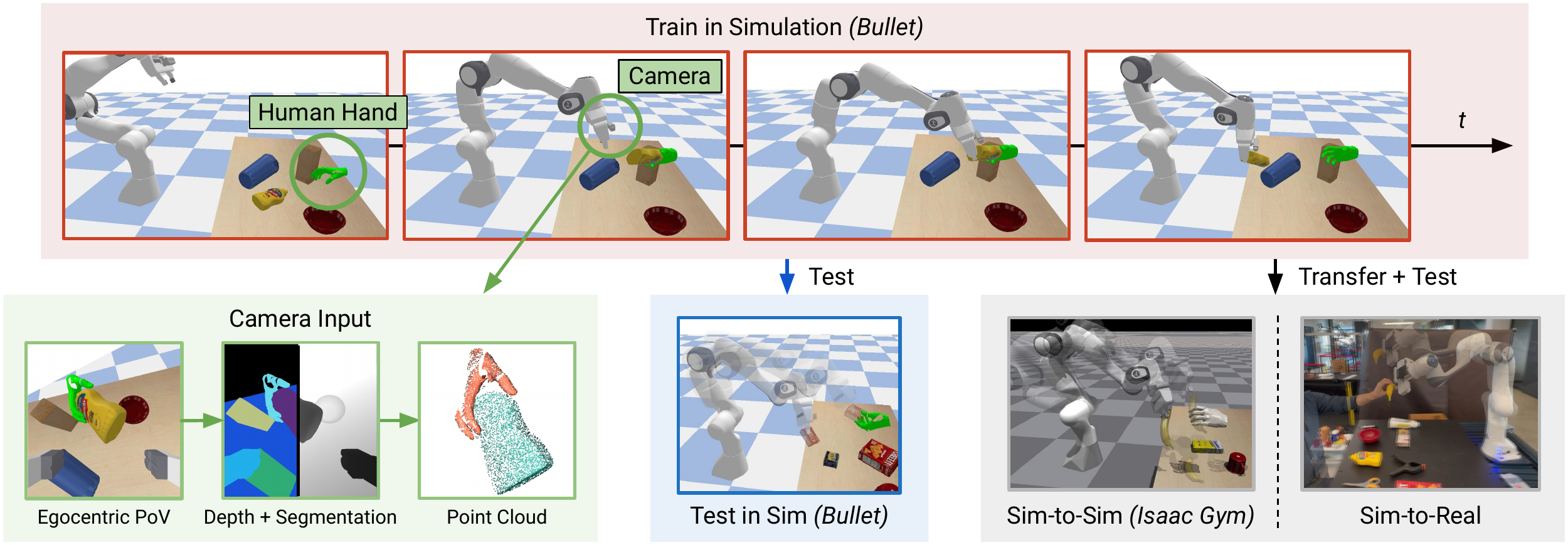}
    \captionof{figure}{We introduce a framework to learn human-to-robot handover policies from point cloud input. Our policies take input from a wrist mounted camera and directly generate action output for the robot's end effector. We train our policies in a simulated handover environment, and evaluate on unseen handover motion and poses. We further transfer the model across physics simulators and to a real robotic platform.}
    \label{fig:teaser}
\end{center}%
}]

\etocdepthtag.toc{mtchapter}
\begin{abstract}
\vspace{-2mm}
We propose the first framework to learn control policies for vision-based human-to-robot handovers, a critical task for human-robot interaction. While research in Embodied AI has made significant progress in training robot agents in simulated environments, interacting with humans remains challenging due to the difficulties of simulating humans. Fortunately, recent research has developed realistic simulated environments for human-to-robot handovers. Leveraging this result, we introduce a method that is trained with a human-in-the-loop via a two-stage teacher-student framework that uses motion and grasp planning, reinforcement learning, and self-supervision. We show significant performance gains over baselines on a simulation benchmark, sim-to-sim transfer and sim-to-real transfer. Video and code are available at \footnotesize{\url{https://handover-sim2real.github.io}}.
\end{abstract}
\vspace{-5mm}

\section{Introduction}

\def\thefootnote{*}\footnotetext{This work was done during an internship at NVIDIA.}

Handing over objects between humans and robots is an important tasks for human-robot interaction (HRI)~\cite{ortenzi:tro2021}. It allows robots to assist humans in daily collaborative activities, such as helping to prepare a meal, or to exchange tools and parts with human collaborators in manufacturing settings. To complete these tasks successfully and safely, intricate coordination between human and robot is required. This is challenging, because the robot has to react to human behavior, while only having access to sparse sensory inputs such as a single camera with limited field of view. Therefore, a need for methods that solve interactive tasks such as handovers purely from vision input arises.

Bootstrapping robot training in the real world can be unsafe and time-consuming. Therefore, recent trends in Embodied AI have focused on training agents to act and interact in simulated (sim) environments~\cite{deitke:cvpr2020,xiang:cvpr2020,shridhar:cvpr2020,srivastava:corl2021,szot:neurips2021,gan:neuripstdb2021,deitke2022retrospectives}. With advances in rendering and physics simulation, models have been trained to map raw sensory input to action output, and can even be directly transferred from simulation to the real world~\cite{anderson:corl2020,shen:iros2021}. Many successes have been achieved particularly around the suite of tasks of robot navigation, manipulation, or a combination of both. In contrast to these areas, little progress has been made around tasks pertained to HRI. This is largely hindered by the challenges in embedding realistic human agents in these environments, since modeling and simulating realistic humans is challenging. 

Despite the challenges, an increasing number of works have attempted to embed realistic human agents in simulated environments~\cite{christen:handshake2019,erickson:icra2020,puig:iclr2021,perezdarpino:icra2021,pang:roman2021,wang:corl2021a,chao:icra2022}. Notably, a recent work has introduced a simulation environment (``HandoverSim'') for human-to-robot handover (H2R)~\cite{chao:icra2022}. To ensure a realistic human handover motion, they use a large motion capture dataset~\cite{chao:cvpr2021} to drive the movements of a virtual human in simulation. However, despite the great potential for training robots, the work of~\cite{chao:icra2022} only evaluates off-the-shelf models from prior work, and has not explored any policy training with humans in the loop in their environment.


We aim to close this gap by introducing a vision-based learning framework for H2R handovers that is trained with a human-in-the-loop (see \Fig{teaser}). In particular, we propose a novel mixed imitation learning (IL) and reinforcement learning (RL) based approach, trained by interacting with the humans in HandoverSim. Our approach draws inspiration from a recent method for learning polices for grasping static objects from point clouds~\cite{wang:corl2021b}, but proposes several key changes to address the challenges in H2R handovers. In contrast to static object grasping, where the policy only requires object information, we additionally encode human hand information in the policy's input. Also, compared to static grasping without a human, we explicitly take human collisions into account in the supervision of training. Finally, the key distinction between static object grasping and handovers is the dynamic nature of the hand and object during handover. To excel on the task, the robot needs to react to dynamic human behavior. Prior work typically relies on open-loop motion planners~\cite{wang:rss2020} to generate expert demonstrations, which may result in suboptimal supervision for dynamic cases. To this end, we propose a two-stage training framework. In the first stage, we fix the humans to be stationary and train an RL policy that is partially guided by expert demonstrations obtained from a motion and grasp planner. In the second stage, we finetune the RL policy in the original dynamic setting where the human and robot move simultaneously. Instead of relying on a planner, we propose a self-supervision scheme, where the pre-trained RL policy serves as a teacher to the downstream policy.
We evaluate our method in three ``worlds'' (see \Fig{teaser}). First, we evaluate on the ``native'' test scenes in HandoverSim~\cite{chao:icra2022}, which use the same backend physics simulator (Bullet~\cite{coumans:2021}) as training but unseen handover motions from the simulated humans. Next, we perform sim-to-sim evaluation on the test scenes implemented with a different physics simulator (Isaac Gym~\cite{makoviychuk:neuripstdb2021}). Lastly, we investigate sim-to-real transfer by evaluating polices on a real robotic system and demonstrate the benefits of our method. 

We contribute: i) the first framework to train human-to-robot handover tasks from vision input with a human-in-the-loop, ii) a novel teacher-student method to train in the setting of a jointly moving human and robot, iii) an empirical evaluation showing that our approach outperforms baselines on the HandoverSim benchmark, iv) transfer experiments indicating that our method leads to more robust sim-to-sim and sim-to-real transfer compared to baselines.

\section{Related Work}

\paragraph{Human-to-Robot Handovers}

%
Encouraging progress in hand and object pose estimation~\cite{hasson:cvpr2019,liu:cvpr2021,li:cvpr2022} has been achieved, aided by the introduction of large hand-object interaction datasets~\cite{garcia-hernando:cvpr2018,hampali:cvpr2020,brahmbhatt:eccv2020,moon:eccv2020,taheri:eccv2020,chao:cvpr2021,ye:iccv2021,kwon:iccv2021,liu:cvpr2022, fan2023arctic}. These developments enable applying model-based grasp planning~\cite{bicchi:icra2000,miller:ram2004,bohg:tro2014}, a well-studied approach in which full pose estimation and tracking are needed, to H2R handovers~\cite{sanchez-matilla:ral2020,chao:cvpr2021}. However, these methods require the 3D shape models of the object and cannot handle unseen objects.
Alternatively, some recent works~\cite{rosenberger:ral2021,yang:icra2021,yang:icra2022,duan:tcds, marturi:2019dynamic} achieve H2R handover by employing learning-based grasp planners to generate grasps for novel objects from raw vision inputs such as images or point clouds~\cite{morrison:rss2018,mousavian:iccv2019}. 
While promising results have been shown, these methods work only on an open-loop sequential setting in which the human hand has to stay still once the robot starts to move~\cite{rosenberger:ral2021}, or need complex hand-designed cost functions for grasp selection~\cite{yang:icra2021} and robot motion planning~\cite{yang:icra2022, marturi:2019dynamic} for reactive handovers, which requires expertise in robot motion and control. 
Hence, these methods are difficult to reproduce and deploy to new environments.
Progress towards dynamic simultaneous motion has been shown by a learning-based method \cite{wang:corl2021a}, using state inputs, leaving an open challenge for training policies that receive visual input directly.
In contrast, we propose to learn control policies together with grasp prediction for handovers in an end-to-end manner from segmented point clouds with a deep neural net. 
To facilitate easy and fair comparisons among different handover methods, \cite{chao:icra2022} propose a physics-simulated environment with diverse objects and realistic human handover behavior collected by a mocap system~\cite{chao:cvpr2021}. They provide benchmark results of several previous handover systems, including a learning-based grasping policy trained with static objects~\cite{wang:corl2021b}. 
However, learning a safe and efficient handover policy is not trivial with a human-in-the-loop, which we address in this work.


\vspace{-3mm}
\paragraph{Policy Learning for Grasping}
Object grasping is an essential skill for many robot tasks, including handovers. 
Prior works usually generate grasp poses given a known 3D object geometry such as object shape or pose~\cite{bicchi:icra2000,miller:ram2004,bohg:tro2014}, which is nontrivial to obtain from real-world sensory input such as images or point clouds. 
To overcome this, recent works train deep neural networks to predict grasps from sensor data~\cite{kleeberger2020survey} and compute trajectories to reach the predicted grasp pose. 
Though 3D object geometry is no longer needed, the feasibility is not guaranteed since the grasp prediction and trajectory planning are computed separately. 
Some recent works directly learn grasping policies given raw sensor data. 
\cite{kalashnikov:corl2018} propose a self-supervised RL framework based on RGB images to learn a deep Q-function from real-world grasps. 
To improve data efficiency, \cite{song:ral2020} use a low-cost handheld device to collect grasping demonstrations with a wrist-mounted camera. They train an RL-based 6-DoF closed-loop grasping policy with these demonstrations.
\cite{wang:corl2021b} combines imitation learning from expert data with RL to learn a control policy for object grasping from point clouds. 
Although this method performs well in HandoverSim \cite{chao:icra2022} when the human hand is not moving, it has difficulty coordinating with a dynamic human hand since the policy is learned with static objects. 
Instead, our policy is directly learned from large-scale dynamic hand-object trajectories obtained from the real world.
To facilitate the training for the dynamic case, we propose a two-stage teacher-student framework, that is conceptually inspired by \cite{chen:corl2021}, which has been proven critical through experiments.




\begin{figure*}[t]
\begin{center}
   \includegraphics[width=0.93\textwidth]{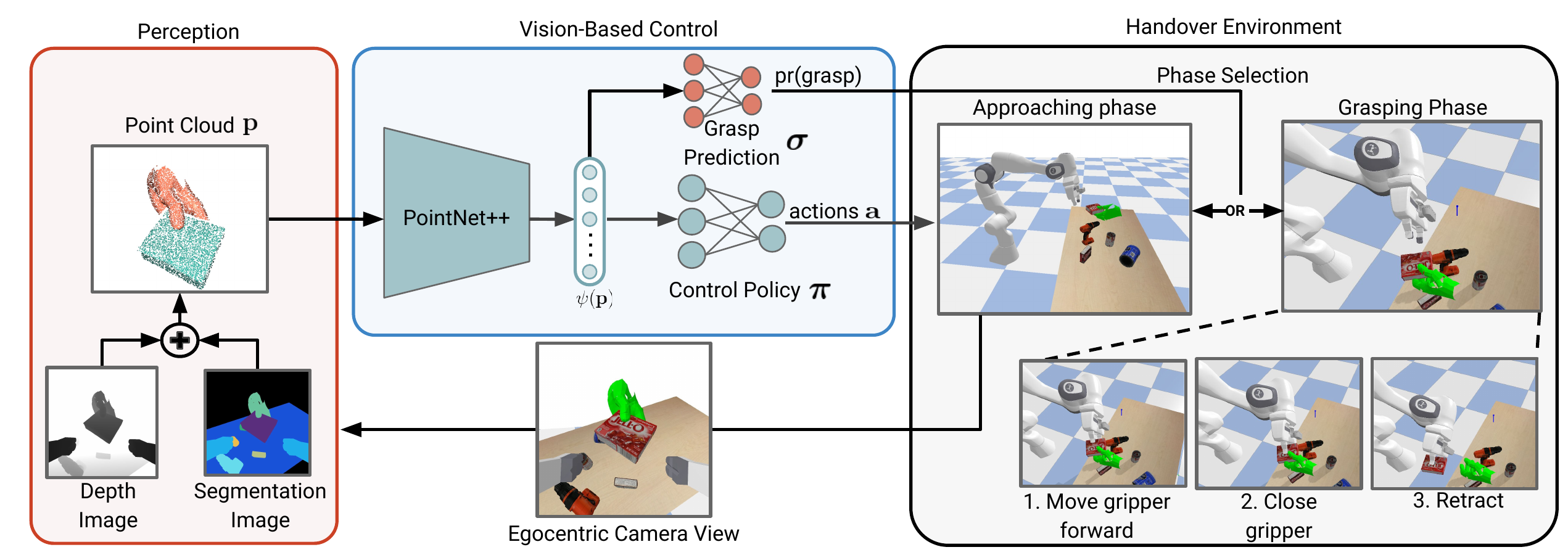}
\end{center}
   \vspace{-0.45cm}
   \caption{\textbf{Method Overview}. The \textbf{Perception} module takes egocentric RGB-D and segmentation images from the environment and outputs a hand/object segmented point cloud. Next, the segmented point cloud is passed to the the \textbf{Vision-based Control} module and processed by PointNet++\cite{qi:nips2017} to obtain a lower-dimensional representation. This embedding is used as input to both the control policy and the grasp predictor. Each task episode in the \textbf{Handover Environment} follows two phases: during the approaching phase, the robot moves towards a pre-grasp pose, driven by the control policy $\policy$ that outputs end-effector actions $\actions$. A learned grasp predictor monitors the motion and determines when the robot should switch into the grasping phase, which follows the steps: 1. moving the gripper forward from a pre-grasp to a grasping pose 2. closing the gripper 3. retracting the object to a designated location, after which the episode ends. }
   \vspace{-0.9em}
\label{fig:method_overview}
\end{figure*}

\section{Background}

\subsection{Reinforcement Learning}
\label{sec:background_rl}
\paragraph{MDP}~We formalize RL as a Markov Decision Process (MDP), that consists of a 5-tuple $\mdp = (\mathcal{S}, \mathcal{A}, \mathcal{R}, \mathcal{T}, \gamma)$, where $\mathcal{S}$ is the state space, $\mathcal{A}$ the action space, $\mathcal{R}$ a scalar reward function, $\mathcal{T}$ a transition function that maps state-action pairs to distributions over states, and $\gamma$ a discount factor. The goal is to find a policy that maximizes the long-term reward: $\policy^* = \argmax_{\policy} \mathbb{E} \sum_{t=0}^{t=T} \gamma^{t}\mathcal{R}(\statevec_t)$, with $\statevec_t \sim \mathcal{T}(\statevec_{t-1}, \actions_{t-1})$ and $\actions_{t-1} \sim \policy(\statevec_{t-1})$.

\paragraph{Learning Algorithm}~In this work, we use TD3 \cite{fujimoto:icml2018}, a common algorithm for continuous control. It is an actor-critic method, which consists of a policy $\policy_\theta(\statevec)$ (actor) and a Q-function approximator $Q_\phi(\statevec,\actions)$ (critic) that predicts the expected return from a state-action pair. Both are represented by neural networks with parameters $\theta$ and $\phi$. TD3 is off-policy, and hence there is a replay buffer in which training transitions are stored. During training, both the actor and critic are updated using samples from the buffer.
To update the critic, we minimize the Bellman error:
{
\medmuskip=0mu
\thinmuskip=0mu
\thickmuskip=0mu
\delimitershortfall=-1pt
\begin{equation}
L_{\text{BE}}(\phi) = \mathbb{E}_{\mdp} \left [ \left( Q_{\phi}(\statevec_{t},  \actions_{t}) - \reward(\statevec_{t}, \actions_{t}) + \discountRate Q_{\phi}(\statevec_{t+1}, \actions_{t+1} ) \right) ^2   \right]
\label{eq:q_bellman}
\end{equation}
}

For the actor network, the policy parameters are trained to maximize the Q-values:
\begin{equation}
L_{\text{DDPG}}(\theta) = \mathbb{E}_{\policy} \left [Q_{\phi}(\statevec_{t}, \actions_{t})    \vert \statevec_{t}, \actions_{t} = \policy_{\theta}(\statevec_{t}) \right ]
\label{eq:actor_critic}
\end{equation} 
For more details, we refer the reader to \cite{fujimoto:icml2018}.



\subsection{HandoverSim Benchmark}
\label{sec:handoversim}
HandoverSim \cite{chao:icra2022} is a benchmark for evaluating H2R handover policies in simulation. The task setting consists of a tabletop with different objects, a Panda 7DoF robotic arm with a gripper and a wrist-mounted RGB-D camera, and a simulated human hand. The task starts with the human grasping an object and moving it to a handover pose. The robot should move to the object and grasp it. The task is successful if the object has been grasped from the human without collision and brought to a designated position without dropping. To accurately model the human, trajectories from the DexYCB dataset \cite{chao:cvpr2021}, which comprises a large amount of human-object interaction sequences, are replayed in simulation. Several baselines~\cite{wang:rss2020,yang:icra2021,wang:corl2021b} are provided for comparison. The setup in HandoverSim has only been used for handover performance evaluation purposes, whereas in this work we utilize it as a learning environment.

\section{Method}
\label{method}

The overall pipeline is depicted in \Fig{method_overview} and consists of three different modules: perception, vision-based control, and the handover environment. The perception module receives egocentric visual information from the handover environment and processes it into segmented point clouds. The vision-based control module receives the point clouds and predicts the next action for the robot and whether to approach or to grasp the object. This information is passed to the handover environment, which updates the robot state and sends the new visual information to the perception module. Note that the input to our method comes from the wrist-mounted camera, i.e., there is no explicit information, such as object or hand pose, provided to the agent. We will now explain each of the modules of our method in more detail.

\subsection{Handover Environment}
\label{sec:handover_env}
We split the handover task into two distinct phases (see \Fig{method_overview}).
First, during the \emph{approaching phase}, the robot moves to a pre-grasp pose that is close to the object by running the learned control policy $\policy$. A learned grasp predictor $\grasppred$ continuously computes a grasp probability to determine when the system can proceed to the second phase. Once the pre-grasp pose is reached and the grasp prediction is confident to take over the object from the human, the task will switch to the \emph{grasping phase}, in which the end-effector moves forward to the final grasp pose in open-loop fashion and closes the gripper to grasp the object. Finally, after object grasping, the robot follows a predetermined trajectory to retract to a base position and complete the episode. This task logic is used in both our simulation environment and the real robot deployment.
Sequencing based on a pre-grasp pose is widely used in literature for dynamic grasping~\cite{akinola:iros2021}.
 

We follow the HandoverSim task setup \cite{chao:icra2022}, where the human hand and objects are simulated by replaying data from the DexYCB dataset \cite{chao:cvpr2021} (see \Sec{handoversim}).  
First, actions $\actions$ in the form of the next 6DoF end-effector pose (translation and rotation) are received from the policy $\policy(\actions| \statevec)$. We then convert the end-effector pose into a target robot configuration using inverse kinematics. Thereafter, we use PD-controllers to compute torques, which are applied to the robot. Finally, the visual information is rendered from the robot's wrist-mounted RGB-D camera and sent to the perception module. 

\subsection{Perception}

Our policy network takes a segmented hand and object point cloud as input. 
In the handover environment, we first render an egocentric RGB-D image from the wrist camera. Then we obtain the object point cloud $\pcvec_o$ and hand point cloud $\pcvec_h$ by overlaying the ground-truth segmentation mask with the RGB-D image.
Since the hand and object may not always be visible from the current egocentric view, we keep track of the last available point clouds. The latest available point clouds are then sent to the control module. 

\subsection{Vision-Based Control}
\label{sec:static_handover}

\paragraph{Input Representation}
Depending on the amount of points contained in the hand point cloud $\pcvec_h$ and object point cloud $\pcvec_o$, we down- or upsample them into constant size. Next, we concatenate the two point clouds into a single point cloud $\pcvec$ and add two one-hot-encoded vectors to indicate the locations of object and hand points within $\pcvec$. We then encode the point cloud into a lower dimensional representation $\psi(\pcvec)$ by passing it through PointNet++ \cite{qi:nips2017}. Finally, the lower dimensional encoding $\psi(\pcvec)$ is passed on to the control policy $\policy$ and the grasp prediction network $\grasppred$. 

\vspace{-3mm}
\paragraph{Control Policy} The policy network $\policy(\actions|\psi(\pcvec))$ is a small, two-layered MLP that takes the PointNet++ embedding as input state ($\statevec=\psi(\pcvec)$) and predicts actions $\actions$ that correspond to the change in 6DoF end-effector pose. These are passed on to the handover environment. 

\vspace{-3mm}
\paragraph{Grasp Prediction}
We introduce a grasp prediction network $\grasppred(\psi(\pcvec))$ that predicts when the robot should switch from approaching to executing the grasping motion (cf. \Fig{method_overview}). We model grasp prediction as a binary classification task. The input corresponds to the PointNet++ embedding $\psi(\pcvec)$, which is fed through a 3-layered MLP. The output is a probability that indicates the likelihood of a successful grasp given the current point cloud feature. If the probability is above a tunable threshold, we execute an open-loop grasping motion. The model is trained offline with pre-grasp poses attained from \cite{eppner:isrr2019}. We augment the dataset by adding random noise to pre-grasp poses. To determine the labels, we initialize the robot with the pre-grasp poses in the physics simulation and execute the forward grasping motion. The label is one if the grasp is successful, and zero otherwise. We use a binary cross-entropy loss for training. 
\begin{figure*}[t]
\begin{center}
   \includegraphics[width=0.95\textwidth]{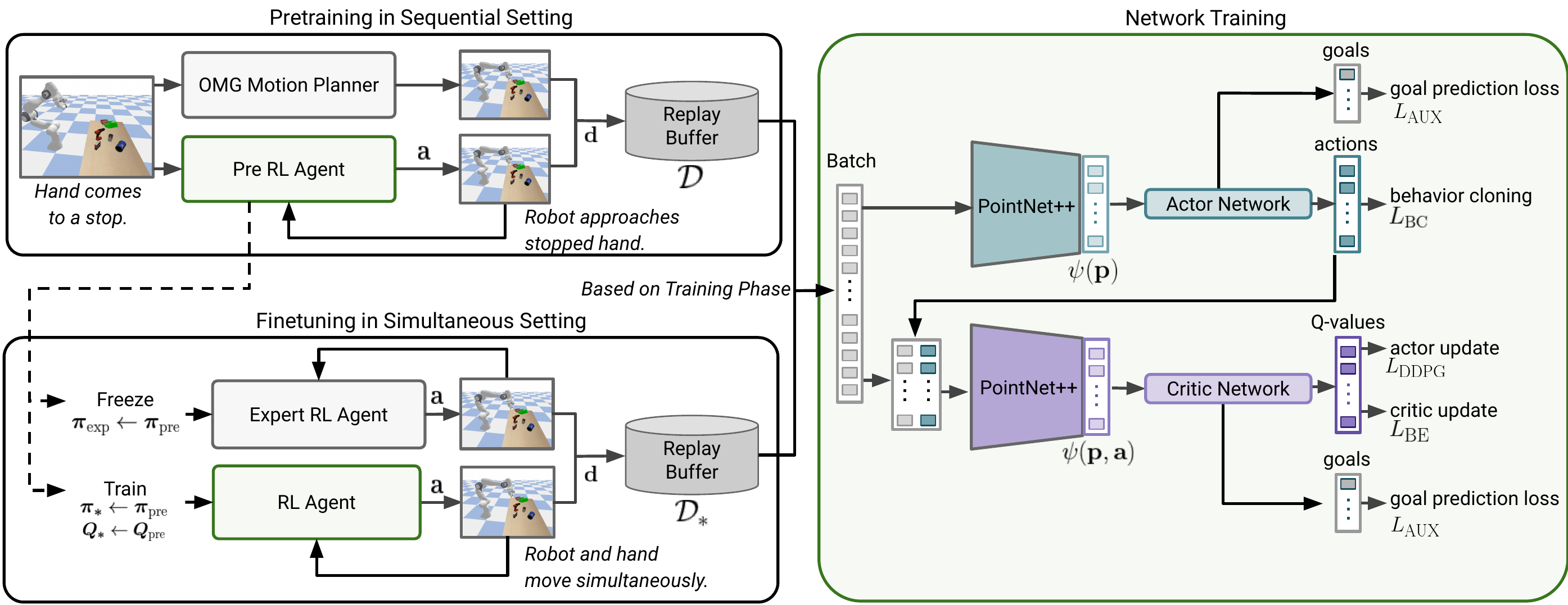}
\end{center}
   \vspace{-0.23cm}
   \caption{\textbf{Training Procedure}. In the \textbf{pretraining stage} (top left box), the human hand is stationary. We alternate between collecting expert demonstrations via motion planning and exploration data with the RL policy $\policypre$. Transitions $\transitionvec$ are stored in a replay buffer $\replay$. During training (green box, right), a batch of randomly sampled transitions from the replay buffer is passed through PointNet++ and the actor and critic networks. In the \textbf{finetuning stage} (bottom left box), the human and robot move concurrently. The expert motion planner is replaced by the expert policy $\policyexp$, which shares the weights of the pretrained policy $\policypre$. This policy network will be kept frozen for the rest of training and serves as a regularizer for the RL agent. The RL agent's actor network $\policyft$ and critic network $\criticft$ are also initialized with the weights of pretrained agent's networks, but the model will be updated during finetuning. In this stage, transitions are stored in a new replay buffer $\replay_*$. Data is sampled solely from this buffer during finetuning.
   }
   \vspace{-0.16cm}
\label{fig:training_procedure}
\end{figure*}

\subsection{Two-Stage Teacher-Student Training}\label{sec:training_procedure}
We aim at training a handover policy capable of moving simultaneously with the human. Training this policy directly in the setting of dynamic motion is challenging because expert demonstrations with open-loop planners to guide training can only be obtained when the human is stationary. A key contribution of our work is a two-stage training scheme for handovers that incrementally trains the policy to alleviate this challenge. In the first stage, we pretrain in a setting where the robot only starts moving once the human has stopped (\textbf{sequential}). This pretrained policy is further finetuned in the second stage in which the human and robot move simultaneously (\textbf{simultaneous}).

\begin{table*}[t]
 \centering
 \small
 \begin{tabular}{l|l|cccc|cccc}

  \hline
  & & \multirow{2}{*}{success (\%)} & \multicolumn{3}{c|}{mean accum time (s)} & \multicolumn{4}{c}{failure (\%)} \\
  & & & exec & plan & total & contact & drop & timeout & total \\
  \hline
  \parbox[t]{2mm}{\multirow{5}{*}{\rotatebox[origin=c]{90}{Sequential}}}
  & OMG Planner~\cite{wang:rss2020} $\dagger$        & 62.50 & 8.309          & 1.414          &  9.722          &  27.78          & ~~8.33 & ~~1.39 & 37.50 \\
  & \gc Yang et al.~\cite{yang:icra2021} $\dagger$  & \gc 64.58  & \gc 4.864  & \gc
  0.036& \gc 4.900          &  \gc 17.36          &  \gc 11.81          & \gc ~~6.25 & \gc 35.42         \\  
\cline{2-10}
  & GA-DDPG~\cite{wang:corl2021b}   & 50.00          & 7.139        & 0.142 & 7.281          & ~\textbf{4.86} &  19.44          &  25.69 & 50.00        \\
  & \gc GA-DDPG~\cite{wang:corl2021b} finetuned  & \gc 57.18         & \gc  \textbf{6.324}        & \gc \textbf{0.086} & \gc \textbf{6.411}      & \gc ~6.48 &  \gc 27.08          &  \gc ~9.26  & \gc 42.82      \\
 
  & Ours    & \textbf{75.23} & 7.743 & 0.177 & 7.922  & ~9.26 & \textbf{13.43} & ~\textbf{2.08}  & \textbf{24.77}    \\
\hline

  \parbox[t]{2mm}{\multirow{3}{*}{\rotatebox[origin=c]{90}{Simult.}}}
  
  & GA-DDPG~\cite{wang:corl2021b} &  36.81          & \textbf{4.664} & 0.132         & \textbf{4.796} &  ~9.03          &  25.00          &  29.17  &   63.19     \\
  & \gc GA-DDPG~\cite{wang:corl2021b} finetuned  & \gc 54.86          & \gc 4.832          & \gc \textbf{0.082} & \gc 4.914         & \gc ~\textbf{6.71} &  \gc 26.39          & \gc 12.04    & \gc 45.14      \\
 
  & Ours   & \textbf{68.75} & 6.232 & 0.178 & 6.411  & ~8.80 & \textbf{17.82} & ~\textbf{4.63} & \textbf{31.25}      \\
  
\bottomrule
  
 \end{tabular}
 \vspace{-1mm}
 \caption{\textbf{HandoverSim Benchmark Evaluation.} Comparison of our method against various baselines from the HandoverSim benchmark \cite{chao:icra2022}. In the sequential setting, we find that our baseline achieves better overall success rates than the baselines. In the simultaneous setting, we outperform the applicable baselines by large margins. The results for our method are averaged across 3 random seeds. $\dagger$: both methods \cite{wang:rss2020,yang:icra2021} are evaluated with ground-truth states in \cite{chao:icra2022} and thus are not directly comparable with ours.}
 \vspace{-1em}
 \label{tab:quantitative}
\end{table*}

\vspace{-2mm}
\paragraph{Pretraining in Sequential Setting}
\label{sec:sequential}
In the sequential setting, the robot starts moving once the human has come to a stop (see \Fig{training_procedure}, top left). To grasp the object from the stationary human hand, we 
 leverage motion planning to provide expert demonstrations. During data collection, we alternate between motion planning and RL-based exploration. In both cases, we store the transitions $ \transitionvec_{t}= \{ \pcvec_t,\actions_t, \goalvec_t, \rewardvec_t, \pcvec_{t+1}, \expertvec_t \}$ in a replay buffer $\replay$, from which we sample during network training. The term $\pcvec_t$ and $\pcvec_{t+1}$ indicate the point cloud and the next point cloud, $\actions_t$ the action, $\goalvec_t$ the pre-grasp goal pose, $\rewardvec_t$ the reward, and $\expertvec_t$ an indicator of whether the transition is from the expert. 
 
Inspired by~\cite{wang:corl2021b}, we collect expert trajectories with the OMG planner \cite{wang:rss2020} \update{that leverages ground-truth states}. Note that some expert trajectories generated by the planner result in collision with the hand, which is why we introduce an offline pre-filtering scheme. We first parse the ACRONYM dataset \cite{eppner:acronym2020} for potential grasps. We then run collision checking to filter out grasps where the robot and human hand collide. For the set of remaining collision-free grasps, we plan trajectories to grasp the object and execute them in open-loop fashion. 
On the other hand, the RL policy $\policypre$ explores the environment and receives a sparse reward, i.e., the reward is one if the task is completed successfully, otherwise zero. Hence, collisions with the human will get implicitly penalized by not receiving any positive reward.

\vspace{-3mm}
\paragraph{Finetuning in Simultaneous Setting}
In this setting, the human and robot move at the same time. Hence, we cannot rely on motion and grasp planning to guide the policy. On the other hand, simply taking the pre-trained policy $\policypre$ from the sequential setting and continue training it without an expert leads to an immediate drop in performance. Hence, we introduce a self-supervision scheme for stability reasons, i.e., we want to keep the finetuning policy close to the pre-trained policy. To this end, we replace the expert planner from the sequential setting by an expert policy $\policyexp$, which is initialized with the weights of the pre-trained policy $\policypre$ that already provides a reasonable prior policy (see \Fig{training_procedure} bottom left). Therefore, we have two policies: i) the expert policy $\policyexp$ as proxy for the motion and grasping planner. We freeze the network weights of this policy, ii) the finetuning policy $\policyft$ and critic $\criticft$, which are initialized with the weights of the pre-trained policy $\policypre$ and critic $\criticpre$, respectively. We proceed to train these two networks using the loss functions which we describe next.

\vspace{-3mm}
\paragraph{Network Training}
During training, we sample a batch of random transitions from the replay buffer $\replay$. The policy network is trained using a combination of behavior cloning, RL-based losses and an auxiliary objective. In particular, the policy is updated using the following loss function:
\begin{equation}
    L(\theta) = \lambda L_{\text{BC}}+(1-\lambda)L_{\text{DDPG}}+L_{\text{AUX}},
    \label{eq:loss_actor}
\end{equation}
where $L_{\text{BC}}$ is a behavior cloning loss that keeps the policy close to the expert policy, $L_{\text{DDPG}}$ is the standard actor-critic loss described in \Eq{actor_critic}, and $L_{\text{AUX}}$ is an auxiliary objective that predicts the grasping goal pose of the end-effector. The coefficient $\lambda$ balances the behavior cloning and the RL objective. The critic loss is defined as:
\vspace{-1mm}
\begin{equation}
    \vspace{-1mm}
L(\phi) =  L_{\text{BE}}+L_{\text{AUX}},
\label{eq:loss_critic}
    \vspace{-1mm}
\end{equation}  
where $L_{\text{BE}}$ indicates the Bellman error from \Eq{q_bellman} and $L_{\text{AUX}}$ is the same auxiliary loss used in \Eq{loss_actor}.
We refer the reader to supplementary material or \cite{wang:corl2021b} for more details.


\section{Experiments}
We first evaluate our approach in simulation using the HandoverSim benchmark (Sec.~\ref{sec:simulation_evaluation}). Next, we investigate the performance of sim-to-sim transfer by evaluating the trained models on the test environments powered by a different physics engine (Sec.~\ref{sec:sim2sim}). Finally, we apply the trained model to a real-world robotic system and analyze the performance of sim-to-real transfer (Sec.~\ref{sec:sim2real}).

\subsection{Simulation Evaluation}
\label{sec:simulation_evaluation}

\begin{figure*}
    \centering
    \begin{subfigure}[t]{0.9\textwidth}
         \centering
            \includegraphics[width=\textwidth]{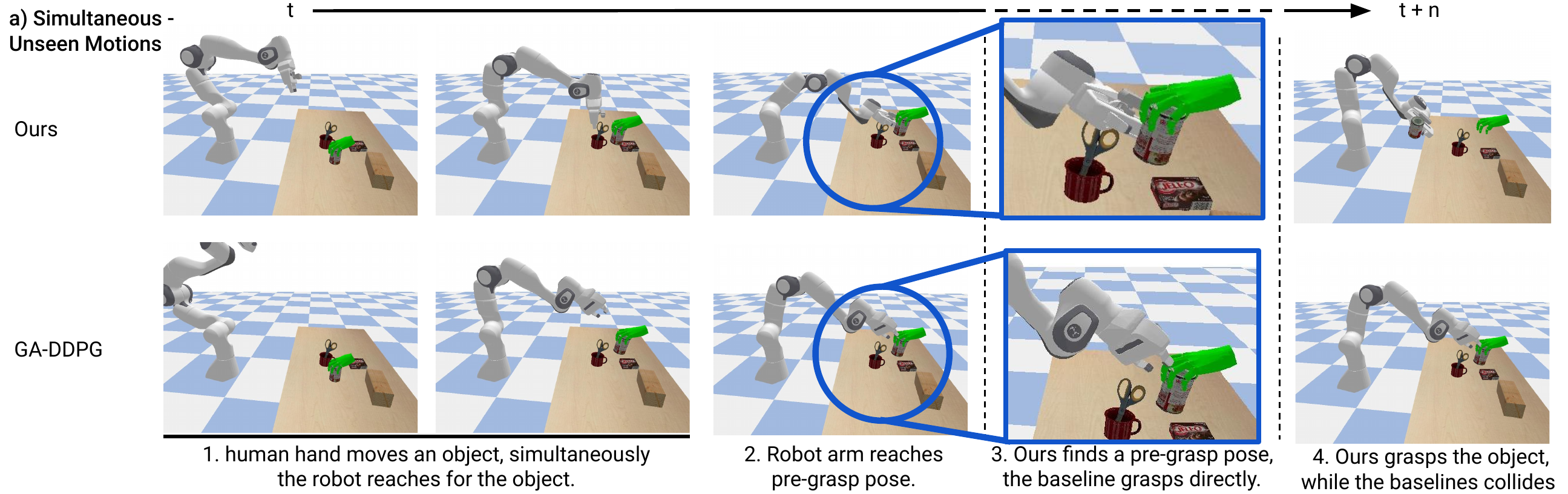}
            \label{fig:quali_a}
    \end{subfigure}\vspace{-0.4cm}
    \begin{subfigure}[b]{0.9\textwidth}
         \centering
            \includegraphics[width=\textwidth]{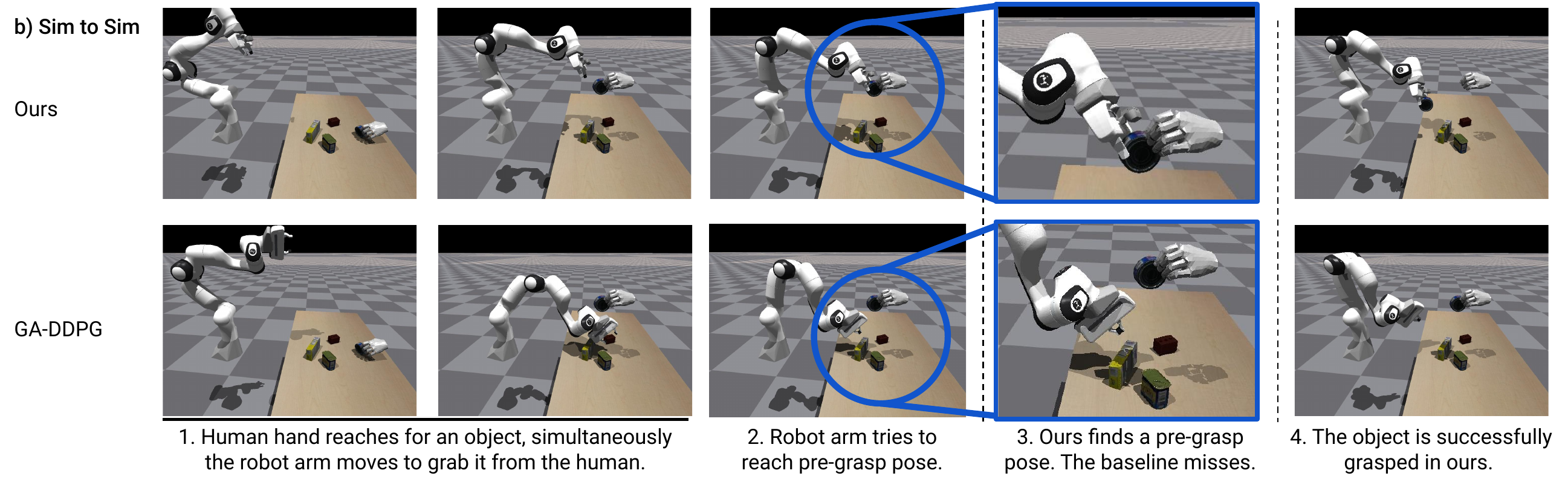}
            \label{fig:quali_b}
    \end{subfigure}\vspace{-0.4cm}
    \begin{subfigure}[b]{0.9\textwidth}
         \centering
            \includegraphics[width=\textwidth]{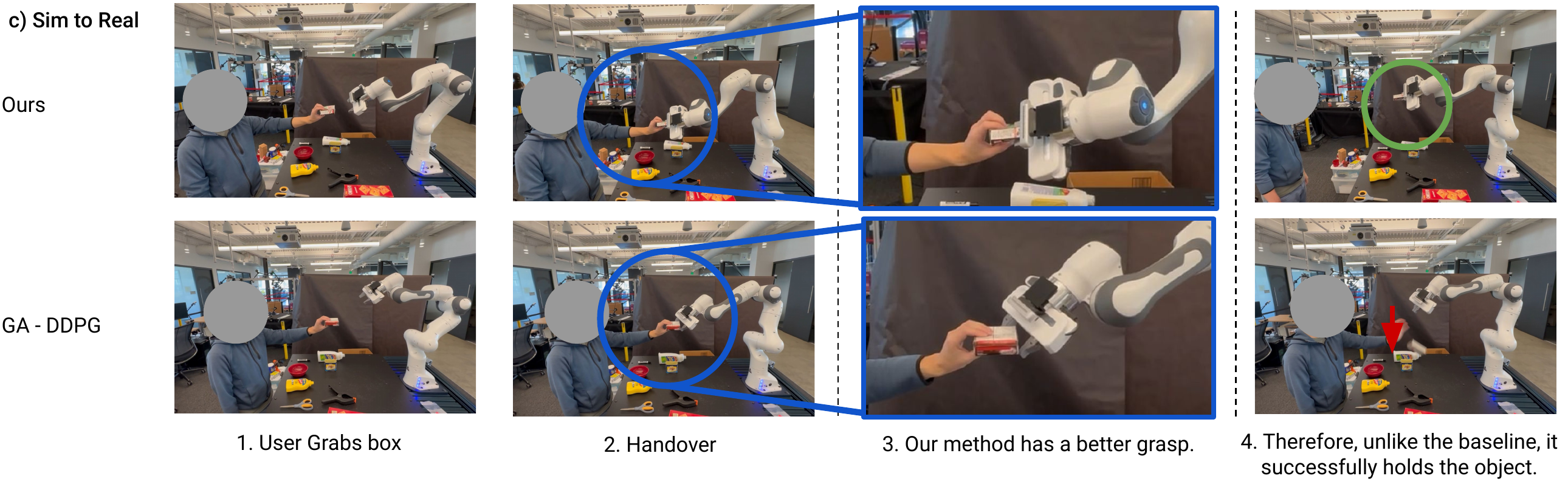}
            \label{fig:quali_c}
     \end{subfigure}
    \vspace{-0.6cm}
    \caption{\textbf{Qualitative results.} We provide a comparison to show our methods' advantages over GA-DDPG \cite{wang:corl2021b}. (a) Our method reacts to the moving human, while the baseline tries to go for a grasp directly, which leads to collision. (b) In the sim-to-sim transfer, we often find that the baseline does not find a grasp on the object. (c) In the sim-to-real experiment, GA-DDPG usually tries to get to a grasp directly, while our method adjusts the gripper into a stable grasping pose first. See the video in \supmat for more qualitative examples.
    }
    \vspace{-1.2em}
    \label{fig:qualitative}
\end{figure*}
\paragraph{Setup}~HandoverSim~\cite{chao:icra2022} contains 1,000 unique H2R handover scenes divided into train, val, and test splits. Each scene contains a unique human handover motion. We evaluate on the ``s0'' setup which contains 720 training and 144 testing scenes. See the \supmat for evaluations on unseen objects, subjects, and handedness. Following the evaluation of GA-DDPG~\cite{wang:corl2021b} in~\cite{chao:icra2022}, we consider two settings: (1) the ``sequential'' setting where the robot is allowed to move only after the human hand reaches the handover location and remains static there (i.e., ``hold'' in~\cite{chao:icra2022}), and (2) the ``simulataneous'' setting where the robot is allowed to move from the beginning of the episode (i.e., ``w/o hold'' in~\cite{chao:icra2022}).

\vspace{-3mm}
\paragraph{Metrics}~We follow the evaluation protocol in HandoverSim~\cite{chao:icra2022}. A handover is considered successful if the robot grasps the object from the human hand and moves it to a designated location. A failure is claimed and the episode is terminated if any of the following three conditions occur: (1) the robot collides with the hand (\emph{contact}), (2) the robot drops the object (\emph{drop}), or (3) a maximum time limit is reached (\emph{timeout}). Besides efficacy, the benchmark also reports efficiency in time. The time metric is further broken down into (1) the execution time (\emph{exec}), i.e., the time to physically move the robot, and (2) the planning time (\emph{plan}), i.e., the time spent on running the policy. All reported metrics are averaged over the rollouts on the test scenes.

\vspace{-3mm}
\paragraph{Baselines}~Our primary baseline is GA-DDPG~\cite{wang:corl2021b}. Besides comparing with the original model (i.e., trained in~\cite{wang:corl2021b} for table-top grasping and evaluated in~\cite{chao:icra2022}), we additionally compare with a variant finetuned on HandoverSim (``GA-DDPG~\cite{wang:corl2021b} finetuned''). For completeness, we also include two other baselines from~\cite{chao:icra2022}: ``OMG Planner~\cite{wang:rss2020}'' and ``Yang et al.~\cite{yang:icra2021}''. However, both of them are evaluated with ground-truth state input in~\cite{chao:icra2022} and thus are not directly comparable with our method.


\vspace{-3mm}
\paragraph{Results}~\Tab{quantitative} reports the evaluation results on the test scenes. 
In the sequential setting, our method significantly outperforms all the baselines in terms of success rate, even compared to methods that use state-based input. Our method is slightly slower on average than GA-DDPG in terms of total time needed for handovers.
In the simultaneous setting, our method clearly outperforms GA-DDPG, which has low success rates. Qualitatively, we observe that GA-DDPG directly tries to grasp the object from the user while it is still moving, while our method follows the hand and finds a feasible grasp once the hand has come to a stop, resulting in a trade-off on the overall execution time. We provide a qualitative example of this behavior in \Fig{qualitative} (a) \update{and in the supplementary video. We also refer to the \supmat for a discussion of limitations and a robustness analysis of our pipeline under noisy observations.}


\textbf{\begin{table}[!]
 \vspace{-0mm}
 \centering
 \small
  \resizebox{0.9\columnwidth}{!}{%
 \begin{tabular}{l|cccc}
  \hline
  \multicolumn{5}{c}{Ablation Study} \\
  \hline
  & \multirow{2}{*}{success (\%)} &  \multicolumn{3}{c}{failure (\%)} \\
  & & contact & drop & timeout \\
  \hline
 w/ RGBDM + ResNet18 & 34.10  & ~~\textbf{6.20} &  45.80     &  13.90          \\
  \rowcolor{Gray}

    w/ third person view & 60.42  & ~~9,95 &  25.69     &  ~~3.94          \\

   w/o hand point cloud & 59.03 & 24.07 &  \textbf{11.58}  &  ~~5.32         \\

     \rowcolor{Gray}
   w/o aux prediction & 70.60  & 10.65 &  16.20     & ~~2.54          \\

   w/o standoff  & 52.55  & ~~7.87 &  36.80     &  ~~2.78          \\
      \rowcolor{Gray}
  w/o finetuning & 73.38 & ~~9.03 & 13.89 & ~~3.70      \\
  

  Ours & \textbf{75.23} & ~~9.26 & 13.43 & ~~\textbf{2.08} \\ 
  \hline
    \rowcolor{Gray}
  w/o finetuning simult. & 62.27 & 11.81 & 20.37 & ~~5.56      \\

  Ours simult. & \textbf{68.75} & \textbf{8.8} & \textbf{17.82} & ~~\textbf{4.63} \\ 
  \hline
 \end{tabular}
 }
 \vspace{-2mm}
 \caption{\textbf{Ablation.} \update{We ablate the vision backbone, hand perception, and egocentric view. We also study the effect of finetuning, the auxiliary prediction, and splitting the task into two phases. All design choices are crucial aspects of our method with regards to overall performance.} Results are averaged over 3 random seeds.
 }
 \vspace{-0mm}
 \label{tab:ablation}
\end{table}}

\vspace{-6mm}
\paragraph{Ablations}~\update{We evaluate our design choices in an ablation study and report the results in \Tab{ablation}. We analyze the vision backbone by replacing PointNet++ with a ResNet18~\cite{he2016resnet} that processes the RGB and depth/segmentation (DM) images. Similar to the findings in GA-DDPG, the PointNet++ backbone performs better. Next, we train our method from third person view instead of egocentric view and without active hand segmentation (\textit{w/o hand point cloud}), i.e., the policy only perceives the object point cloud but not the hand point cloud. We also ablate the auxiliary prediction (\textit{w/o aux prediction}) and evaluate a variant that directly learns to approach and grasp the object instead of using the two task phases of approaching and grasping (\textit{w/o standoff}). Lastly, we compare against our pretrained model, which was only trained in the sequential setting without finetuning (\textit{w/o finetuning}). We find that the ablated components comprise important elements of our method.  The results indicate an increased amount of hand collision or object drop in all ablations. A closer analysis in the simultaneous setting shows that our finetuned model outperforms the pretrained model.}

\subsection{Sim-to-Sim Transfer}
\label{sec:sim2sim}
Instead of directly transferring to the real world, we first evaluate the robustness of the models by transferring them to a different physics simulator. We re-implement the HandoverSim environment following the mechanism presented in~\cite{chao:icra2022} except for replacing the backend physics engine from Bullet~\cite{coumans:2021} to Isaac Gym~\cite{makoviychuk:neuripstdb2021}. We then evaluate the models trained on the original Bullet-based environment on the test scenes powered by Isaac Gym. The results are presented in \Tab{sim2sim_singlecol}. We observe a significant drop for GA-DDPG on the success rates (i.e., to below 20\%) in both settings. Qualitatively, we see that grasps are often either missed completely or only partially grasped (see \Fig{qualitative} (b)). On the other hand, our method is able to retain higher success rates. Expectedly, it also suffers from a loss in performance. We analyze the influence of our grasp predictor on transfer performance and compare against a variant where we execute the grasping motion after a fixed amount of time (\emph{Ours w/o grasp pred.}), which will leave the robot enough time to find a pre-grasp pose. Part of the performance drop is caused by the grasp predictor initiating the grasping phase at the wrong time, which can be improved upon in future work.

\begin{table}[t]
 \centering
 \small
 \resizebox{0.9\columnwidth}{!}{%
 \begin{tabular}{l|l|cccc}
  \hline
  \multicolumn{6}{c}{Sim-to-Sim} \\
  \hline
  & & \multirow{2}{*}{success (\%)} &  \multicolumn{3}{c}{failure (\%)} \\
  & & & contact & drop & timeout  \\
  \hline
  \parbox[t]{2mm}{\multirow{4}{*}{\rotatebox[origin=c]{90}{Sequential}}}
  & GA-DDPG~\cite{wang:corl2021b} &  19.44         & ~~\textbf{4.86} & 47.22          & 28.47      \\
  & \gc GA-DDPG~\cite{wang:corl2021b} finetuned  & \gc 11.81      & \gc ~~6.25 &  \gc 68.75          &  \gc 13.19      \\
  
  &  Ours & 44.21  & ~~9.49 & 40.51 & ~~5.79   \\ 
  
  & \gc Ours w/o grasp & \gc \textbf{54.40} & \gc ~~7.87 & \gc \textbf{33.34} & \gc ~~\textbf{4.40}  \\

  \hline

  \parbox[t]{2mm}{\multirow{4}{*}{\rotatebox[origin=c]{90}{Simult.}}}
  
  & GA-DDPG~\cite{wang:corl2021b} & 11.11           & 15.97          & 48.61          & 24.31       \\
  & \gc GA-DDPG~\cite{wang:corl2021b} finetuned  & \gc 16.67   & \gc ~~9.72 &  \gc 63.89          & \gc ~~9.72     \\
 
  & Ours    & 39.58 & ~~\textbf{9.03} & 43.75  & ~~7.64    \\
  & \gc Ours w/o grasp pred. & \gc \textbf{47.92} & \gc 10.65 & \gc \textbf{35.88} & \gc ~~\textbf{5.56}   \\ 
  
\bottomrule
  
 \end{tabular}
 }
 \vspace{-2mm}
 \caption{\textbf{Sim-to-Sim Experiment}. We evaluate sim-to-sim transfer of the learning-based method to Isaac Gym \cite{makoviychuk:neuripstdb2021}, Our method shows better transfer capabilities than GA-DDPG \cite{wang:corl2021b}. }
 \vspace{-4mm}
 \label{tab:sim2sim_singlecol}
\end{table}

\subsection{Sim-to-Real Transfer}
\label{sec:sim2real}

Finally, we deploy the models trained in HandoverSim on a real robotic platform. We follow the perception pipeline used in~\cite{yang:icra2021,wang:corl2021b} to generate segmented hand and object point clouds for the policy, and use the output to update the end effector's target position. We compare our method against GA-DDPG~\cite{wang:corl2021b} with two sets of experiments: (1) a pilot study with controlled handover poses and (2) a user evaluation with free-form handovers. For experimental details and the full results, please see the \supmat.




\vspace{-3mm}
\paragraph{Pilot Study}~We first conduct a pilot study with two subjects. The subjects are instructed to handover 10 objects from HandoverSim by grasping and presenting the objects in controlled poses. For each object, we test with 6 poses (3 poses for each hand) with varying object orientation and varying amount of hand occlusion, resulting in 60 poses per subject. The same set of poses are used in testing both our model and GA-DDPG~\cite{wang:corl2021b}. The success rates are shown \Tab{pilot_study}. Results indicate that our method outperforms GA-DDPG~\cite{wang:corl2021b} for both subjects on the overall success rate (i.e., 41/60 versus 21/60 for Subject 1). Qualitatively, we observe that GA-DDPG~\cite{wang:corl2021b} tends to fail more from unstable grasping as well as hand collision. Fig.~\ref{fig:qualitative} (c) shows two examples of the real world handover trials.

\begin{table}[]
 \centering
 \footnotesize
 \setlength{\tabcolsep}{2pt}
  \resizebox{0.9\columnwidth}{!}{%
 \begin{tabular}{l|C{1.27cm}C{1.27cm}|C{1.27cm}C{1.27cm}}
   \hline
                           & \multicolumn{2}{c}{Subject 1}                  & \multicolumn{2}{|c}{Subject 2}                 \\
  \cline{2-5}
                           & GA-DDPG                & \multirow{2}{*}{Ours} & GA-DDPG                & \multirow{2}{*}{Ours} \\
                           & ~\cite{wang:corl2021b} &                       & ~\cite{wang:corl2021b} &                       \\
  \hline
  011\_banana              & ~~3 / ~~6              & \textbf{~~6 / ~~6}    & \textbf{~~6 / ~~6}     & ~~5 / ~~6             \\
  \rowcolor{Gray}
  037\_scissors            & ~~2 / ~~6              & \textbf{~~5 / ~~6}    & ~~3 / ~~6              & \textbf{~~5 / ~~6}    \\
  006\_mustard\_bottle     & ~~1 / ~~6              & \textbf{~~3 / ~~6}    & ~~2 / ~~6              & \textbf{~~4 / ~~6}    \\
  \rowcolor{Gray}
  024\_bowl                & ~~3 / ~~6              & \textbf{~~4 / ~~6}    & \textbf{~~3 / ~~6}     & \textbf{~~3 / ~~6}    \\
  040\_large\_marker       & ~~0 / ~~6              & \textbf{~~4 / ~~6}    & ~~4 / ~~6              & \textbf{~~5 / ~~6}    \\
  \rowcolor{Gray}
  003\_cracker\_box        & \textbf{~~3 / ~~6}     & ~~2 / ~~6             & ~~0 / ~~6              & \textbf{~~2 / ~~6}    \\
  052\_extra\_large\_clamp & ~~1 / ~~6              & \textbf{~~4 / ~~6}    & \textbf{~~5 / ~~6}     & \textbf{~~5 / ~~6}    \\
  \rowcolor{Gray}
  008\_pudding\_box        & ~~3 / ~~6              & \textbf{~~6 / ~~6}    & \textbf{~~4 / ~~6}     & \textbf{~~4 / ~~6}    \\
  010\_potted\_meat\_can   & \textbf{~~2 / ~~6}     & \textbf{~~2 / ~~6}    & ~~3 / ~~6              & \textbf{~~4 / ~~6}    \\
  \rowcolor{Gray}
  021\_bleach\_cleanser    & ~~3 / ~~6              & \textbf{~~5 / ~~6}    & ~~3 / ~~6              & \textbf{~~4 / ~~6}    \\
  \hline
  \textbf{total}                      & 21 / 60                & \textbf{41 / 60}      & 33 / 60                & \textbf{41 / 60}   \\
    \hline
 \end{tabular}
 }
 \vspace{-1mm}
 \caption{\textbf{Sim-to-Real Experiment}. Success rates of the pilot study. Our method outperforms GA-DDPG~\cite{wang:corl2021b} for both subjects.}
 \label{tab:pilot_study}
\end{table}

\vspace{-3mm}
\paragraph{User Evaluation}~We further recruited $6$ users to compare the two methods and collected feedback from a questionnaire with Likert-scale and open-ended questions. In contrast to the pilot study, we asked the users to handover the 10 objects in ways that are most comfortable to them. We repeated the same experimental process for both methods, and counterbalanced the order to avoid bias. \update{From participants' feedback, the majority agreed that the timing of our method is more appropriate and our method can adjust between different object poses better. The interpretability of the robot's motion was also acknowledged by their comments}. Please see the \supmat for more details.
\vspace{-0mm}

\section{Conclusion}
In this work, we have presented a learning-based framework for human-to-robot handovers from vision input with a simulated human-in-the-loop. We have introduced a two-stage teacher-student training procedure. In our experiments we have shown that our method outperforms baselines by a significant margin on the HandoverSim benchmark \cite{chao:icra2022}. Furthermore, we have demonstrated that our approach is more robust when transferring to a different physics simulator and a real robotic system. 

\vspace{-1mm}
\paragraph*{\small Acknowledgements}\small~We thank Tao Chen and Adithyavairavan Murali for laying the groundwork, Lirui Wang for the help with GA-DDPG, and Mert Albaba, Christoph Gebhardt, Thomas Langerak and Juan Zarate for their feedback on the manuscript.

{\small
\bibliographystyle{ieee_fullname}
\bibliography{egbib}
}
\appendix



\newpage
{\large \textbf{Appendix - Supplementary Material}}
\vspace{2mm}

The supplementary material of this work contains \textbf{a video} and this document. See the table of contents below for an overview. The video and code are available at {\footnotesize{\url{https://handover-sim2real.github.io/}}}.

\appendix

\addcontentsline{toc}{section}{Appendix} 

\etocdepthtag.toc{mtappendix}
\etocsettagdepth{mtchapter}{none}
\etocsettagdepth{mtappendix}{subsection}
{
  \hypersetup{
    linkcolor = black
  }
  \tableofcontents
}

\section{Method Details}
\label{app_method}

\subsection{Task and Method}
\paragraph{Action Space}
Actions are defined as transformations of the end-effector pose. In particular, an action contains the relative transformation of the 3D translation and 3D rotation of the end-effector. More formally, given the current gripper pose $\mathcal{T}_t \in \mathbb{SE}(3)$ at time $t$, the policy $\policy$ predicts an action $\actions_t=\mathcal{T}_{\policy}$ that indicates the relative transformation of the end-effector. The next end-effector pose it then computed by transforming the current pose $\mathcal{T}_{t+1} =\mathcal{T}_t\mathcal{T}_{\policy}$. We then use inverse kinematics to compute the next robot pose that is sent to the PD-controllers of the physics engine.

\vspace{-3mm}
\paragraph{Goal Space}
Goals are defined in a similar way. They correspond to the relative transformation between the current 6DoF end-effector pose $\mathcal{T}_t$ and the pre-grasp pose $\mathcal{T}_{\text{pre}}$, i.e., $\goalvec_t = \mathcal{T}_{\text{pre}}\mathcal{T}_t^{-1}$. Note that in contrast to related work \cite{wang:corl2021b}, we define the goal as the pre-grasp pose from where the object can be grasped via a forward grasping motion (see main paper Sec. 4.1) instead of the actual grasp pose.

\begin{figure}[t]

 \centering
 \includegraphics[width=0.8\linewidth]{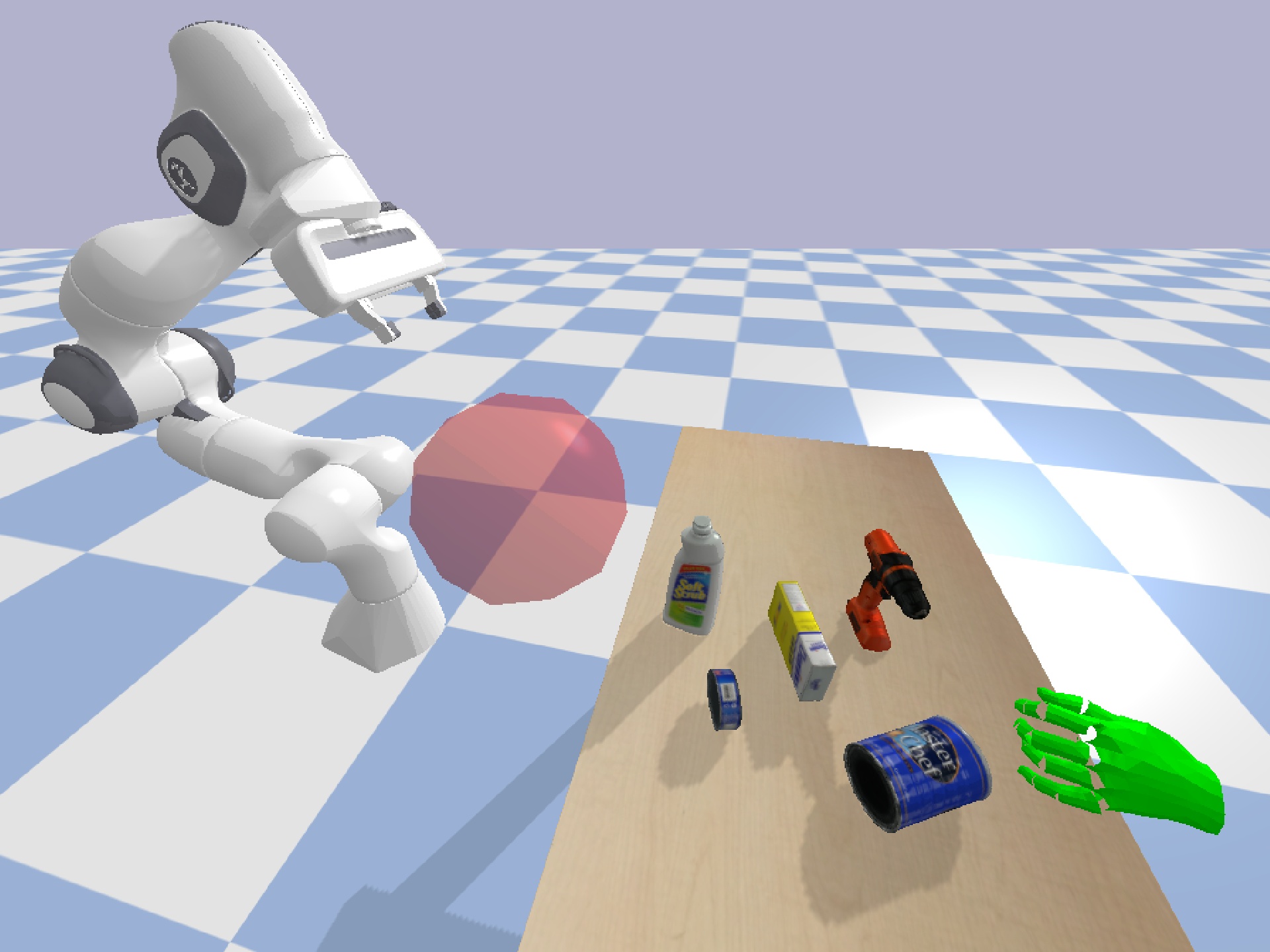}
 \caption{\small The simulation environment from HandoverSim. The red sphere indicates the goal region. Image source: Chao \etal \cite{chao:icra2022}.}

 \label{fig:environment}
\end{figure}

\vspace{-3mm}
\paragraph{Reward Function}
We use a sparse reward function. The reward is 1 if a task has been successfully completed, i.e., the robot has successfully taken over the object from the human and moved it to a predefined goal region (see red sphere in \Fig{environment}) without dropping or collision with the human hand. The reward is 0 in all other cases. If the robot collides with the human hand or the object is dropped, the episode terminates early and the reward stays 0.

\begin{figure*}[t]

 \centering
 \includegraphics[width=\linewidth]{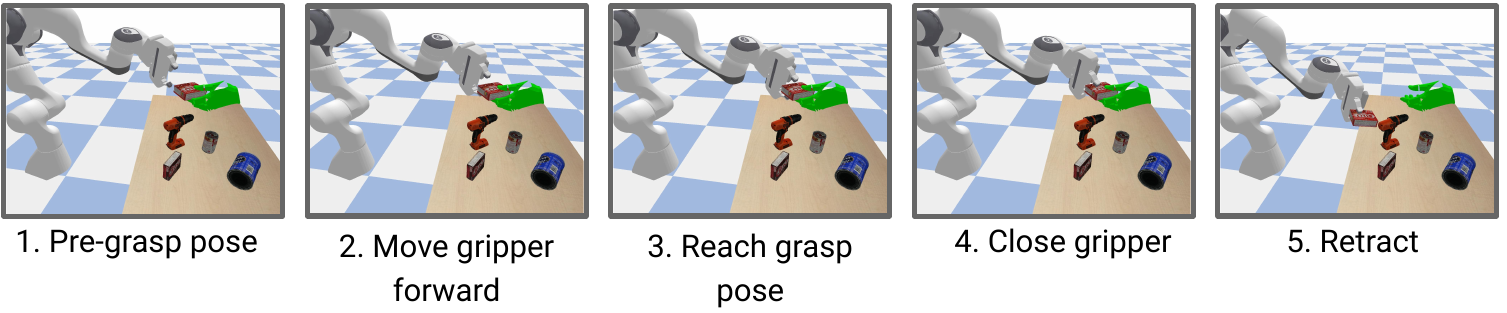}
 \vspace{-6mm}
 \caption{\small The grasping phase starts with a forward grasping motion that moves the end-effector from a pre-grasp pose to a grasp pose, then grasps the object, and finally moves it to a pre-defined goal region.}

 \label{fig:grasping_phase}
\end{figure*}

\vspace{-3mm}
\paragraph{Grasping Phase} The grasping phase comprises the forward motion from the pre-grasp pose to the grasp pose, the closing of the gripper, and the retraction of the object (see \Fig{grasping_phase}). The forward motion moves the end-effector 8 cm into the z-direction of the gripper. Then, the gripper is closed to grasp the object. Finally, a retraction trajectory is executed in open-loop fashion to bring the object into a goal location. We compute the trajectory by linearly interpolating the path between the end-effector and the end-effector goal location, and then transform the trajectory into robot poses using inverse kinematics. Note that the grasping phase is non-learning based and purely based on heuristics and open-loop control.

\vspace{-3mm}
\paragraph{Perception Module} To transform the segmented  object point cloud $\pcvec_o$ and hand point cloud $\pcvec_h$ into a single point cloud $\pcvec$ of constant size, we down- or up-sample both point clouds. Since there are usually more points contained in the segmented object point cloud (e.g., due to the hand being occluded), we use a ratio of 87.5\% to 12.5\% for the object and hand point clouds when sampling the single point cloud $\pcvec$. We add two one-hot encoded vectors to the point cloud $\pcvec$ to indicate which points are from the object or hand point clouds, respectively. We keep track of the latest available point cloud $\pcvec_{t-1}$. If there is no point cloud available at time $t$, we use the latest available point cloud $\pcvec_t=\pcvec_{t-1}$.

\subsection{Loss Functions}
We provide a more detailed description of the loss functions used to train our method. We mostly follow the description of \cite{wang:corl2021b}. 

\vspace{-3mm}
\paragraph{Policy Loss} As stated in the main paper (Eq.\ 3), our loss function for the policy is defined as:
\begin{equation}
    L(\theta) = \lambda L_{\text{BC}}+(1-\lambda)L_{\text{DDPG}}+L_{\text{AUX}}.
\end{equation}
We first introduce the point matching loss function \cite{li:eccv2018}:
\begin{equation}
    L_{\text{POSE}}(\mathcal{T}_1, \mathcal{T}_2) = \frac{1}{|X_g|} \sum_{x \in X_g} ||\mathcal{T}_1(x)-\mathcal{T}_2(x)||_1,
\end{equation}
where $X_g$ is a set of pre-defined points on the end effector. The loss computes the L1 norm between of these points after applying pose transformations $\mathcal{T}_1$ and $\mathcal{T}_2$ to the end-effector.
The behavior cloning loss is defined as :
\begin{equation}
    L_{\text{BC}}(\actions_*, \actions) = L_{\text{POSE}}(\actions_*, \actions).
\end{equation}
The loss computes the L1 norm between these points after applying the relative transformation $\actions$ predicted by the policy and the relative transformation $\actions_*$ of the expert to the end effector.
The auxiliary loss is defined similarly:
\begin{equation}
    \vspace{-1mm}
    L_{\text{AUX}}(\goalvec_*, \goalvec) = L_{\text{POSE}}(\goalvec_*, \goalvec),
    \vspace{-1mm}
    \label{eq:loss_aux}
\end{equation}
where $\goalvec$ is an additional output of the policy that predicts the pre-grasp pose and $\goalvec_*$ indicates the pre-grasp pose of the expert.  

\vspace{-3mm}
\paragraph{Critic Loss} The critic loss is defined as:
\begin{equation}
    \vspace{-1mm}
L(\phi) =  L_{\text{BE}}+L_{\text{AUX}},
\label{eq:loss_critic}
    \vspace{-1mm}
\end{equation}  
where $L_{\text{AUX}}$ is identical to \Eq{loss_aux} and $L_{\text{BE}}$ is the Bellman equation defined in Eq.\ 1 of the main paper.

\vspace{-3mm}
\paragraph{Grasp Prediction Loss} The loss for the grasp prediction network is defined as:
\begin{equation}
    \vspace{-1mm}
L(\zeta) =  L_{\text{CE}}(\grasppred_{\zeta}(\psi(\pcvec)), \mathbf{y}), 
\label{eq:loss_critic}
    \vspace{-1mm}
\end{equation}  
where $L_{\text{CE}}$ is a binary cross-entropy loss between the output predictions of the model $\grasppred_{\zeta}(\psi(\pcvec))$ and the binary labels $\mathbf{y}$. The labels indicate whether a pre-grasp pose will lead to a successful grasp or not.

\subsection{Training Details}

\paragraph{Training Techniques} We apply a variety of different techniques to make the policies more robust. To this end, during the sequential phase, we alternate between initializing the robot in a home position and random poses (that have the object and hand in its view). As proposed in \cite{wang:corl2021b}, we occasionally compute optimal actions using DAGGER \cite{ross:2011dagger} during RL exploration and use them to supervise the policy's actions. Additionally, we start episodes of the RL agent with a random amount of initial actions proposed by the expert to further guide the training process. 
In the finetuning phase, we drop most of these techniques and start all the episodes from home position. We do not use DAGGER anymore and rollouts from the RL agent are not started with actions proposed by the expert.

\vspace{-3mm}
\paragraph{Expert Demonstrations}
In the pretraining stage, we use motion and grasp planning \cite{wang:rss2020}. We plan trajectories until the pre-grasp pose and then use the forward grasping motion  (cf. main paper Sec.\ 4.1) to grasp the object.
The ACRONYM dataset \cite{eppner:acronym2020} is a large dataset for robot grasp planning based on physics simulation. It is used for grasp selection of the expert planner. However, because it does not consider the human hand, we first prefilter suitable grasps offline. To this end, we parse grasps from ACRONYM and combine them with the handover poses extracted from DexYCB \cite{chao:cvpr2021}, i.e., the pose at the last frame of the sequences where the hand and object are not moving. We check for collisions between the hand and the gripper using a mesh collider. We then filter out sequences where the robot and hand collide. 
During rollouts of expert trajectories, we frequently add random perturbations to the robot end-effector and replan the trajectory from the current pose. 

\vspace{-3mm}
\paragraph{Hindsight goals} In the pretraining stage, the expert planner \cite{wang:rss2020} provides goal labels for both expert demonstrations and RL rollouts. In the finetuning stage, we cannot rely on the goal selection from the planner anymore. We therefore employ the hindsight scheme from \cite{wang:corl2021b}, which was used for training with novel objects, and utilize it for labeling sequences where the human and robot move simultaneously. In particular, if an episode is successful, we can use the pre-grasp pose from this episode as a label for supervision of the goal-prediction task.

\vspace{-3mm}
\paragraph{Grasp Prediction Data Collection} To collect samples for training the grasp prediction network, we utilize pre-grasps generated from ACRONYM \cite{eppner:acronym2020}. We initialize the hand and object in the final pose of the handover trajectories from HandoverSim \cite{chao:icra2022}. We then use inverse kinematics to compute a robot pose that matches the end-effector pre-grasp pose. We then rollout the forward grasping motion (cf. main paper Sec.\ 4.1) and check for grasp stability. The label is 1 if the grasp is successful, and 0 otherwise (e.g., hand collision or object drop). For each pre-grasp pose, we collect two more samples by adding small random noise (translation and rotation) to the end-effector pose. This is done to balance the positive and negative labels in the dataset \cite{mousavian:iccv2019}, since the pre-grasp poses from ACRONYM will have a higher chance of leading to a stable grasp.
\section{Implementation Details}
\label{app_impl}

\begin{table}[t]
\centering
\resizebox{0.85\columnwidth}{!}{
\begin{tabular}{ll}
\textbf{Training Parameters} & \textbf{Value} \vspace{0.1cm}\\
\hline\\
Num. iterations pretraining & 10000 \\
Num. iterations finetuning & 5000 \\
Buffer size pretraining & 1e6\\
Buffer size finetuning & 4e5\\
Parallel workers & 3 \\

Simulation timestep & 1e-3s \\
Simulation steps per action & 150 \\

Network layers & 3 \\
Hidden size & 256 \\
Activation functions & ReLU \\
Optimizer & Adam\cite{kingma:2015iclr}\\


\end{tabular}
}
\captionof{table}{Overview of the most important parameters.} 
\vspace{-3mm}
\label{tab:params}
\end{table}
\subsection{Network architecture}

We use PointNet++ \cite{qi:nips2017} as backbone for the grasp prediction network, the actor network and the critic network. We use separate backbones for each network. The networks are fully-connected MLPs with three-layers and 256 neurons per layer. The critic network has two heads as output, one for predicting the Q-value and one for the auxiliary goal prediction. Similarly, the actor has one head for the goal prediction and one for the action prediction. The critic network takes as input the concatenation of a point cloud and an action (see Fig.\ 3 in the main paper) . During critic training, transitions from the replay buffer are used as actions and point clouds. On the other hand, during actor training, the actor's action predictions are used together with the point cloud from the replay buffer.
The output of the grasp prediction network is a scalar that indicates a probability. We use Adam \cite{kingma:2015iclr} as optimizer and ReLU activation functions during training of all networks. The learning rate is decreased over the course of training.

\subsection{Training Information}
We use TD3 \cite{fujimoto:icml2018} as our learning algorithm. To ensure a fair comparison with GA-DDPG, we use their implementation of TD3 and only make minimal changes to learning parameters. We therefore refer the reader to \cite{wang:corl2021b} for an exact description of all the parameters used and report only crucial or changed parameters in \Tab{params}. In every iteration, we use three parallel workers to rollout episodes. Thereafter, we update our networks using 20 optimization steps. We run the pretraining for 10k iterations and the finetuning for 5k iterations. The grasp prediction network is trained for 1k iterations. 

We train our method on a single Nvidia V100 32GB. Training the full method takes around 72-96 hours, about 36-48 hours for pretraining and 36-48 hours for finetuning. 
The grasp prediction network is trained offline and training takes roughly 1 hour.

\section{Additional Simulation Evaluation}
\label{app_results}

\paragraph{HandoverSim Benchmark} For completeness, we report the results of the remaining settings from HandoverSim \cite{chao:icra2022}. Namely, we add the settings ``S1 - Unseen Subjects'' in \Tab{s1}, ``S2 - Unseen Handedness'' in \Tab{s2}, and ``S3 - Unseen Objects'' in \Tab{s3}. Overall, we observe that the results are consistent with the main paper. In general, the main baseline GA-DDPG \cite{wang:corl2021b} struggles in the simultaneous setting. Our method has significantly better performance in terms of overall success rates, while retaining a slightly slower mean accumulated time for successful handovers. This is because GA-DDPG often goes for a grasp in the most direct path, whereas our approach searches for a safe pre-grasp pose, from where the object can be grasped. For a qualitative demonstration of this behavior, we refer to the supplementary video. We also compare our final model with the pretrained versions (\emph{Ours w/o ft.}). The results further indicate that finetuning helps improve the model, especially in the simultaneous setting, e.g., the success rate in \Tab{s1} improves from 62.78\% to 73.33\% with the finetuned model.

Notably, results on S2 and S3 suggest that our method can generalize well to unseen subjects and unseen objects. This result is important because in unstructured real world environments, neither objects nor subjects have been encountered during training.

\begin{figure}[]
 \centering
 \includegraphics[width=1.05\linewidth, trim={0cm 0cm 0cm 2cm}, clip]{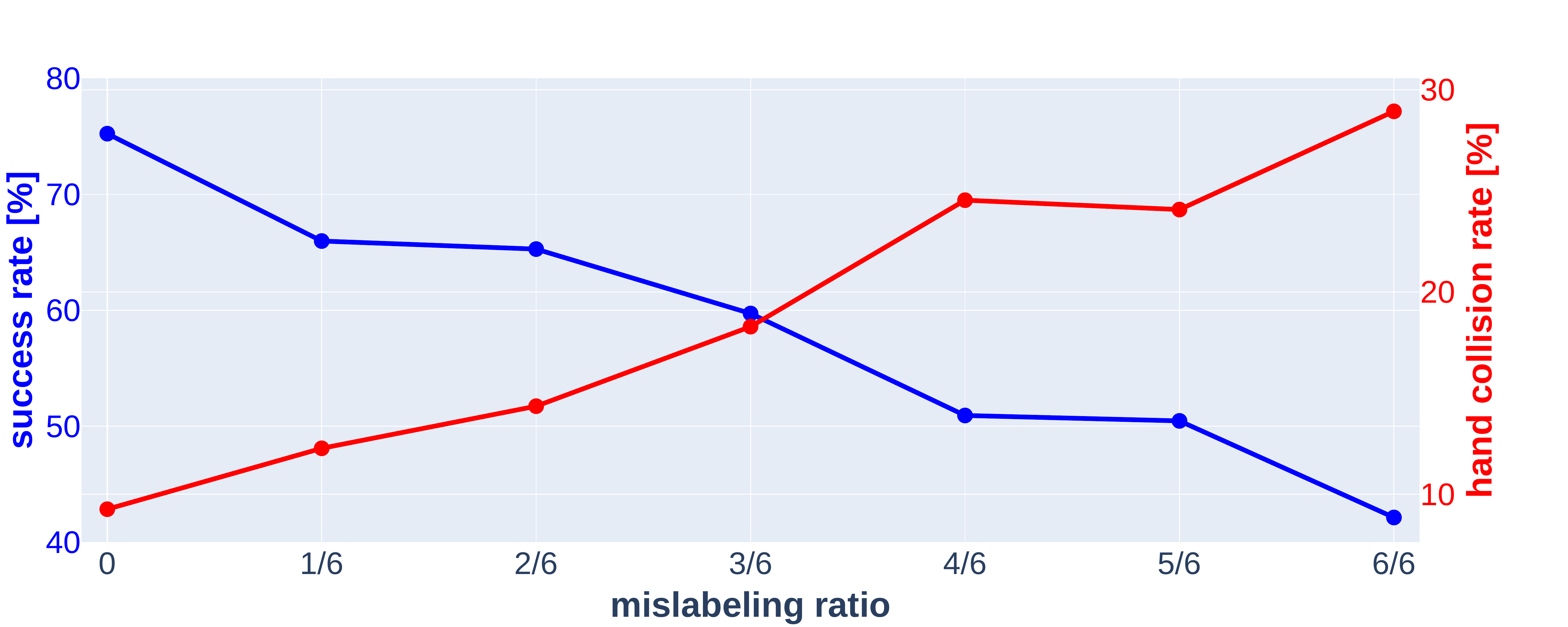}
     \caption{\textbf{Segmentation Mislabeling.} Our method's success (\textcolor{blue}{blue}) and hand collision rate (\textcolor{red}{red}) under increasing degree of mislabeling hand as object segments.}
 \label{fig:seg_err_plot}
\end{figure}

\paragraph{Robustness Analysis} \update{We evaluate in simulation how \textit{noisy observations} affect our pipeline by (1) adding simulated Kinect noise to depth images~\cite{handa:2014icra} and (2) testing with imperfect hand segmentation. For (2), we divide the hand into 6 different parts (fingers and palm) and re-label a subset of parts as object in the segmentation mask. We vary the mislabeling ratio (``0/6'' no parts and ``6/6'' all parts) and sample randomly which parts will be mislabeled for a given episode. As expected, performance degrades with increasing noise in depth (\eg, a 59.49\% success rate in \Tab{kinect}) and increasing mislabeling ratio (\eg, decreasing success rate and increasing hand collisions in \Fig{seg_err_plot}).}

\begin{table}[t]
 \vspace{-0mm}
 \centering
 \small
  \resizebox{1.0\columnwidth}{!}{%
 \begin{tabular}{l|cccc}
  \hline
  & \multirow{2}{*}{success (\%)} &  \multicolumn{3}{c}{failure (\%)} \\
  & & contact & drop & timeout \\
  \hline

   w/ Kinect noise & 59.49 & 13.19 &  19.68 &  ~~7.64         \\ 

  \rowcolor{Gray}
  Ours & \textbf{75.23} & ~~\textbf{9.26} & \textbf{13.43} & ~~\textbf{2.08} \\ 
\hline
 \end{tabular}
 }
 \vspace{-2mm}
 \caption{\update{We analyze the effect of simulated Kinect noise~\cite{handa:2014icra} on our model.} Results are averaged over 3 random seeds.
 }
 \vspace{-0mm}
 \label{tab:kinect}
\end{table}

\begin{table*}[t!]
 \centering
 \begin{tabular}{l|l|cccc|cccc}

  \hline
  \multicolumn{10}{c}{S1: Unseen Subjects} \\
  \hline
  & & \multirow{2}{*}{success (\%)} & \multicolumn{3}{c|}{mean accum time (s)} & \multicolumn{4}{c}{failure (\%)} \\
  & & & exec & plan & total & contact & drop & timeout & total \\
  \hline
  \parbox[t]{2mm}{\multirow{5}{*}{\rotatebox[origin=c]{90}{Sequential}}}
  & OMG Planner~\cite{wang:rss2020} $\dagger$  & 62.78  & 8.012 & 1.355 & 9.366 & 33.33 & ~~2.22 & ~~1.67 & 37.22   \\
  & \gc Yang et al.~\cite{yang:icra2021} $\dagger$  & \gc 62.78 & \gc 4.719  & \gc 0.039
 & \gc 4.758 &  \gc 14.44 &  \gc ~~7.78  & \gc 15.00 & \gc 37.22  \\  
\cline{2-10}
  & GA-DDPG~\cite{wang:corl2021b}   & 55.00 & \textbf{6.791} & \textbf{0.136} & \textbf{6.927} & ~~8.89 & 15.00 & 21.11 & 45.00 \\
  & \gc Ours w/o ft. & \gc 68.15  & \gc 7.151 & \gc 0.164
 & \gc 7.314          &  \gc    ~~6.85     &  \gc 12.96          & \gc 12.04 & \gc    31.85     \\  
  & Ours    & \textbf{75.00} & 7.108 & 0.159 & 7.267 & ~~\textbf{5.00} & \textbf{12.59} & ~~\textbf{7.41}  & \textbf{25.00}    \\
\hline

  \parbox[t]{2mm}{\multirow{3}{*}{\rotatebox[origin=c]{90}{Simult.}}}
  
  & GA-DDPG~\cite{wang:corl2021b} & 33.33 & \textbf{4.261} & \textbf{0.132} & \textbf{4.393} & 15.56 & 21.67 & 29.44 & 66.67   \\
    & \gc Ours w/o ft. & \gc 62.78 & \gc 5.695  & \gc
 0.164 & \gc 5.859   &  \gc     ~~5.93    &  \gc 17.59         & \gc 13.70 & \gc 37.22        \\  
  & Ours   & \textbf{73.33} & 5.633 & 0.158 & 5.791 & ~~\textbf{5.56} & \textbf{15.37} & ~~\textbf{5.74} & \textbf{26.67}  \\
  
\bottomrule
  
 \end{tabular}
 \caption{\textbf{Unseen Subjects} Comparison of our method against various baselines from the HandoverSim benchmark \cite{chao:icra2022} in the setting ``S1: Unseen Subjects''. The results of our method are averaged over 3 random seeds. $\dagger$: both methods \cite{wang:rss2020,yang:icra2021} are evaluated with ground-truth states in \cite{chao:icra2022} and thus are not directly comparable with ours.}
 \label{tab:s1}
\end{table*}
\begin{table*}[t!]
 \centering
 \begin{tabular}{l|l|cccc|cccc}

  \hline
  \multicolumn{10}{c}{S2: Unseen Handedness} \\
  \hline
  & & \multirow{2}{*}{success (\%)} & \multicolumn{3}{c|}{mean accum time (s)} & \multicolumn{4}{c}{failure (\%)} \\
  & & & exec & plan & total & contact & drop & timeout & total \\
  \hline
  \parbox[t]{2mm}{\multirow{5}{*}{\rotatebox[origin=c]{90}{Sequential}}}
  & OMG Planner~\cite{wang:rss2020} $\dagger$  & 62.78  & 8.275 & 1.481 & 9.755 & 30.56 & ~~3.89 & ~~2.78 & 37.22  \\
  & \gc Yang et al.~\cite{yang:icra2021} $\dagger$  & \gc 62.50 & \gc 4.808  & \gc 0.034
 & \gc 4.843 &  \gc 16.11  &  \gc  10.56   & \gc 10.83 & \gc 37.50 \\  
\cline{2-10}
  & GA-DDPG~\cite{wang:corl2021b}   & 55.00 & 7.145 & \textbf{0.129} & 7.274 & ~~\textbf{8.61} & 17.78 & 18.61 & 45.00  \\
  & \gc Ours w/o ft. & \gc 71.76  & \gc \textbf{7.045} & \gc 0.140
 & \gc   \textbf{7.185}        &  \gc  ~~8.80  &  \gc 14.72  & \gc ~~4.72 & \gc  28.24       \\  
  & Ours   & \textbf{72.96} & 7.101 & 0.144 & 7.245 & 11.29 & \textbf{12.69} & ~~\textbf{3.05}  & \textbf{27.04}    \\
\hline

  \parbox[t]{2mm}{\multirow{3}{*}{\rotatebox[origin=c]{90}{Simult.}}}
  
  & GA-DDPG~\cite{wang:corl2021b} & 28.33 & \textbf{4.747} & \textbf{0.133} & \textbf{4.881} & ~~9.17 & 34.44 & 28.06  & 71.67   \\
    & \gc Ours w/o ft. & \gc 64.81  & \gc 5.638  & \gc 0.144
 & \gc  5.783     &  \gc ~~\textbf{8.24} &  \gc 21.02  & \gc ~~5.93 & \gc 35.19        \\  
  & Ours  & \textbf{71.11} & 5.771 & 0.150 & 5.921 & 10.00 & \textbf{15.37} & ~~\textbf{3.61} & \textbf{28.89} \\
  
\bottomrule
  
 \end{tabular}
 \caption{\textbf{Unseen Handedness.} Comparison of our method against various baselines from the HandoverSim benchmark \cite{chao:icra2022} in the setting ``S2: Unseen Handedness''. The results of our method are averaged over 3 random seeds. $\dagger$: both methods \cite{wang:rss2020,yang:icra2021} are evaluated with ground-truth states in \cite{chao:icra2022} and are not directly comparable with ours.}
 \label{tab:s2}
\end{table*}
\begin{table*}[t!]
 \centering
 \begin{tabular}{l|l|cccc|cccc}

  \hline
  \multicolumn{10}{c}{S3: Unseen Objects} \\
  \hline
  & & \multirow{2}{*}{success (\%)} & \multicolumn{3}{c|}{mean accum time (s)} & \multicolumn{4}{c}{failure (\%)} \\
  & & & exec & plan & total & contact & drop & timeout & total \\
  \hline
  \parbox[t]{2mm}{\multirow{5}{*}{\rotatebox[origin=c]{90}{Sequential}}}
  & OMG Planner~\cite{wang:rss2020} $\dagger$ & 69.00 & 8.478 & 1.588 & 10.066 & 23.00 & ~~4.00 & ~~4.00 & 31.00   \\
  & \gc Yang et al.~\cite{yang:icra2021} $\dagger$  & \gc 62.00  & \gc 4.805 & \gc 0.031
 & \gc ~~4.837 &  \gc 18.00 &  \gc ~~9.00  & \gc 11.00 & \gc  38.00 \\  
\cline{2-10}
  & GA-DDPG~\cite{wang:corl2021b}   & 50.00 & \textbf{7.305} & \textbf{0.135} & ~~\textbf{7.440} & ~~\textbf{5.00} & 23.00 & 22.00 & 50.00       \\
  & \gc Ours w/o ft. & \gc  76.33 & \gc 7.410 & \gc 0.151
 & \gc  ~~7.565     &  \gc    ~~9.33     &  \gc   10.67   & \gc ~~\textbf{3.67} & \gc   23.67      \\  
  & Ours    & \textbf{79.67} & 7.499 & 0.156 & ~~7.656 & ~~6.33 & \textbf{10.33}  & ~~\textbf{3.67}  & \textbf{20.33}    \\
\hline

  \parbox[t]{2mm}{\multirow{3}{*}{\rotatebox[origin=c]{90}{Simult.}}}
  
  & GA-DDPG~\cite{wang:corl2021b} & 33.00 & \textbf{4.948} & \textbf{0.123} & ~~\textbf{5.071} &  10.00 & 33.00 & 24.00  & 67.00  \\
    & \gc Ours w/o ft. & \gc 72.00  & \gc 6.242 & \gc 0.168
 & \gc ~~6.410  &  \gc ~~7.33 &  \gc  13.67 & \gc ~~7.00 & \gc 28.00       \\  
  & Ours   & \textbf{75.67} & 6.153 & 0.160 & ~~6.314 & ~~\textbf{5.00} & \textbf{13.33} & ~~\textbf{6.00} & \textbf{24.33}  \\
  
\bottomrule
  
 \end{tabular}
 \caption{\textbf{Unseen Object Evaluation.} Comparison of our method against baselines from the HandoverSim benchmark \cite{chao:icra2022} in the setting ``S3: Unseen Objects''. The results of our method are averaged over 3 random seeds. $\dagger$: both methods \cite{wang:rss2020,yang:icra2021} are evaluated with ground-truth states in \cite{chao:icra2022} and are not directly comparable with ours.}
 \label{tab:s3}
\end{table*}


\newpage
~
\newpage
~

\begin{figure*}[ht!]
 \centering
 \includegraphics[width=0.75\linewidth]{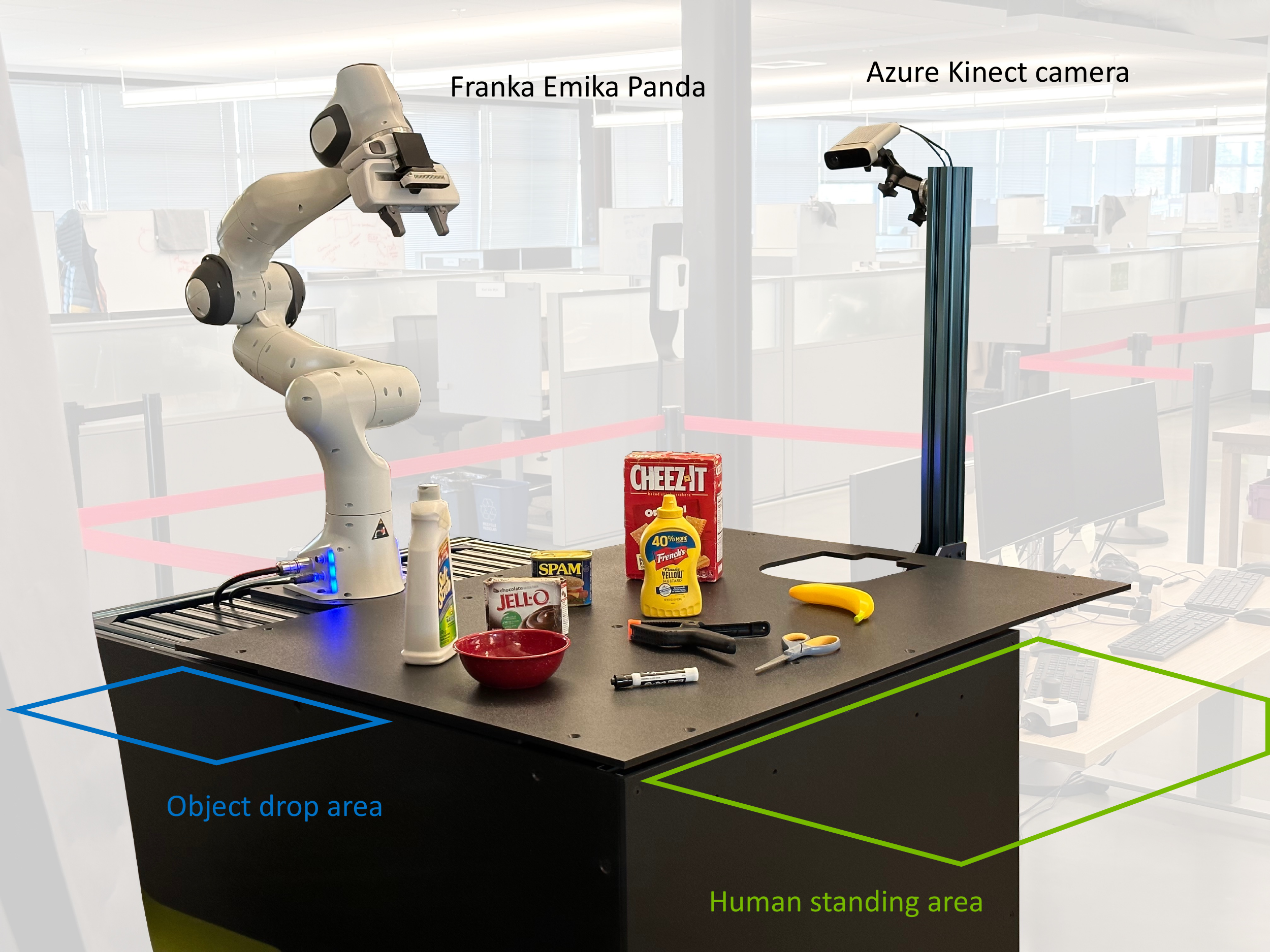}~
 \caption{\small The setup of our real-world handover system. A Franka Emika Panda robot and an Azure Kinect camera are rigidly mounted on a table. The human participant will stand across the table (in the green area), pick up objects, and attempt handovers to the robot. The robot will drop the object in a designated area (blue) after retrieving it from the handover.}
 \label{fig:setup}
\end{figure*}

\section{Sim-to-Real Transfer}
\label{app_sim2real}

\subsection{System Setup}
\label{sec:setup}

\cref{fig:setup} shows the setup of our real-world handover system. The setup consists of a Franka Emika Panda robot and a Azure Kinect camera, both rigidly mounted to a table. The Azure Kinect is mounted externally to the robot with the extrinsics calibrated, and is perceiving the scenes from a third-person view with an RGB-D stream. The objects for handover are initially placed on the table. During handovers, the human participant will stand on the opposite side of the table (in the green area), pick up the objects, and attempt handovers to the robot. If the robot successfully retrieves the object, it will move the end effector to a drop-off area (blue) and drop the object into a box.

Since our policy expects a segmented point cloud at the input, we follow the perception pipeline used in~\cite{yang:icra2021,wang:corl2021b} to generate segmented point clouds for the hand and object. The Azure Kinect is launched to provide a continuous stream of RGB images and point clouds. We first use Azure Kinect's Body Tracking SDK to track the 3D location of the wrist joint of the handover hand. At each time frame, we crop a sub-point cloud around the tracked joint location which includes points on both the hand and the held object. We additionally run a 2D hand segmentation model on the RGB image and use it to label the hand points in the cropped point cloud. We treat all the points not labeled as hand as the object. Since our policies are trained for wrist camera views and we use an external camera in the real-world system, we need to additionally compensate for the view point change. We transform the segmented point cloud from the external camera's frame to the wrist camera's frame using the calibrated robot-camera extrinsics and forward kinematics. This way we can simulate the segmented point cloud input which the policy observes during the training in simulation. Note that this perception pipeline can induce sim-to-real gaps from several sources: (1) noises in the point clouds from real cameras, (2) noises from body tracking and hand segmentation errors, (3) the change in view points (i.e., from the wrist to external camera), \update{and (4) unseen human behavior. To make our transfer policy more robust to diversity in human behvaior, we include human-object trajectories generated with D-Grasp \cite{christen2022dgrasp} during training.}

Compared to GA-DDPG~\cite{wang:corl2021b}, we adapt the control flow of the policy to explicitly incorporate the pre-grasp mechanism in our method. To control the motion of Franka, we follow the pipeline used in~\cite{yang:icra2021,wang:corl2021b}. Given a target end effector pose at a new time step, we use Riemannian Motion Policies (RMPs)~\cite{ratliff:arxiv2018} to generate a smooth trajectory for the robot arm. We use libfranka to control the Franka arm to follow the trajectory. The robot will start moving only when a segmented point cloud is perceived. Once it decides to grasp, we will execute a predefined motion where the robot closes the gripper, lifts the end effector, moves to the drop-off area, and opens the gripper. The robot will return to a standby pose and remain in that state if no segmented point cloud is perceived or after it drops off the object.

\begin{figure}[t!]
 \centering
 \includegraphics[width=\linewidth]{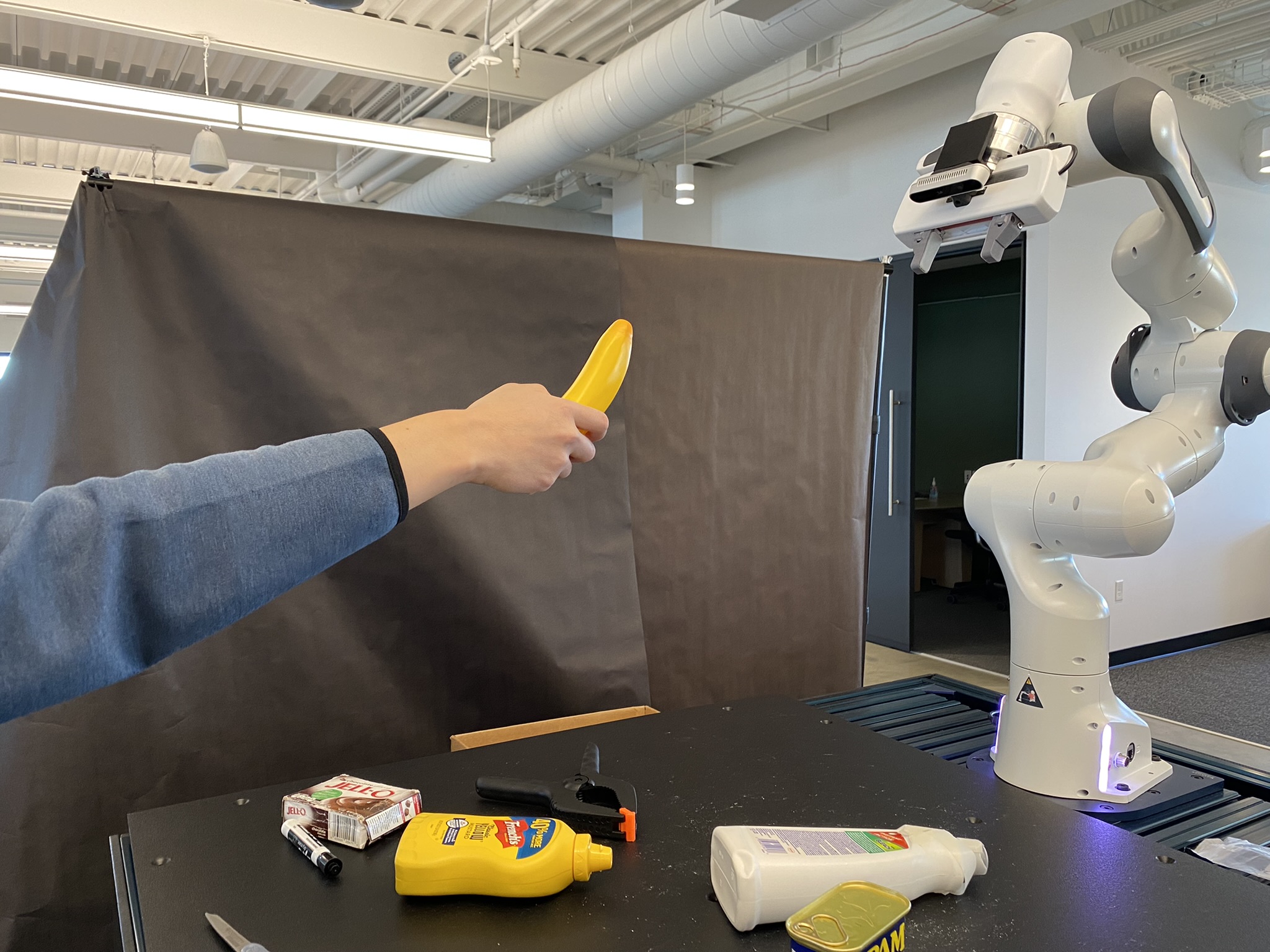}
 \caption{\small We conducted a pilot study by controlling the handover poses from the human subject.}
 \label{fig:pilot}
\end{figure}

\subsection{Pilot Study}

The goal of the pilot study is to provide a standardized benchmarking of the sim-to-real transfer. We instruct the participated subjects to follow a set of pre-determined handover poses when performing the handovers. We keep the instructed handover poses fixed for different methods to ensure a fair comparison.

\begin{figure*}[t!]
 \centering
 \begin{minipage}{0.495\linewidth}
  \centering
  011\_banana
  \\~\\ \vspace{-3mm}
  \begin{minipage}{0.320\linewidth}
   \centering
   \includegraphics[width=\linewidth,trim={270 264 420 170},clip]{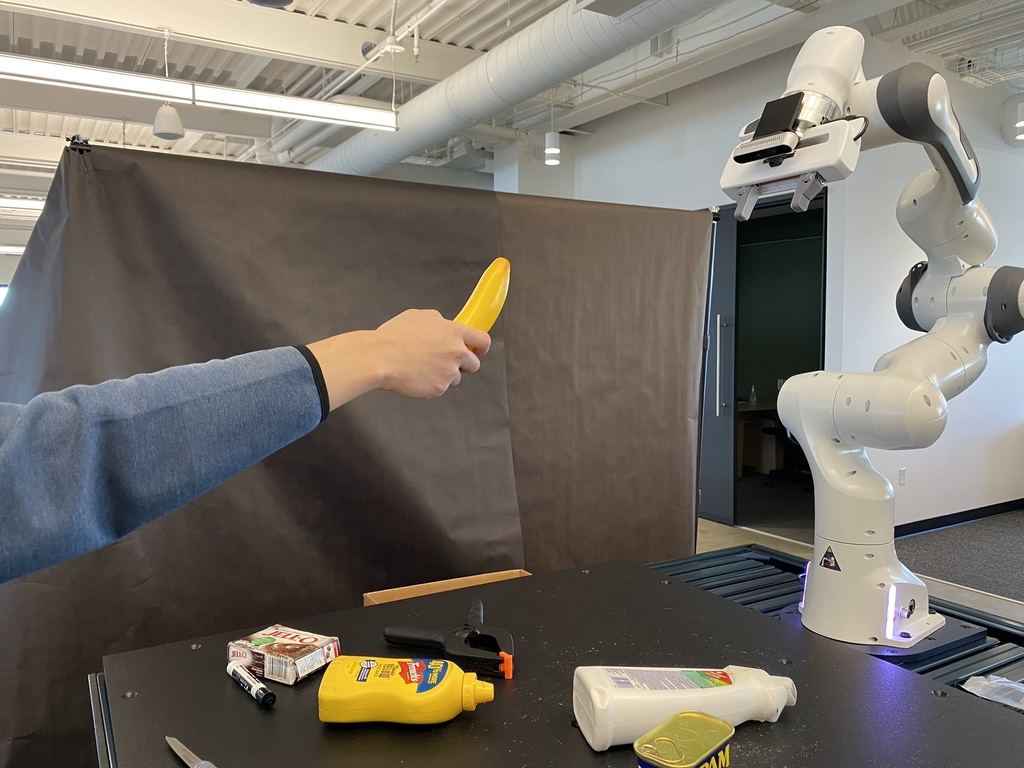}
   \\ \vspace{-1mm}
   pose 1
  \end{minipage}~
  \begin{minipage}{0.320\linewidth}
   \centering
   \includegraphics[width=\linewidth,trim={270 264 420 170},clip]{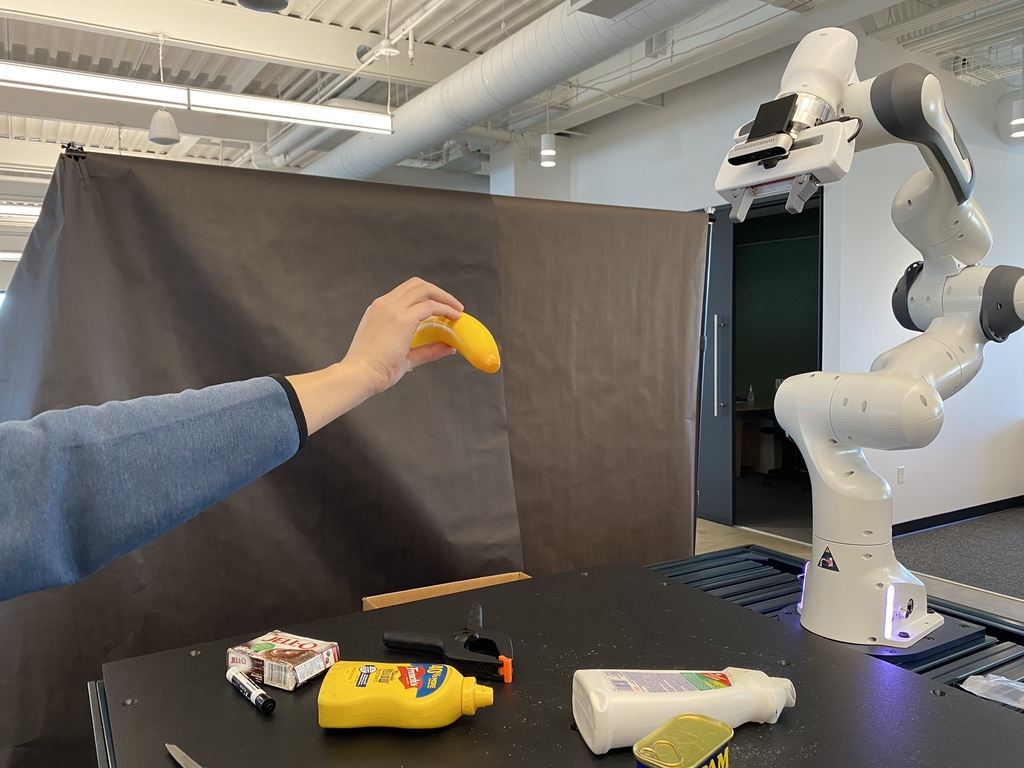}
   \\ \vspace{-1mm}
   pose 2
  \end{minipage}~
  \begin{minipage}{0.320\linewidth}
   \centering
   \includegraphics[width=\linewidth,trim={250 244 440 190},clip]{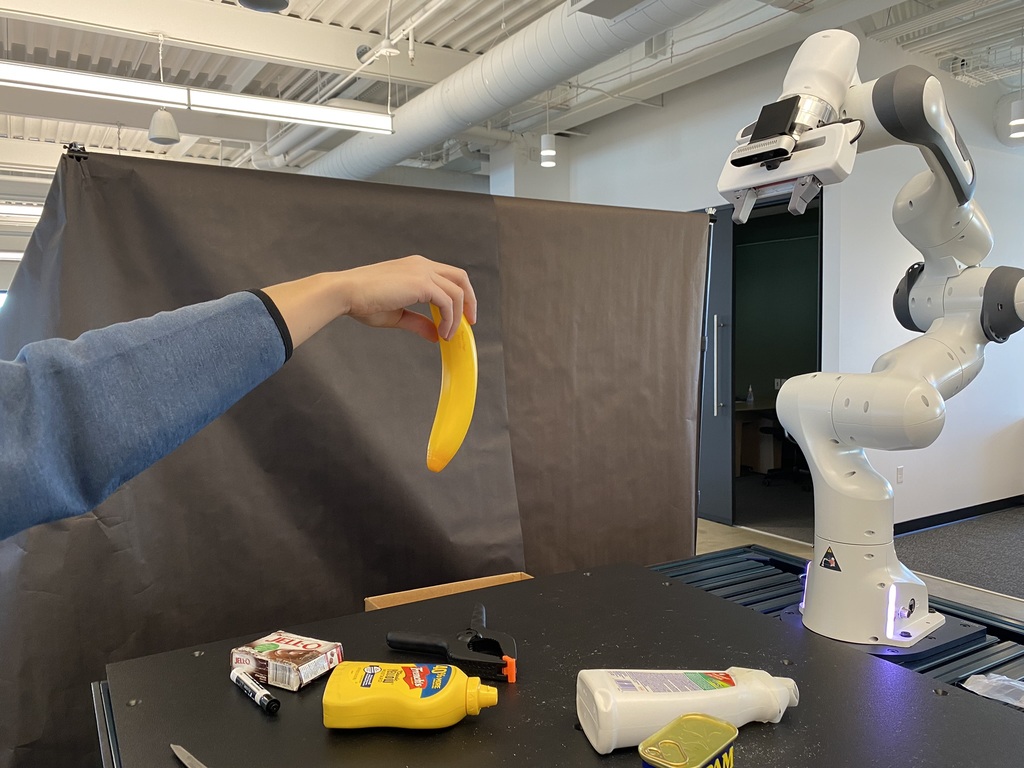}
   \\ \vspace{-1mm}
   pose 3
  \end{minipage}
  \\ \vspace{2mm}
 \end{minipage}~
 \begin{minipage}{0.495\linewidth}
  \centering
  037\_scissors
  \\~\\ \vspace{-3mm}
  \begin{minipage}{0.320\linewidth}
   \centering
   \includegraphics[width=\linewidth,trim={250 204 440 230},clip]{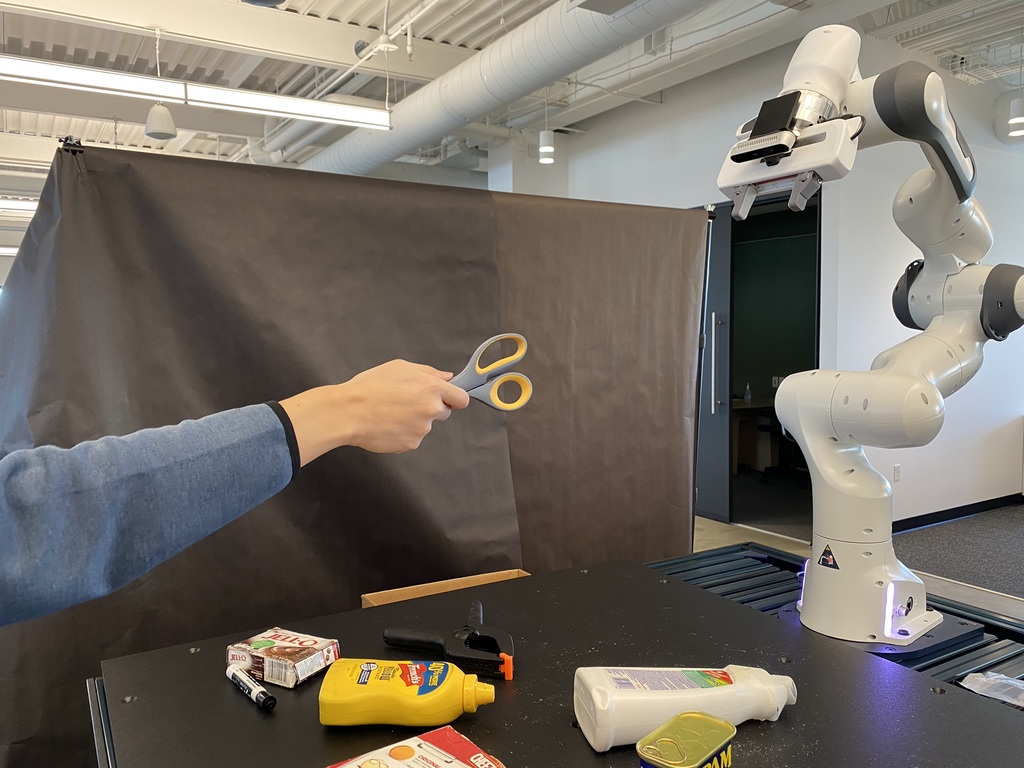}
   \\ \vspace{-1mm}
   pose 1
  \end{minipage}~
  \begin{minipage}{0.320\linewidth}
   \centering
   \includegraphics[width=\linewidth,trim={250 244 440 190},clip]{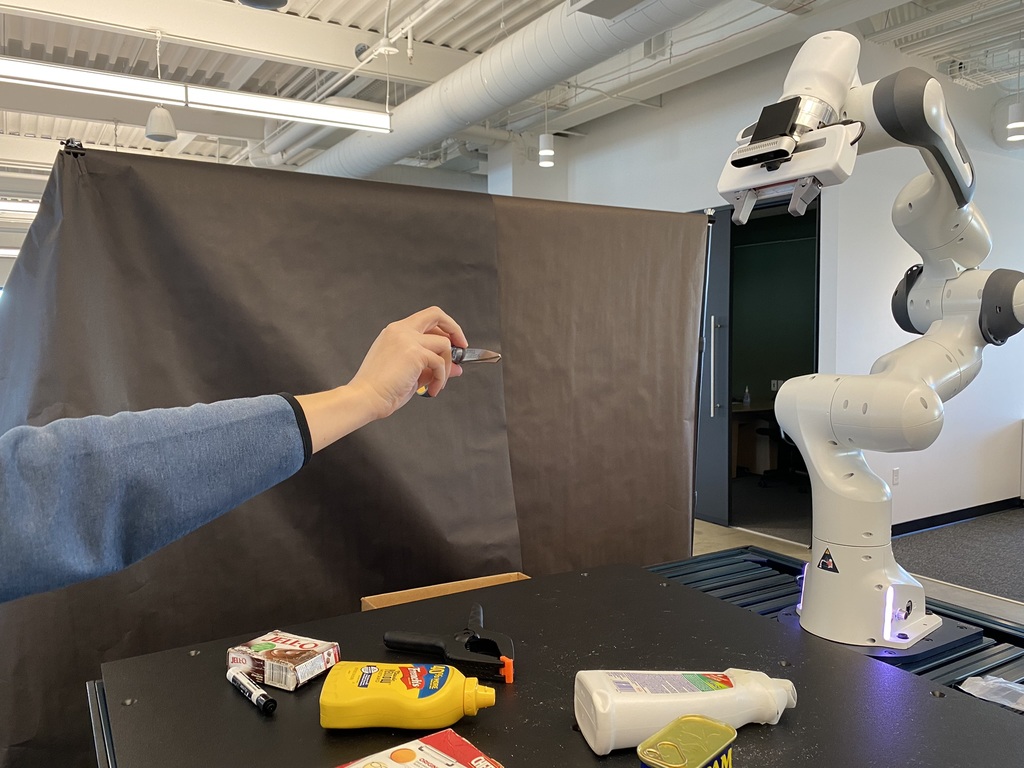}
   \\ \vspace{-1mm}
   pose 2
  \end{minipage}~
  \begin{minipage}{0.320\linewidth}
   \centering
   \includegraphics[width=\linewidth,trim={250 264 440 170},clip]{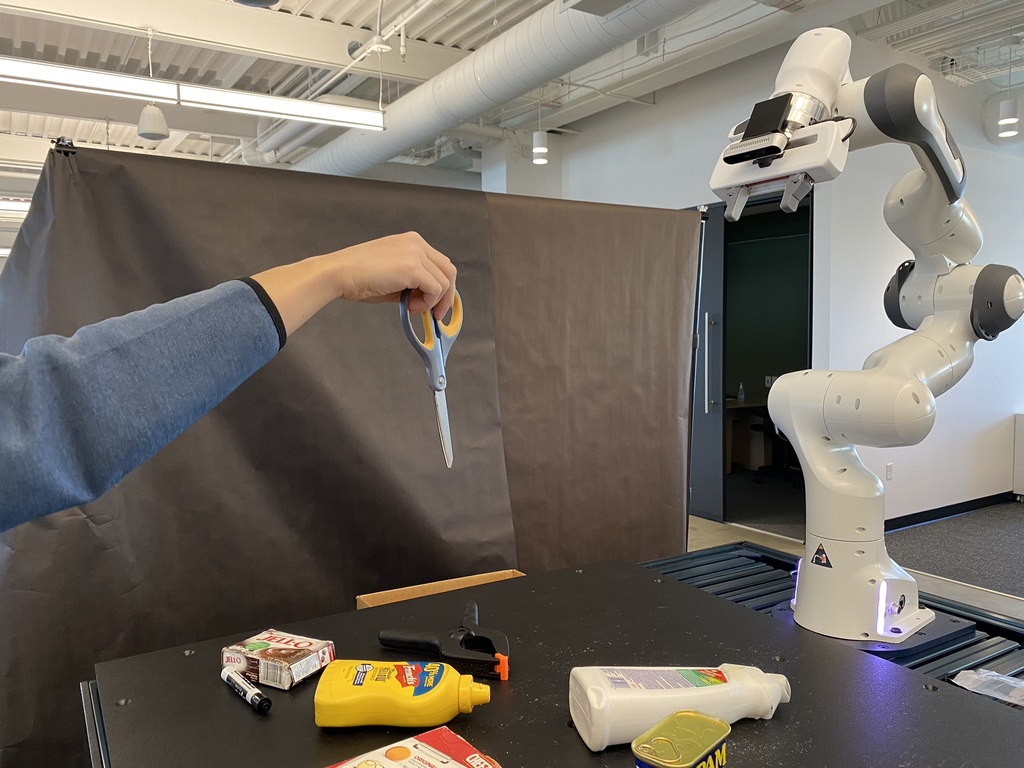}
   \\ \vspace{-1mm}
   pose 3
  \end{minipage}
  \\ \vspace{2mm}
 \end{minipage}
 \\ \vspace{3mm}
 \begin{minipage}{0.495\linewidth}
  \centering
  006\_mustard\_bottle
  \\~\\ \vspace{-3mm}
  \begin{minipage}{0.320\linewidth}
   \centering
   \includegraphics[width=\linewidth,trim={270 204 420 230},clip]{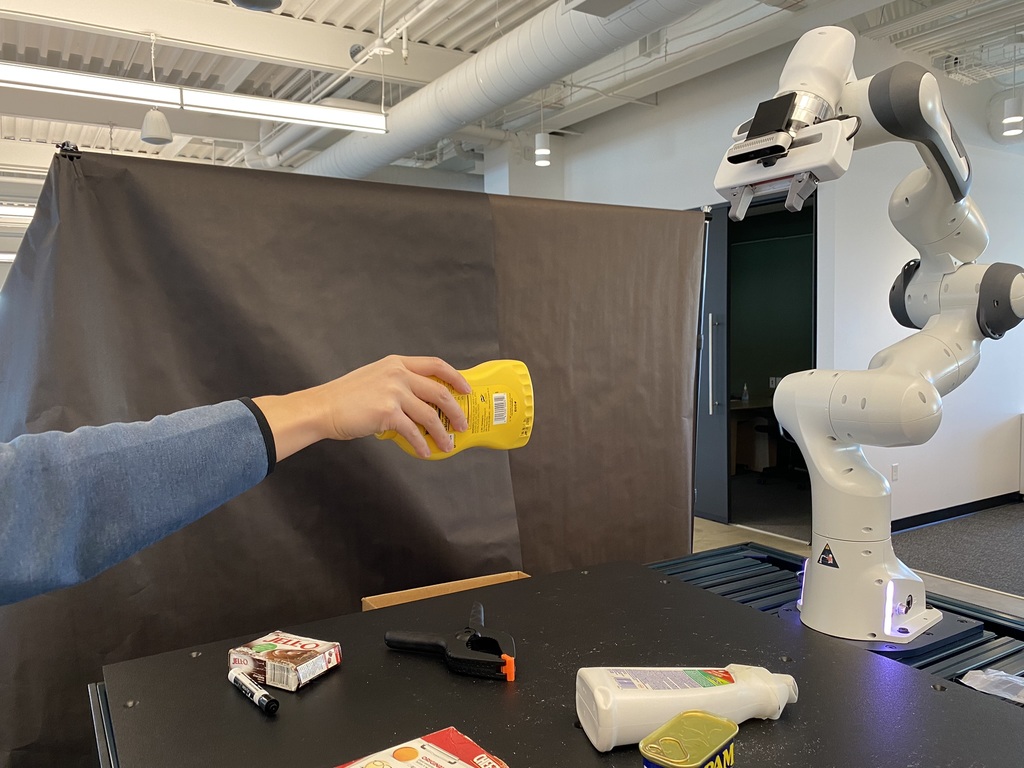}
   \\ \vspace{-1mm}
   pose 1
  \end{minipage}~
  \begin{minipage}{0.320\linewidth}
   \centering
   \includegraphics[width=\linewidth,trim={270 284 420 150},clip]{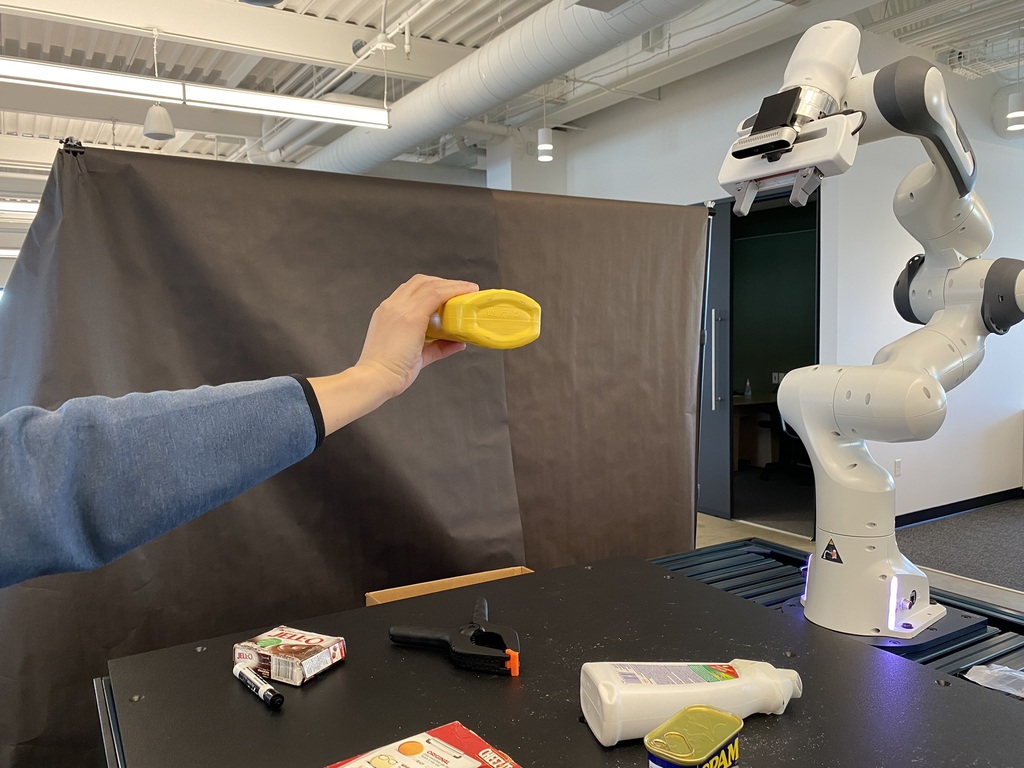}
   \\ \vspace{-1mm}
   pose 2
  \end{minipage}~
  \begin{minipage}{0.320\linewidth}
   \centering
   \includegraphics[width=\linewidth,trim={270 264 420 170},clip]{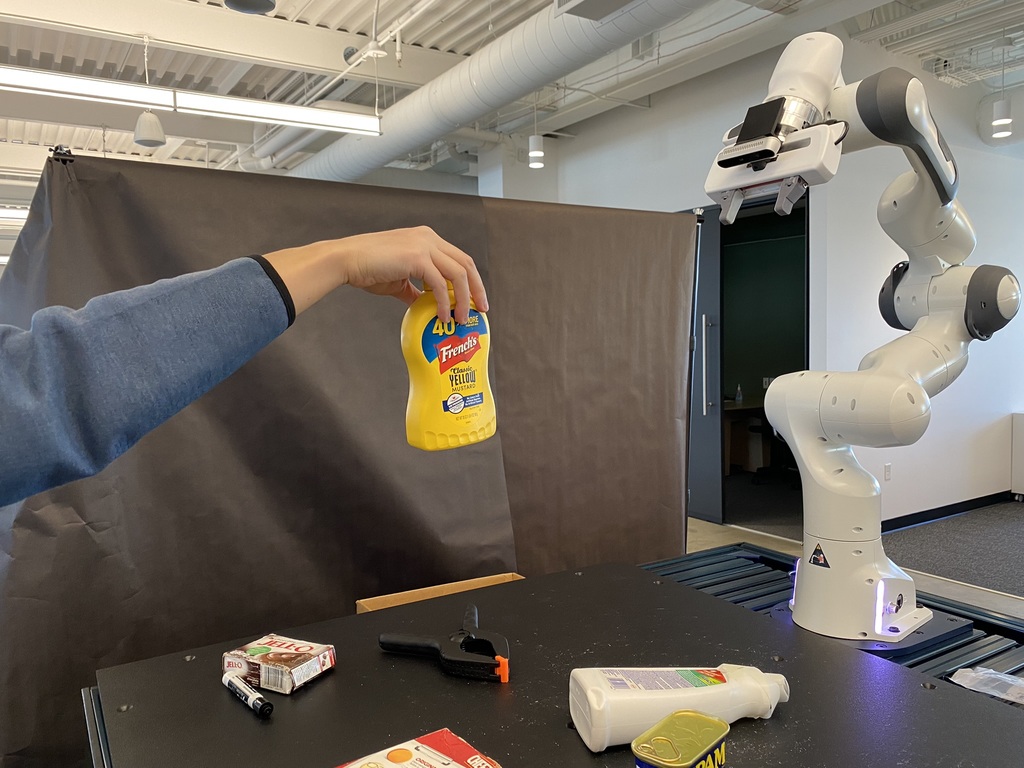}
   \\ \vspace{-1mm}
   pose 3
  \end{minipage}
  \\ \vspace{2mm}
 \end{minipage}~
 \begin{minipage}{0.495\linewidth}
  \centering
  024\_bowl
  \\~\\ \vspace{-3mm}
  \begin{minipage}{0.320\linewidth}
   \centering
   \includegraphics[width=\linewidth,trim={290 164 400 270},clip]{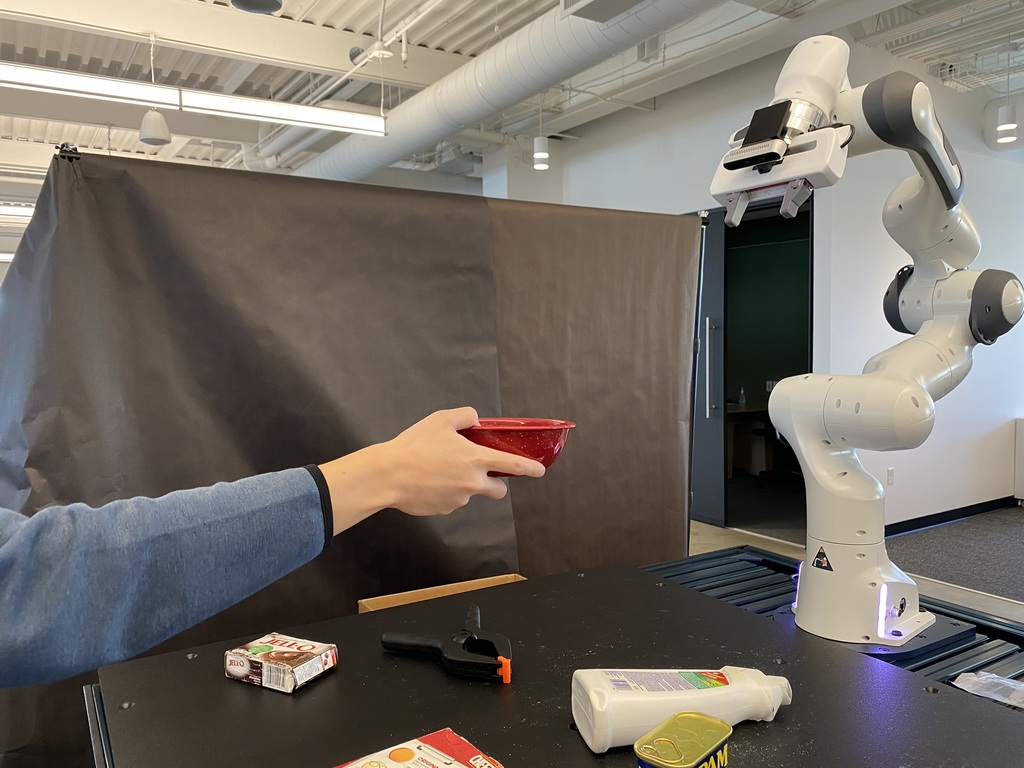}
   \\ \vspace{-1mm}
   pose 1
  \end{minipage}~
  \begin{minipage}{0.320\linewidth}
   \centering
   \includegraphics[width=\linewidth,trim={270 304 420 130},clip]{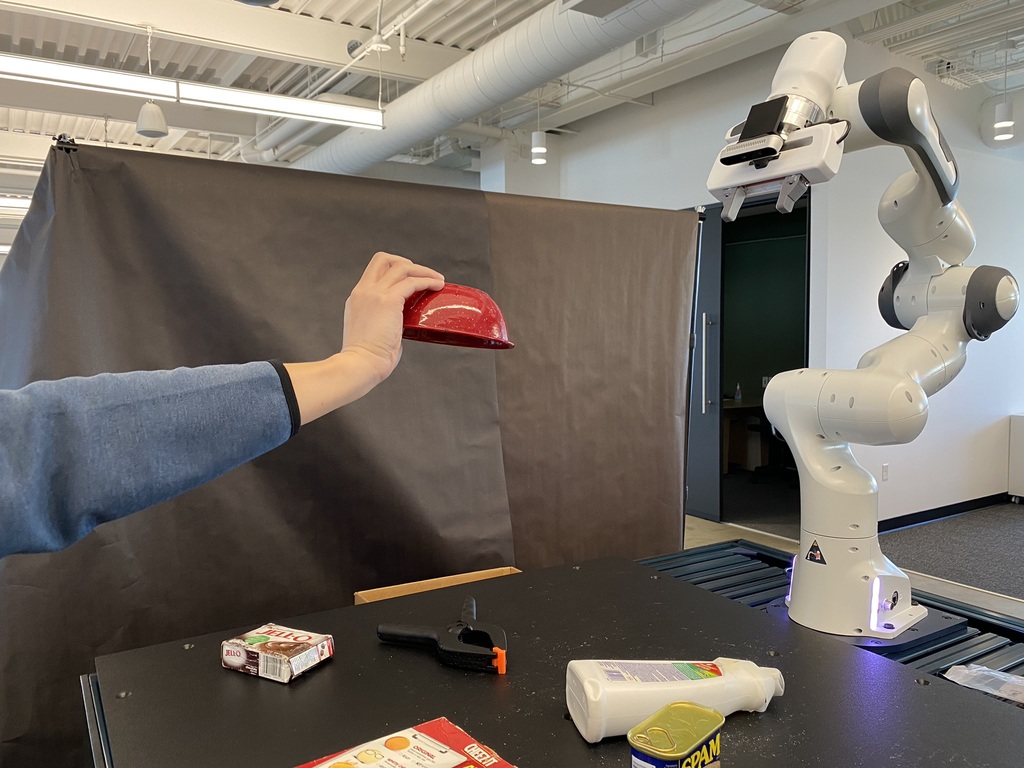}
   \\ \vspace{-1mm}
   pose 2
  \end{minipage}~
  \begin{minipage}{0.320\linewidth}
   \centering
   \includegraphics[width=\linewidth,trim={270 304 420 130},clip]{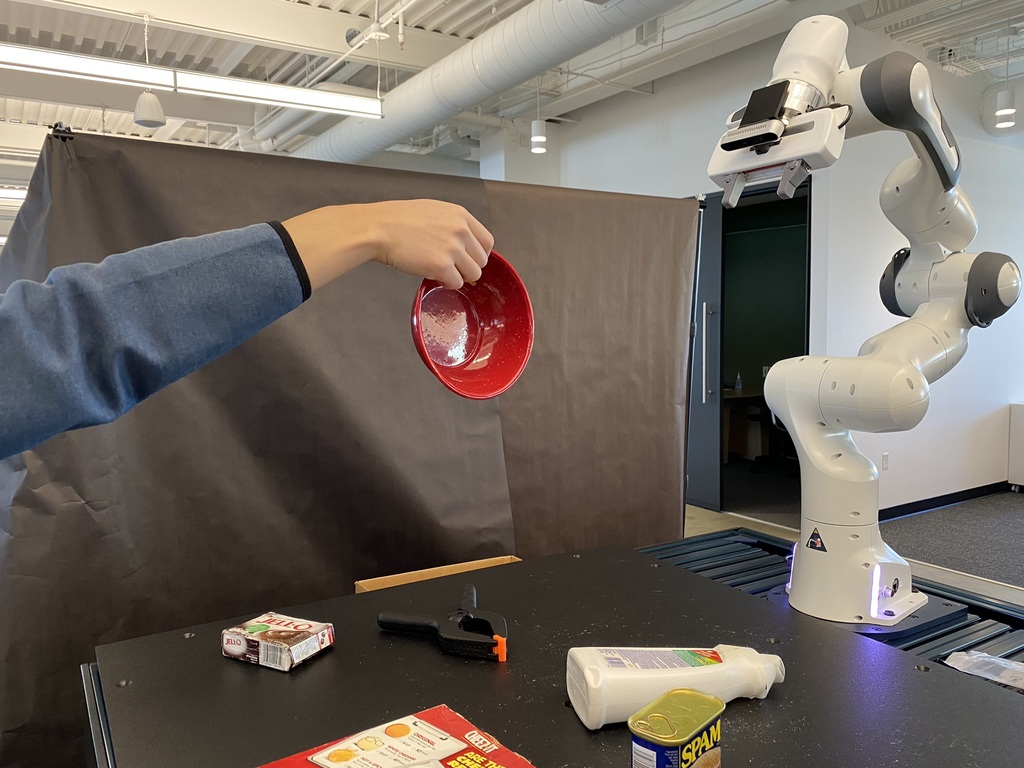}
   \\ \vspace{-1mm}
   pose 3
  \end{minipage}
  \\ \vspace{2mm}
 \end{minipage}
 \\ \vspace{3mm}
 \begin{minipage}{0.495\linewidth}
  \centering
  040\_large\_marker
  \\~\\ \vspace{-3mm}
  \begin{minipage}{0.320\linewidth}
   \centering
   \includegraphics[width=\linewidth,trim={270 224 420 210},clip]{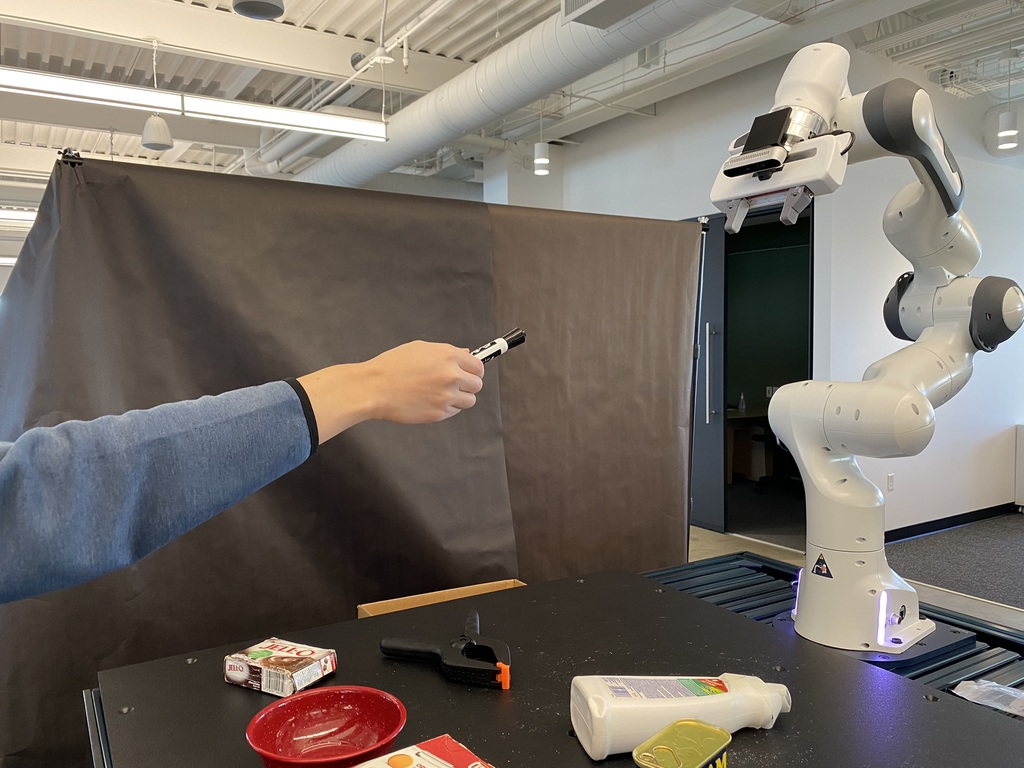}
   \\ \vspace{-1mm}
   pose 1
  \end{minipage}~
  \begin{minipage}{0.320\linewidth}
   \centering
   \includegraphics[width=\linewidth,trim={270 284 420 150},clip]{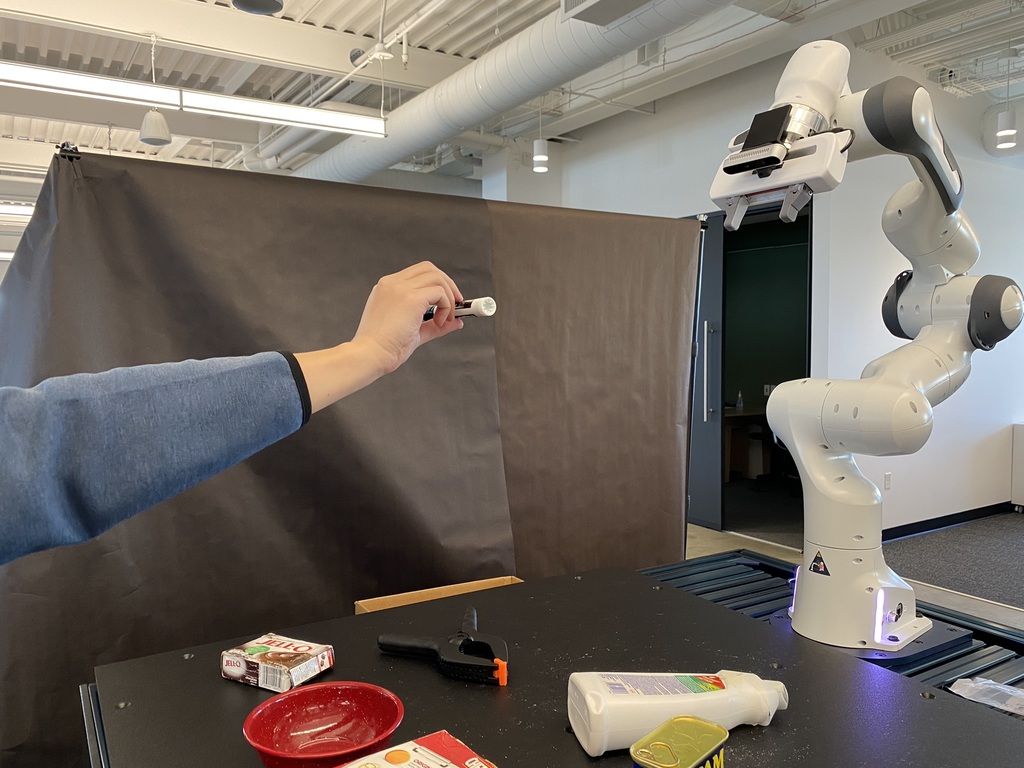}
   \\ \vspace{-1mm}
   pose 2
  \end{minipage}~
  \begin{minipage}{0.320\linewidth}
   \centering
   \includegraphics[width=\linewidth,trim={270 324 420 110},clip]{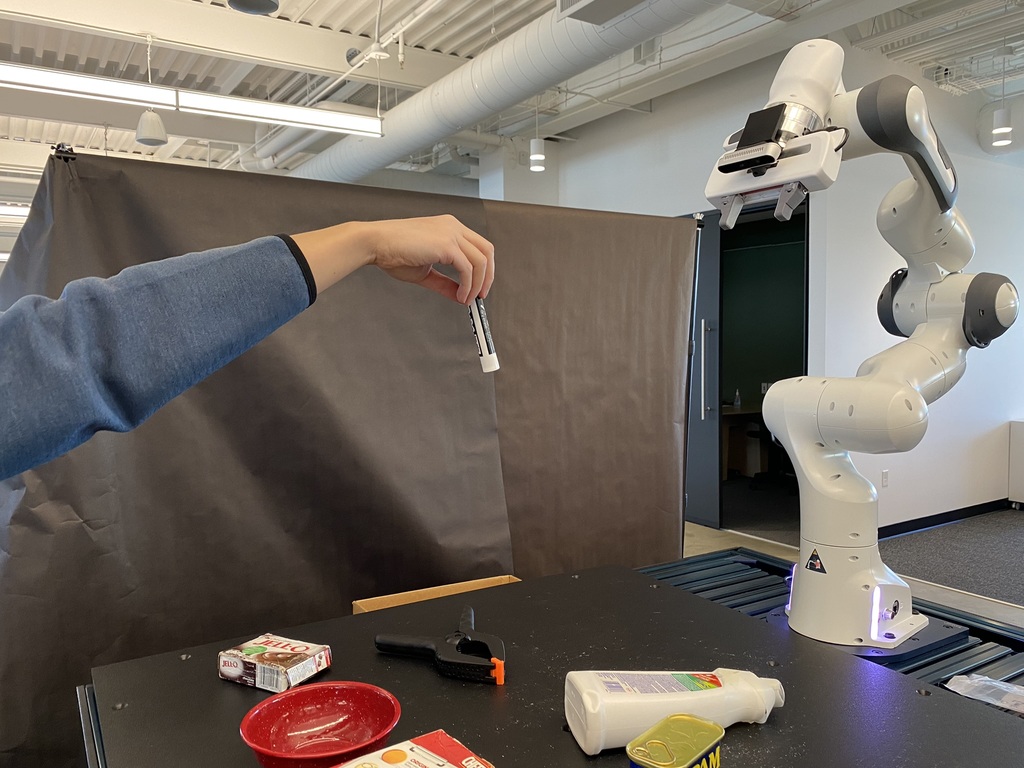}
   \\ \vspace{-1mm}
   pose 3
  \end{minipage}
  \\ \vspace{2mm}
 \end{minipage}~
 \begin{minipage}{0.495\linewidth}
  \centering
  003\_cracker\_box
  \\~\\ \vspace{-3mm}
  \begin{minipage}{0.320\linewidth}
   \centering
   \includegraphics[width=\linewidth,trim={270 264 420 170},clip]{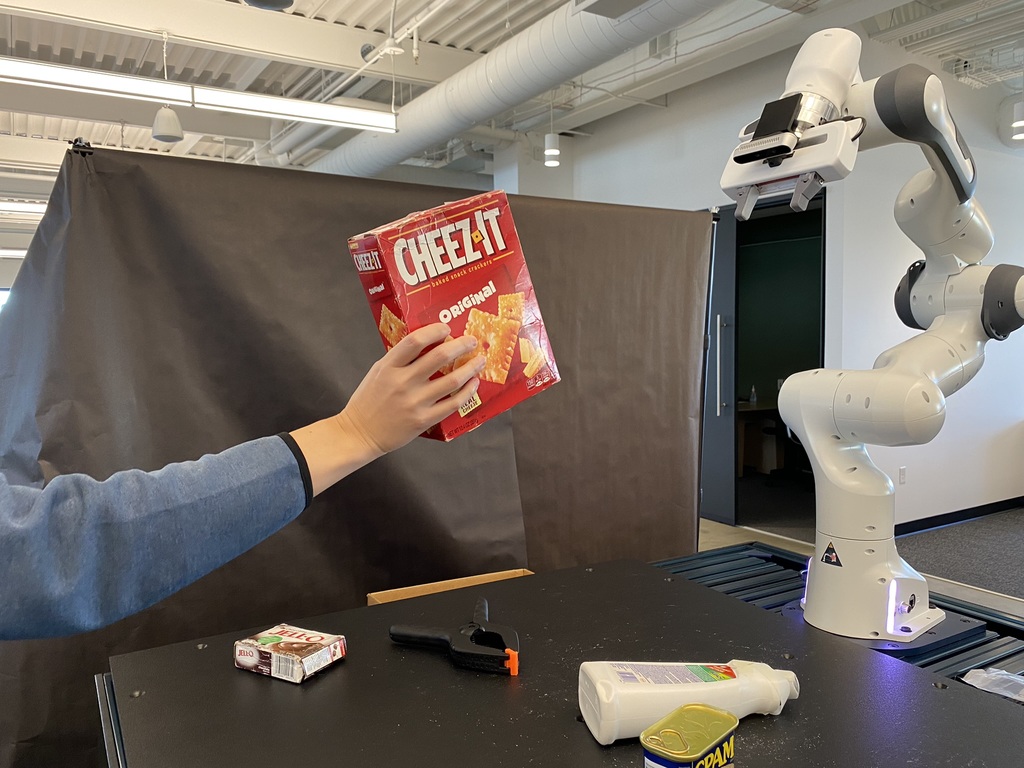}
   \\ \vspace{-1mm}
   pose 1
  \end{minipage}~
  \begin{minipage}{0.320\linewidth}
   \centering
   \includegraphics[width=\linewidth,trim={330 264 360 170},clip]{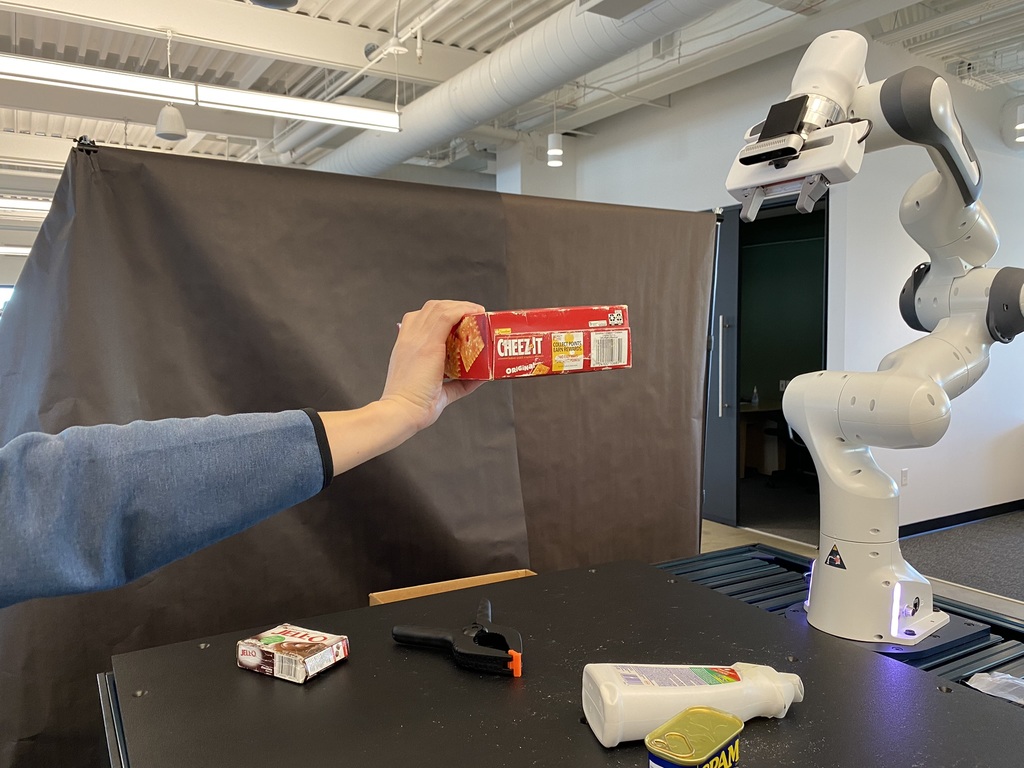}
   \\ \vspace{-1mm}
   pose 2
  \end{minipage}~
  \begin{minipage}{0.320\linewidth}
   \centering
   \includegraphics[width=\linewidth,trim={270 224 420 210},clip]{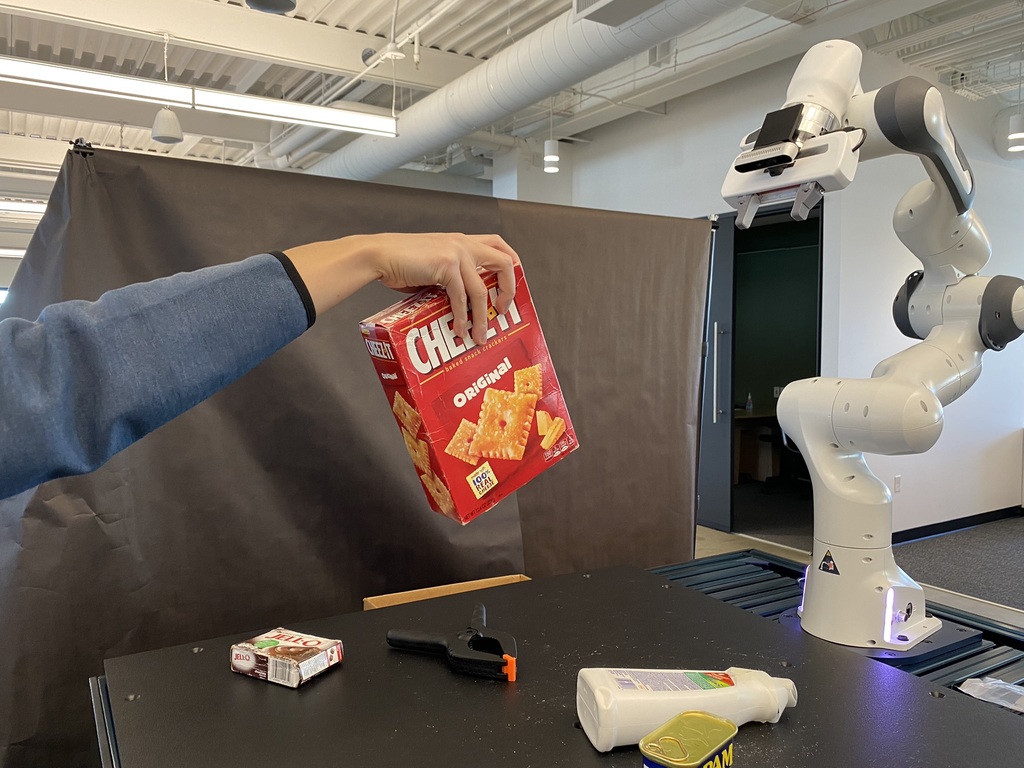}
   \\ \vspace{-1mm}
   pose 3
  \end{minipage}
  \\ \vspace{2mm}
 \end{minipage}
 \\ \vspace{3mm}
 \begin{minipage}{0.495\linewidth}
  \centering
  052\_extra\_large\_clamp
  \\~\\ \vspace{-3mm}
  \begin{minipage}{0.320\linewidth}
   \centering
   \includegraphics[width=\linewidth,trim={310 204 380 230},clip]{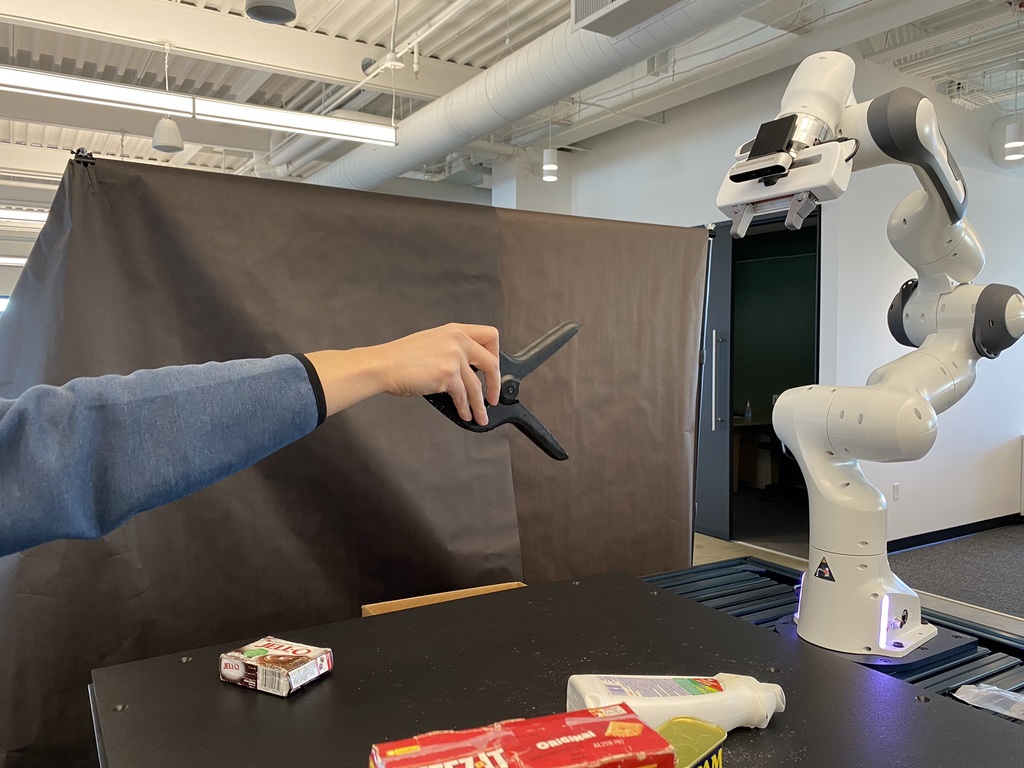}
   \\ \vspace{-1mm}
   pose 1
  \end{minipage}~
  \begin{minipage}{0.320\linewidth}
   \centering
   \includegraphics[width=\linewidth,trim={310 224 380 210},clip]{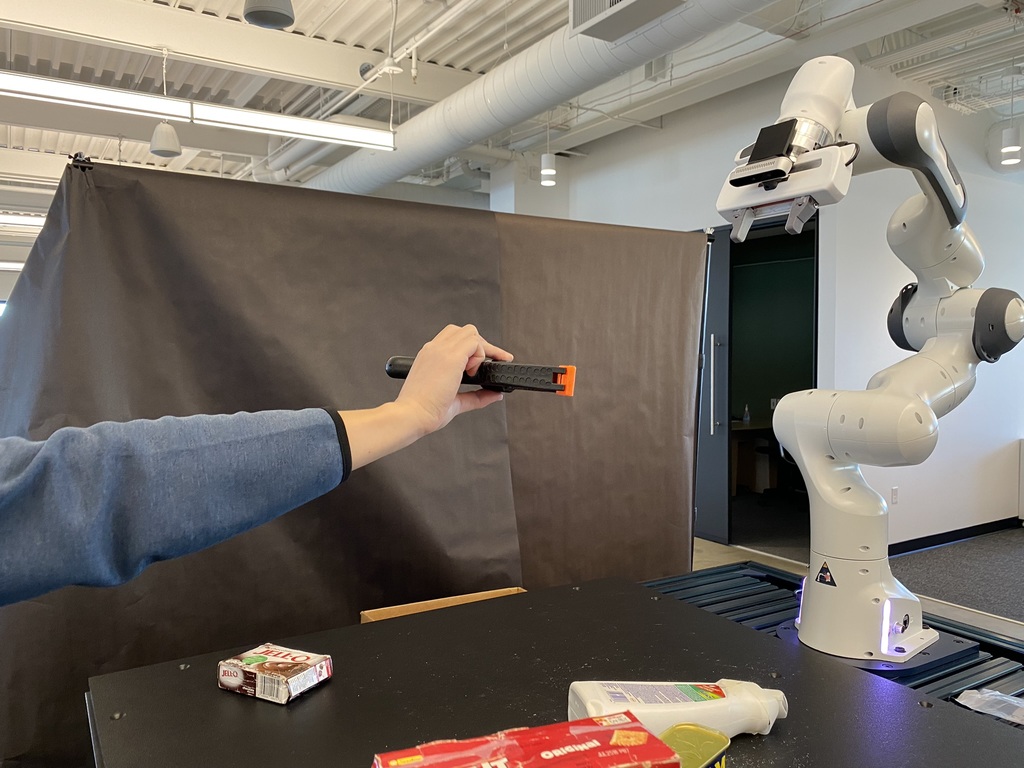}
   \\ \vspace{-1mm}
   pose 2
  \end{minipage}~
  \begin{minipage}{0.320\linewidth}
   \centering
   \includegraphics[width=\linewidth,trim={290 264 400 170},clip]{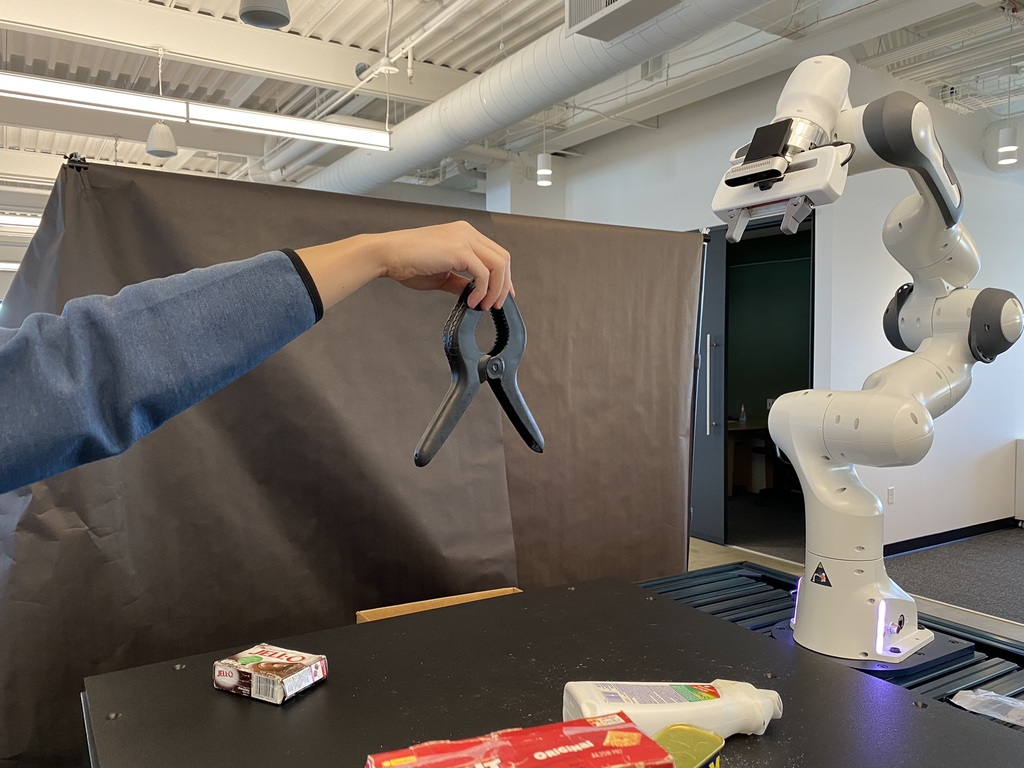}
   \\ \vspace{-1mm}
   pose 3
  \end{minipage}
  \\ \vspace{2mm}
 \end{minipage}~
 \begin{minipage}{0.495\linewidth}
  \centering
  008\_pudding\_box
  \\~\\ \vspace{-3mm}
  \begin{minipage}{0.320\linewidth}
   \centering
   \includegraphics[width=\linewidth,trim={270 174 420 260},clip]{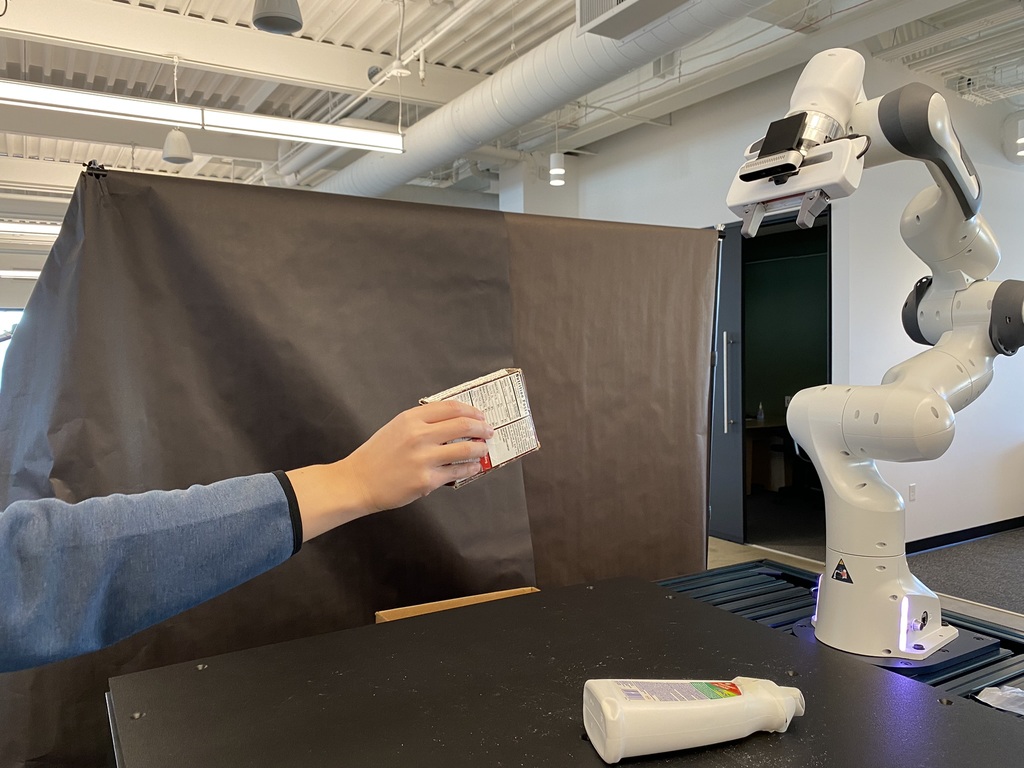}
   \\ \vspace{-1mm}
   pose 1
  \end{minipage}~
  \begin{minipage}{0.320\linewidth}
   \centering
   \includegraphics[width=\linewidth,trim={270 224 420 210},clip]{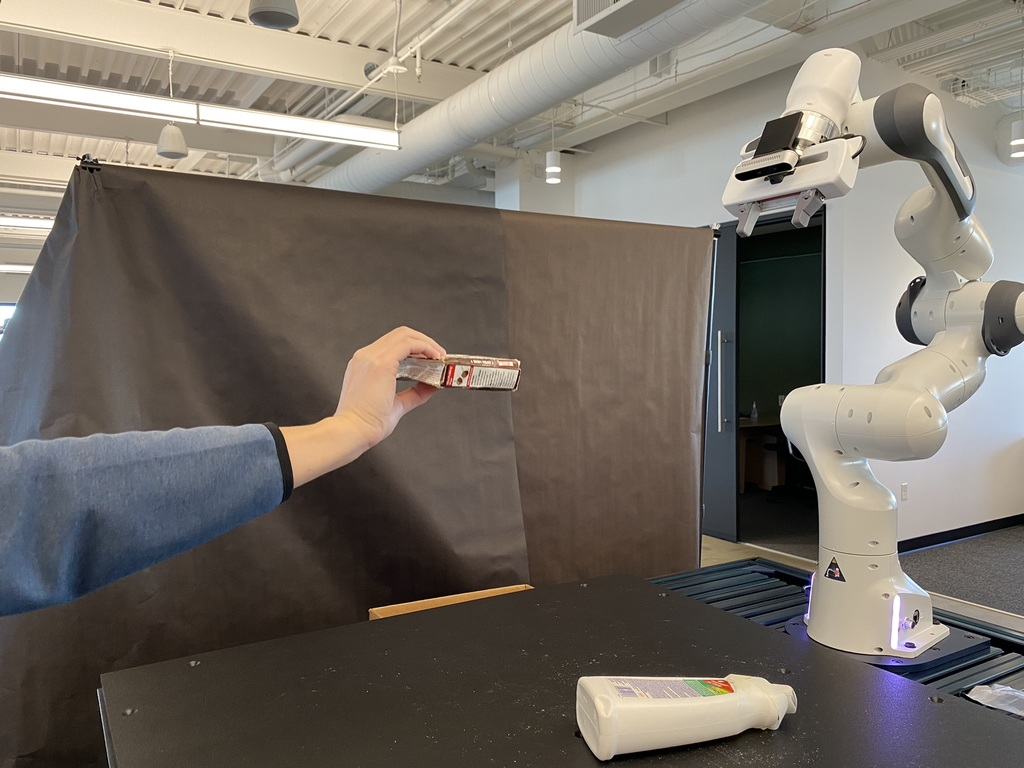}
   \\ \vspace{-1mm}
   pose 2
  \end{minipage}~
  \begin{minipage}{0.320\linewidth}
   \centering
   \includegraphics[width=\linewidth,trim={270 264 420 170},clip]{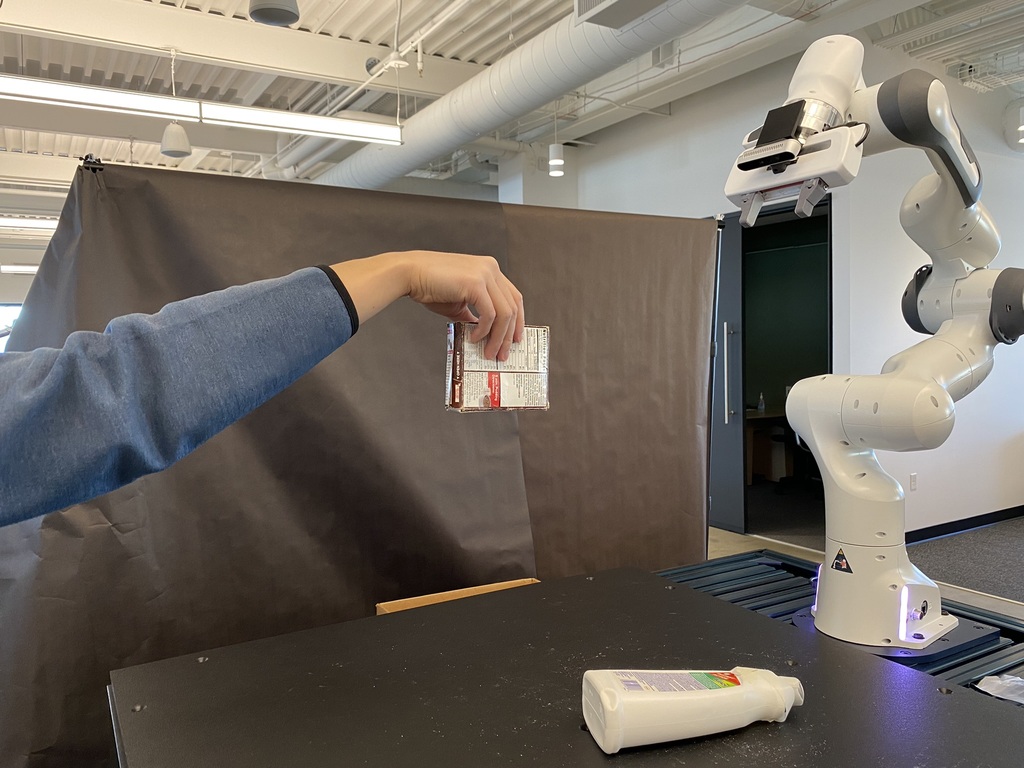}
   \\ \vspace{-1mm}
   pose 3
  \end{minipage}
  \\ \vspace{2mm}
 \end{minipage}
 \\ \vspace{3mm}
 \begin{minipage}{0.495\linewidth}
  \centering
  010\_potted\_meat\_can
  \\~\\ \vspace{-3mm}
  \begin{minipage}{0.320\linewidth}
   \centering
   \includegraphics[width=\linewidth,trim={270 184 420 250},clip]{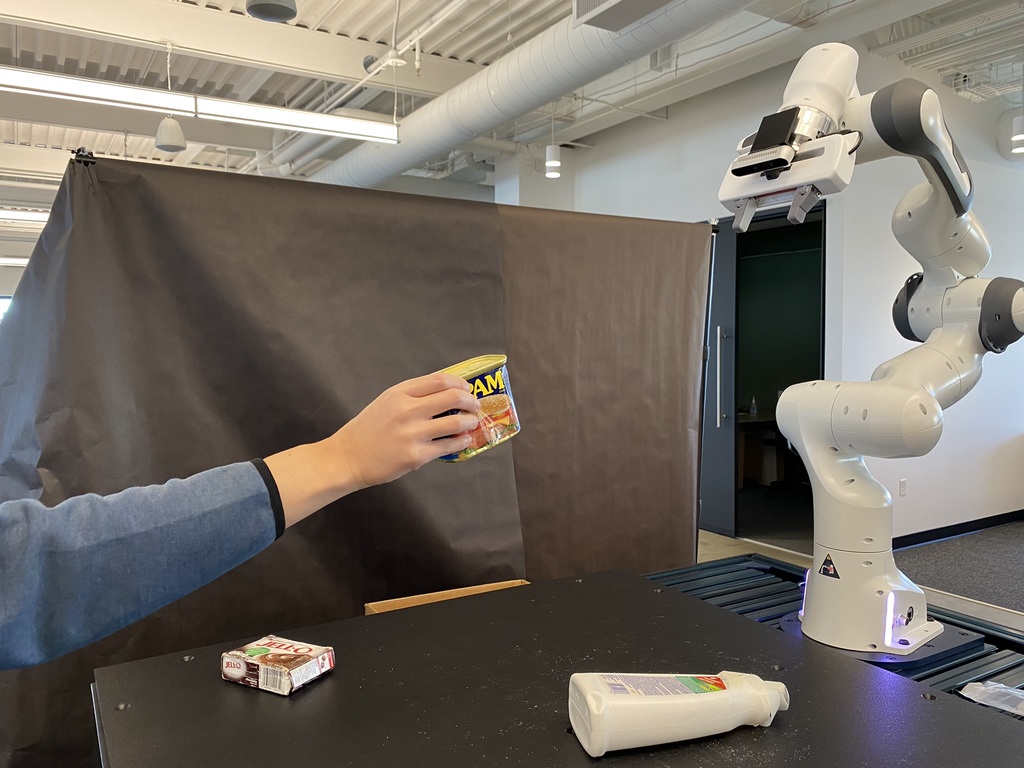}
   \\ \vspace{-1mm}
   pose 1
  \end{minipage}~
  \begin{minipage}{0.320\linewidth}
   \centering
   \includegraphics[width=\linewidth,trim={230 224 460 210},clip]{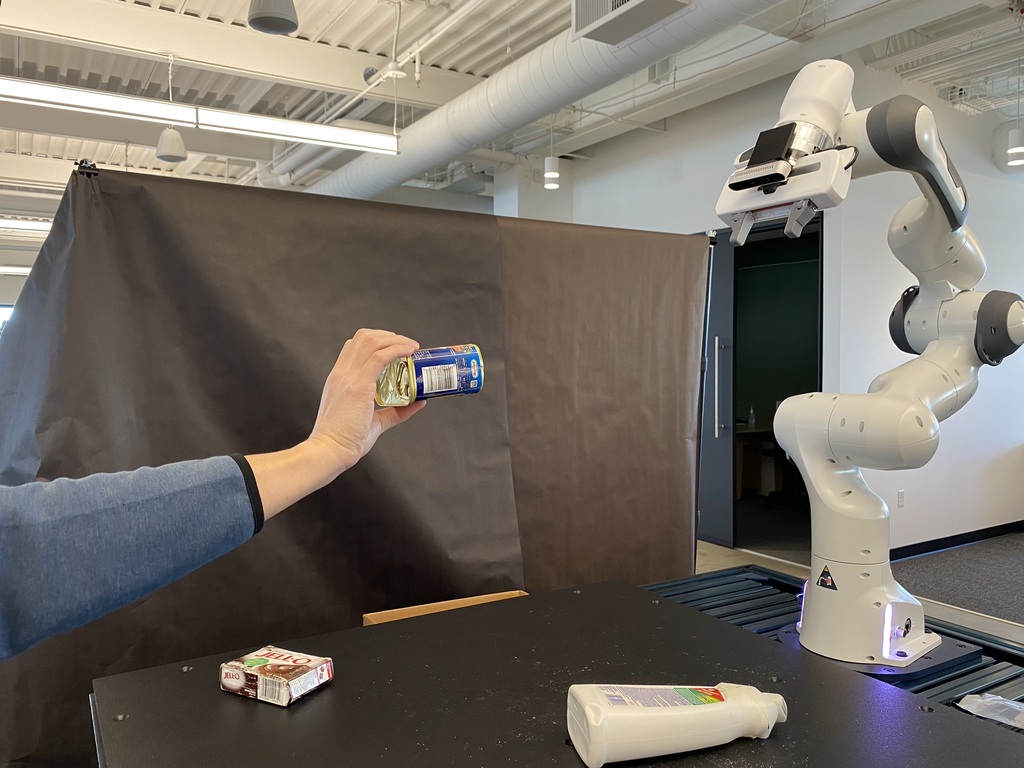}
   \\ \vspace{-1mm}
   pose 2
  \end{minipage}~
  \begin{minipage}{0.320\linewidth}
   \centering
   \includegraphics[width=\linewidth,trim={270 224 420 210},clip]{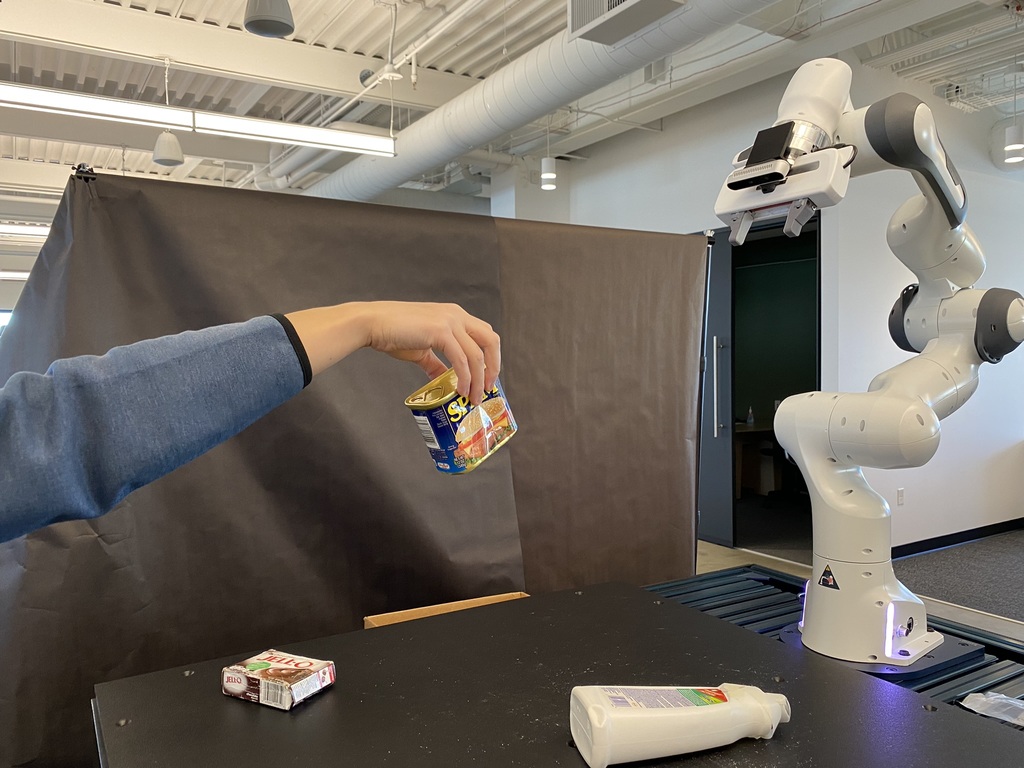}
   \\ \vspace{-1mm}
   pose 3
  \end{minipage}
  \\ \vspace{2mm}
 \end{minipage}~
 \begin{minipage}{0.495\linewidth}
  \centering
  021\_bleach\_cleanser
  \\~\\ \vspace{-3mm}
  \begin{minipage}{0.320\linewidth}
   \centering
   \includegraphics[width=\linewidth,trim={290 184 400 250},clip]{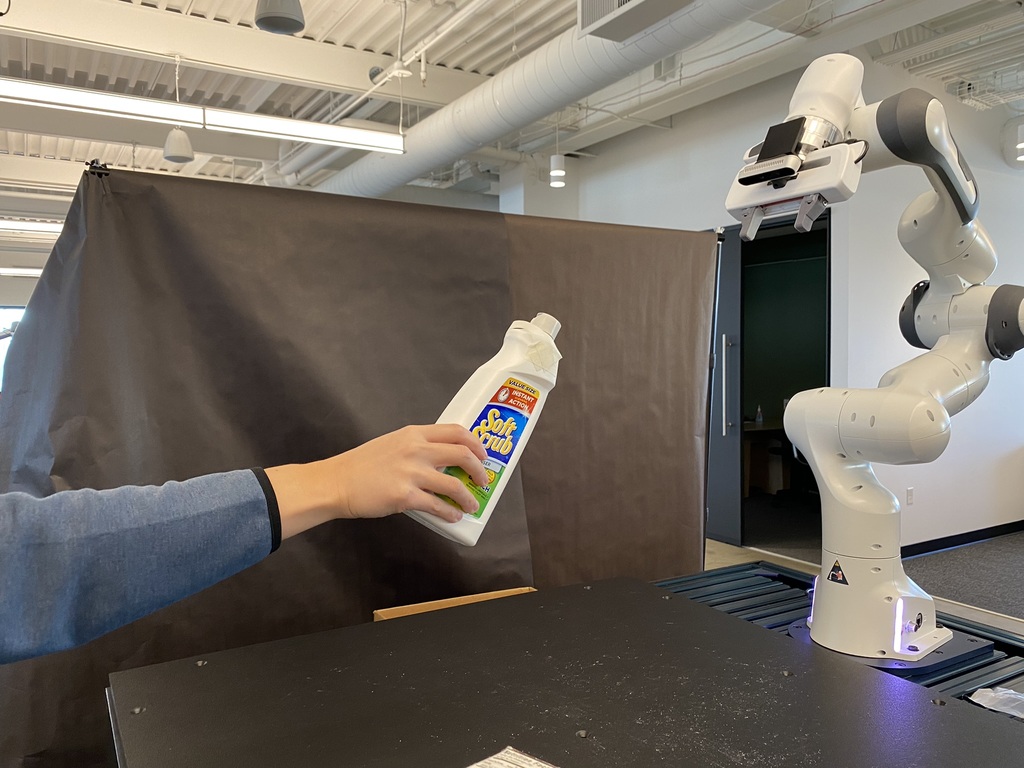}
   \\ \vspace{-1mm}
   pose 1
  \end{minipage}~
  \begin{minipage}{0.320\linewidth}
   \centering
   \includegraphics[width=\linewidth,trim={310 224 380 210},clip]{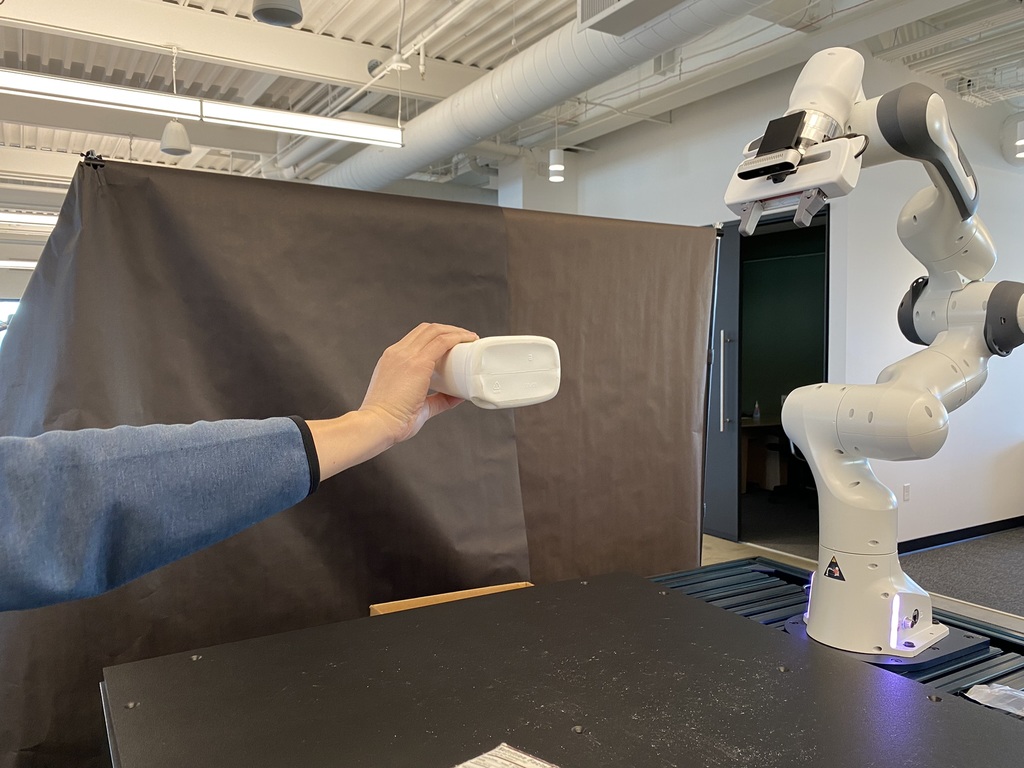}
   \\ \vspace{-1mm}
   pose 2
  \end{minipage}~
  \begin{minipage}{0.320\linewidth}
   \centering
   \includegraphics[width=\linewidth,trim={290 204 400 230},clip]{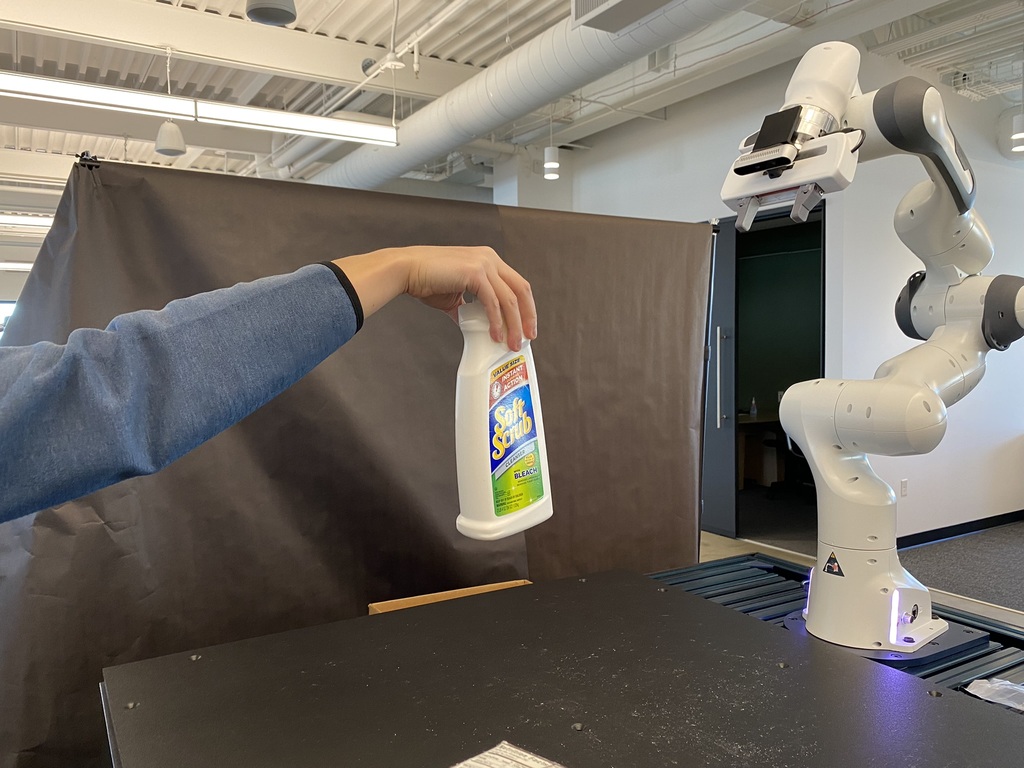}
   \\ \vspace{-1mm}
   pose 3
  \end{minipage}
  \\ \vspace{2mm}
 \end{minipage}
 \caption{\small The instructed handover poses for the \textbf{right hand} in the pilot study. We adopt 10 objects from the YCB-Video dataset~\cite{xiang:rss2018} and pre-select 3 handover poses per object with varying handover difficulties, totaling 30 handover poses for the right hand.}
 \label{fig:pilot_right}
\end{figure*}

\begin{figure*}[t!]
 \centering
 \begin{minipage}{0.495\linewidth}
  \centering
  011\_banana
  \\~\\ \vspace{-3mm}
  \begin{minipage}{0.320\linewidth}
   \centering
   \includegraphics[width=\linewidth,trim={320 274 390 180},clip]{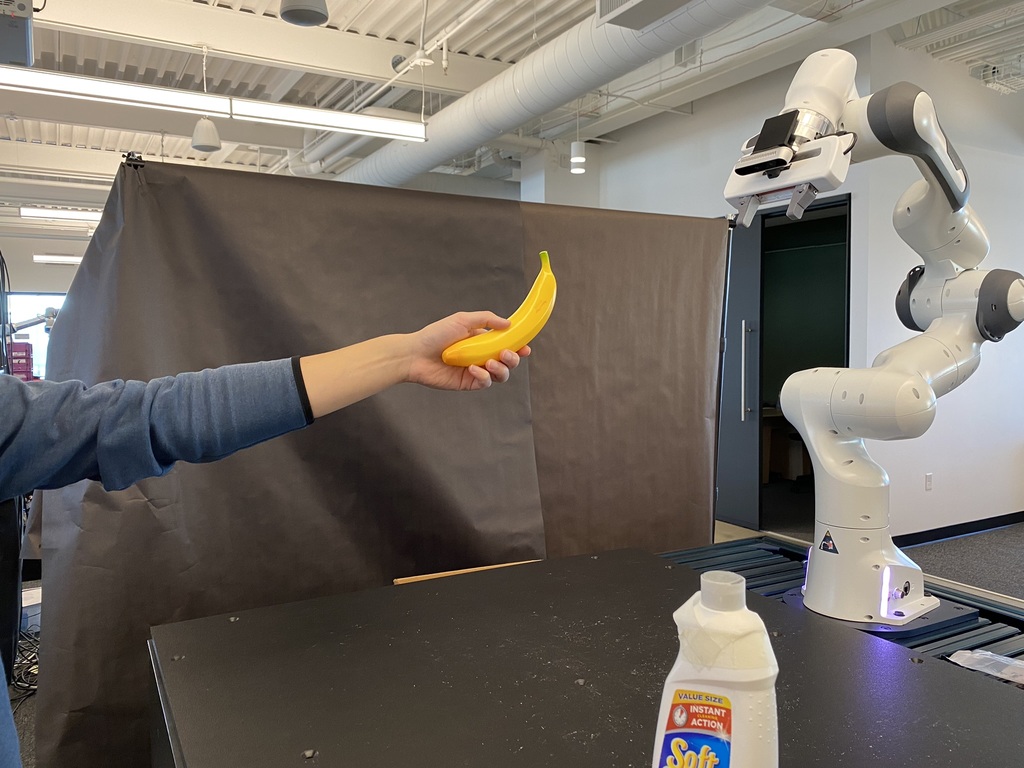}
   \\ \vspace{-1mm}
   pose 1
  \end{minipage}~
  \begin{minipage}{0.320\linewidth}
   \centering
   \includegraphics[width=\linewidth,trim={320 274 390 180},clip]{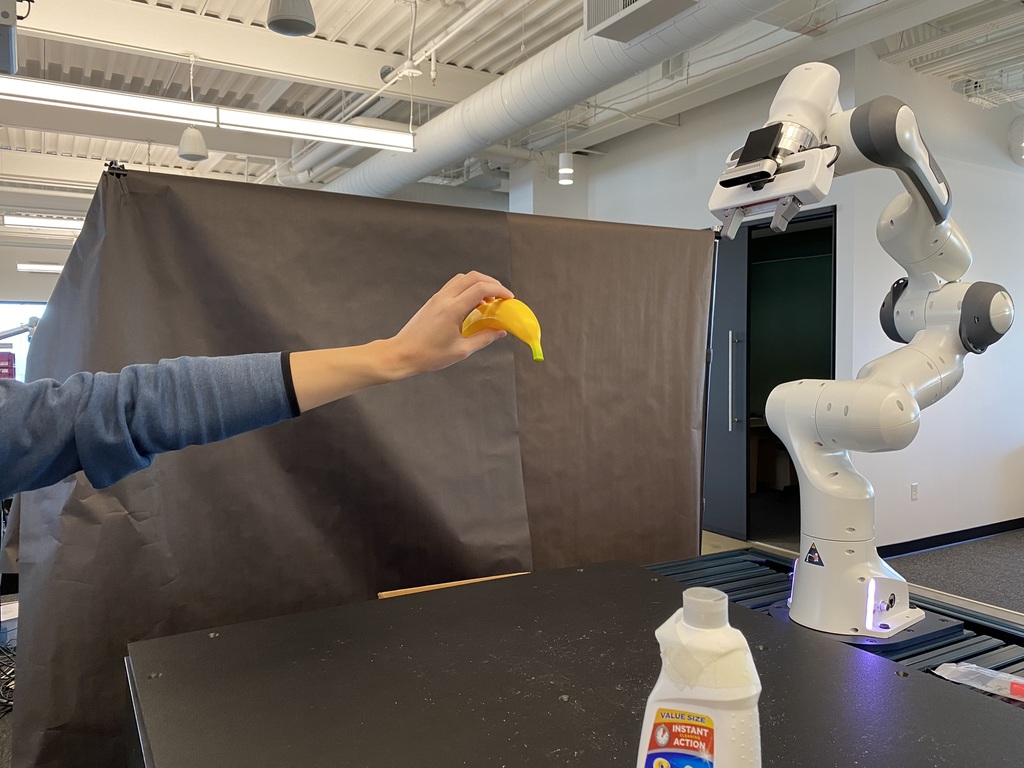}
   \\ \vspace{-1mm}
   pose 2
  \end{minipage}~
  \begin{minipage}{0.320\linewidth}
   \centering
   \includegraphics[width=\linewidth,trim={260 254 450 200},clip]{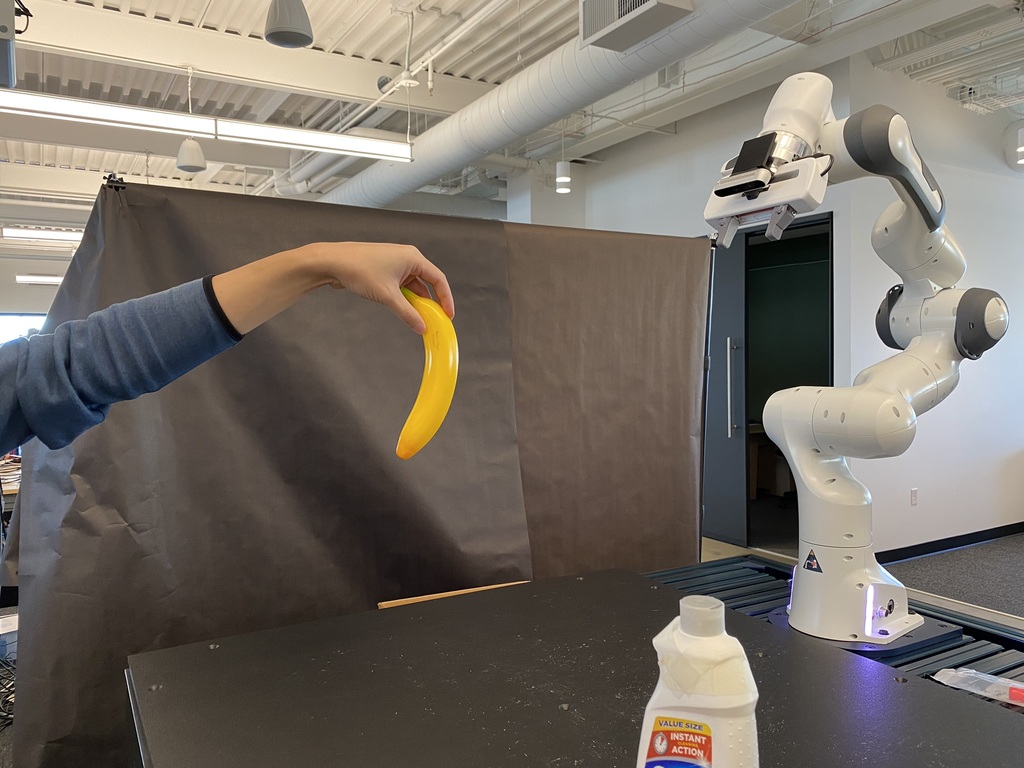}
   \\ \vspace{-1mm}
   pose 3
  \end{minipage}
  \\ \vspace{2mm}
 \end{minipage}~
 \begin{minipage}{0.495\linewidth}
  \centering
  037\_scissors
  \\~\\ \vspace{-3mm}
  \begin{minipage}{0.320\linewidth}
   \centering
   \includegraphics[width=\linewidth,trim={260 234 450 220},clip]{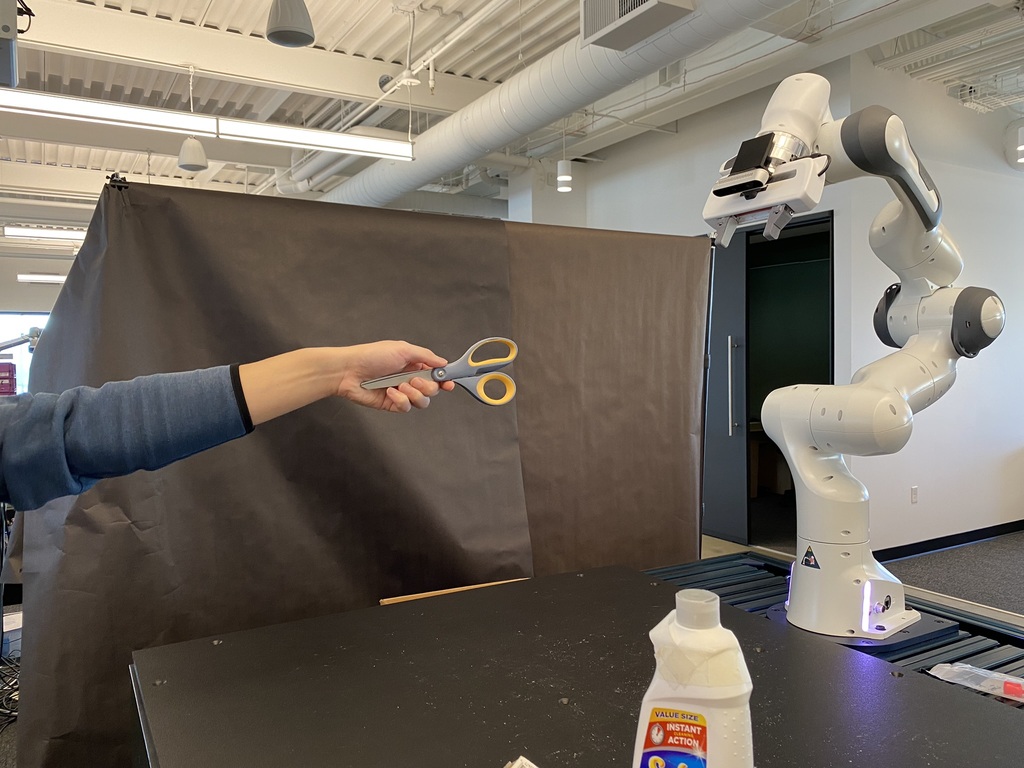}
   \\ \vspace{-1mm}
   pose 1
  \end{minipage}~
  \begin{minipage}{0.320\linewidth}
   \centering
   \includegraphics[width=\linewidth,trim={280 274 430 180},clip]{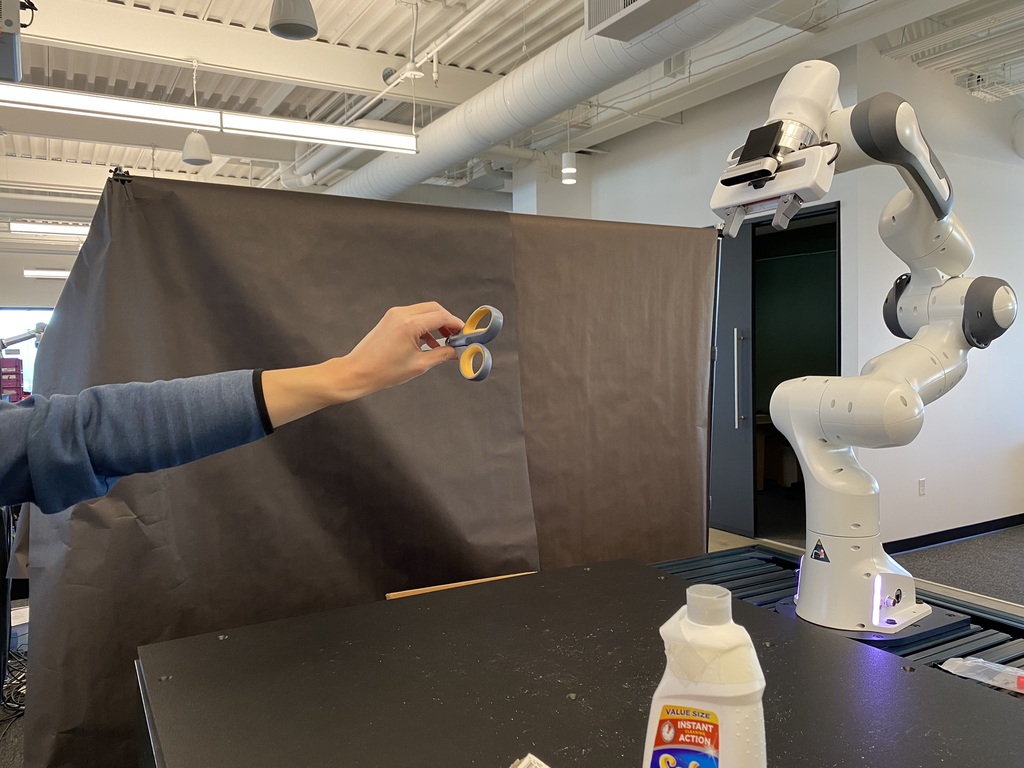}
   \\ \vspace{-1mm}
   pose 2
  \end{minipage}~
  \begin{minipage}{0.320\linewidth}
   \centering
   \includegraphics[width=\linewidth,trim={280 214 430 240},clip]{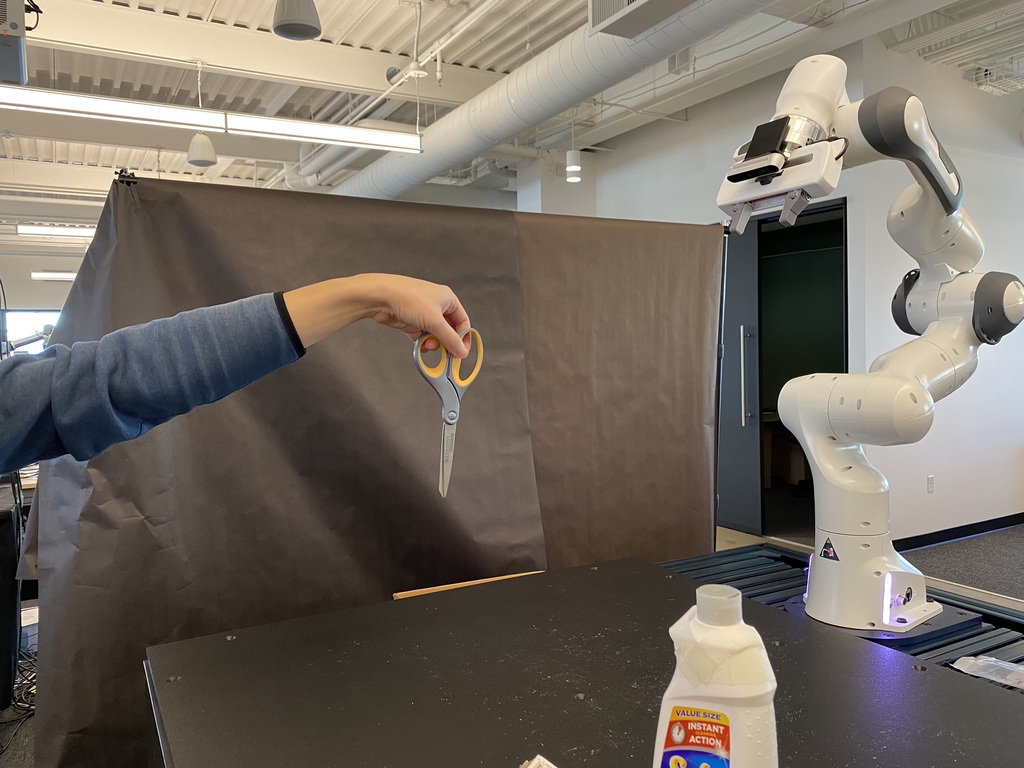}
   \\ \vspace{-1mm}
   pose 3
  \end{minipage}
  \\ \vspace{2mm}
 \end{minipage}
 \\ \vspace{3mm}
 \begin{minipage}{0.495\linewidth}
  \centering
  006\_mustard\_bottle
  \\~\\ \vspace{-3mm}
  \begin{minipage}{0.320\linewidth}
   \centering
   \includegraphics[width=\linewidth,trim={260 214 430 240},clip]{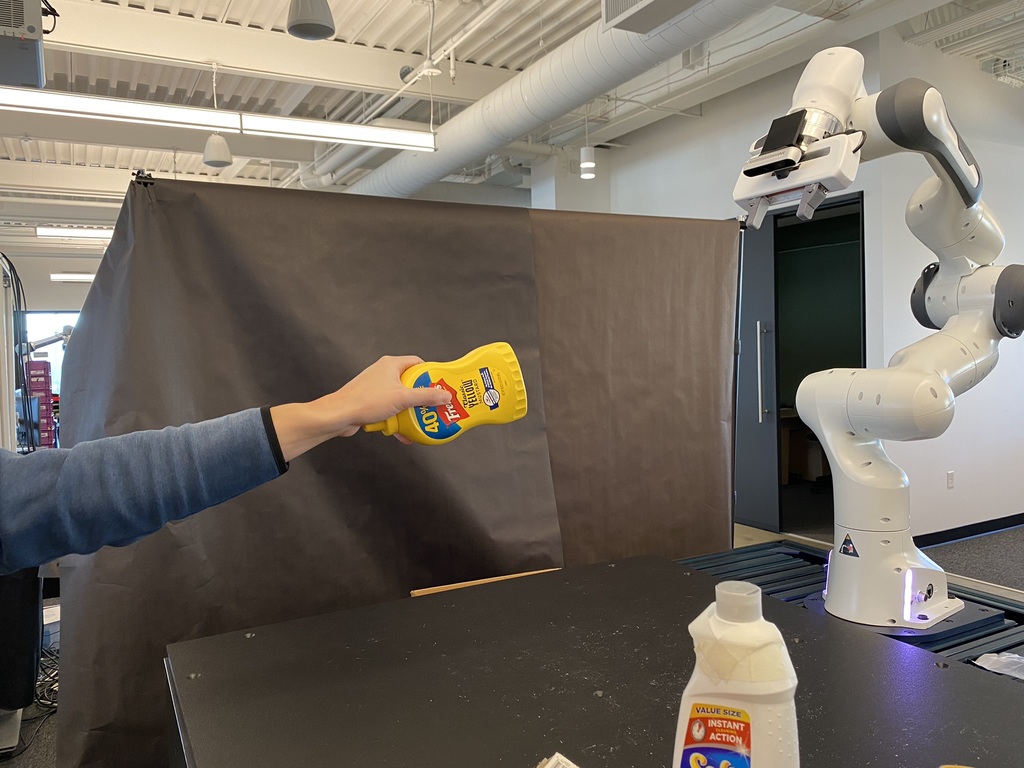}
   \\ \vspace{-1mm}
   pose 1
  \end{minipage}~
  \begin{minipage}{0.320\linewidth}
   \centering
   \includegraphics[width=\linewidth,trim={280 254 430 200},clip]{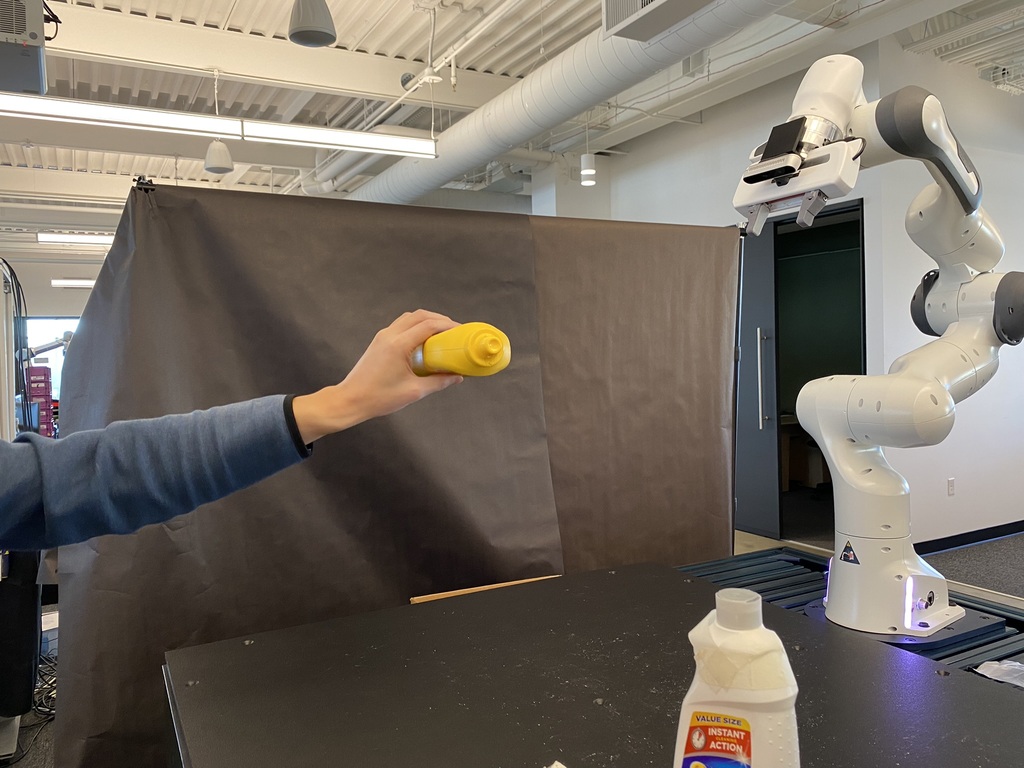}
   \\ \vspace{-1mm}
   pose 2
  \end{minipage}~
  \begin{minipage}{0.320\linewidth}
   \centering
   \includegraphics[width=\linewidth,trim={280 254 430 200},clip]{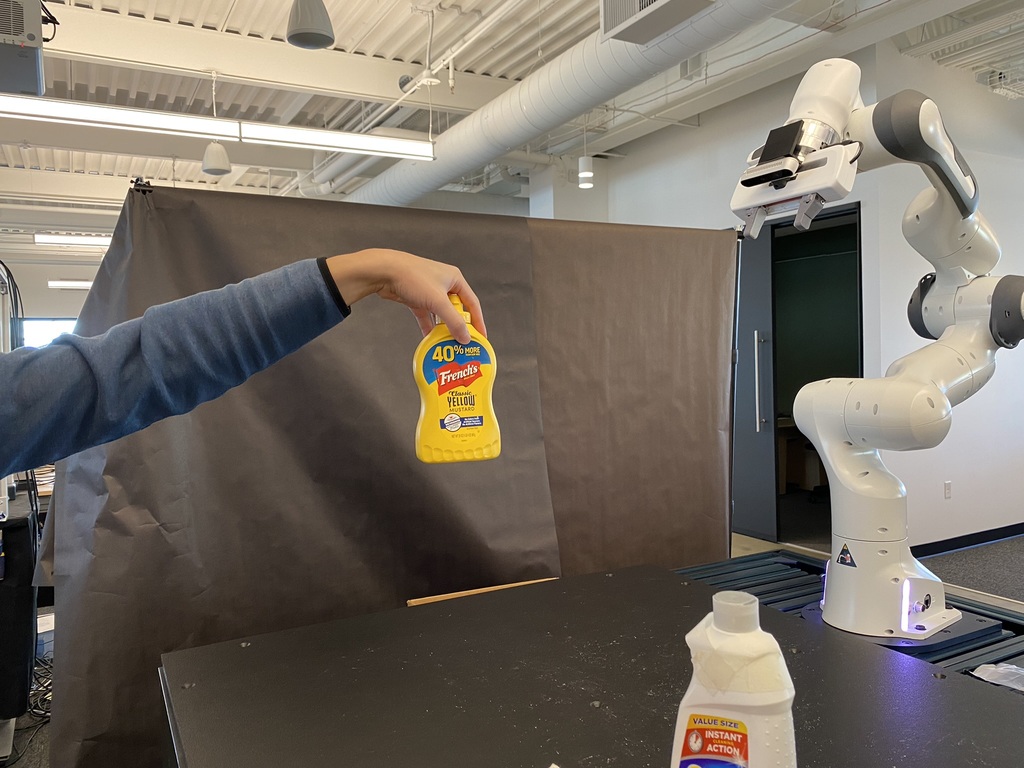}
   \\ \vspace{-1mm}
   pose 3
  \end{minipage}
  \\ \vspace{2mm}
 \end{minipage}~
 \begin{minipage}{0.495\linewidth}
  \centering
  024\_bowl
  \\~\\ \vspace{-3mm}
  \begin{minipage}{0.320\linewidth}
   \centering
   \includegraphics[width=\linewidth,trim={360 214 350 240},clip]{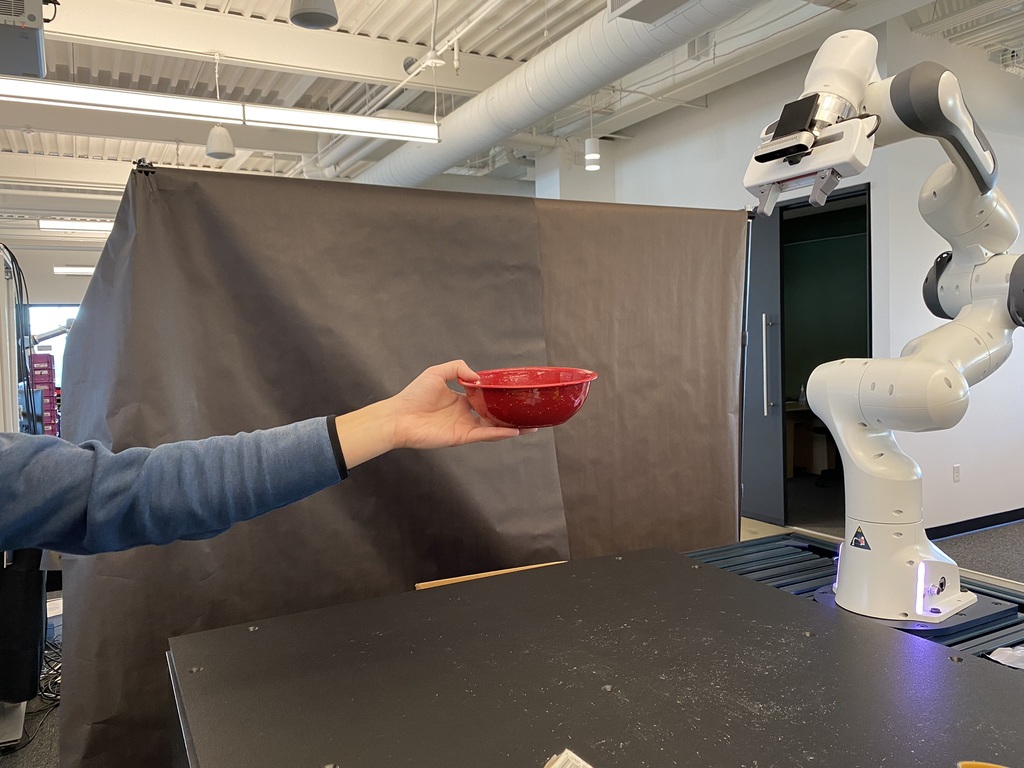}
   \\ \vspace{-1mm}
   pose 1
  \end{minipage}~
  \begin{minipage}{0.320\linewidth}
   \centering
   \includegraphics[width=\linewidth,trim={280 314 430 140},clip]{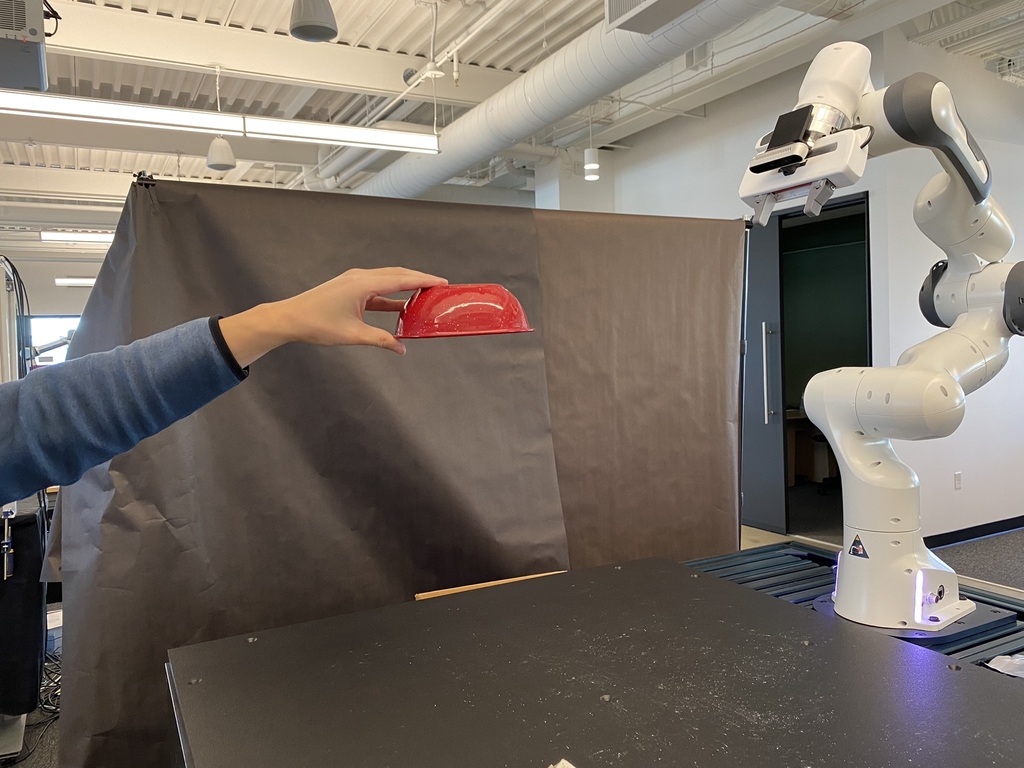}
   \\ \vspace{-1mm}
   pose 2
  \end{minipage}~
  \begin{minipage}{0.320\linewidth}
   \centering
   \includegraphics[width=\linewidth,trim={280 254 430 200},clip]{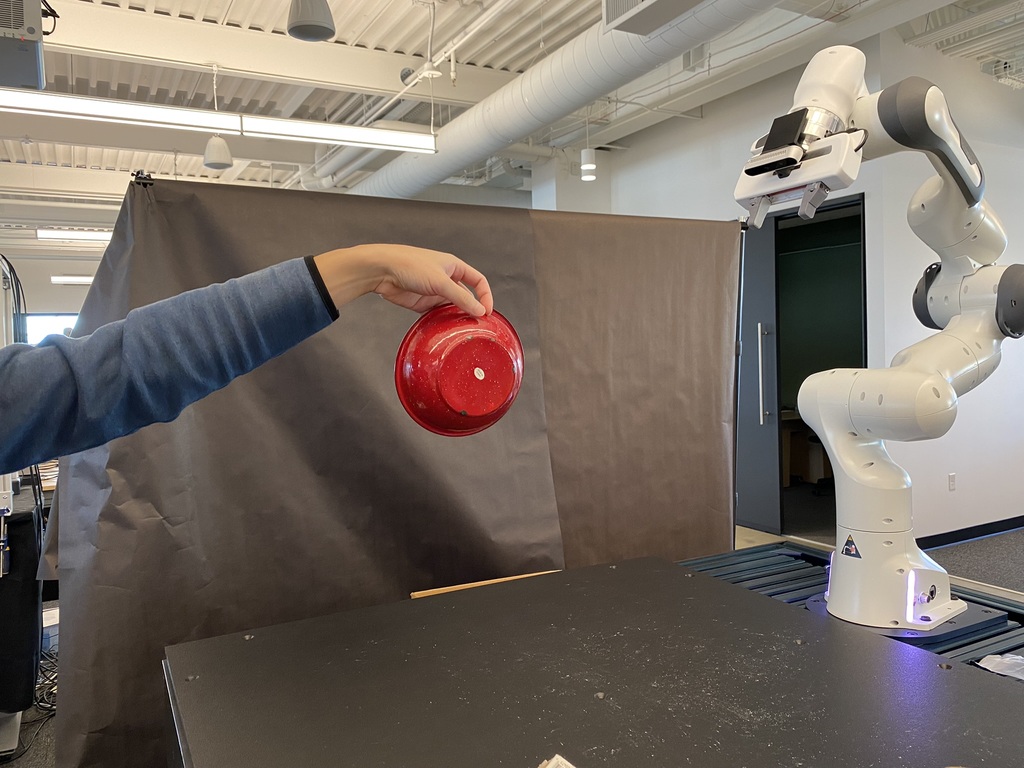}
   \\ \vspace{-1mm}
   pose 3
  \end{minipage}
  \\ \vspace{2mm}
 \end{minipage}
 \\ \vspace{3mm}
 \begin{minipage}{0.495\linewidth}
  \centering
  040\_large\_marker
  \\~\\ \vspace{-3mm}
  \begin{minipage}{0.320\linewidth}
   \centering
   \includegraphics[width=\linewidth,trim={320 234 390 220},clip]{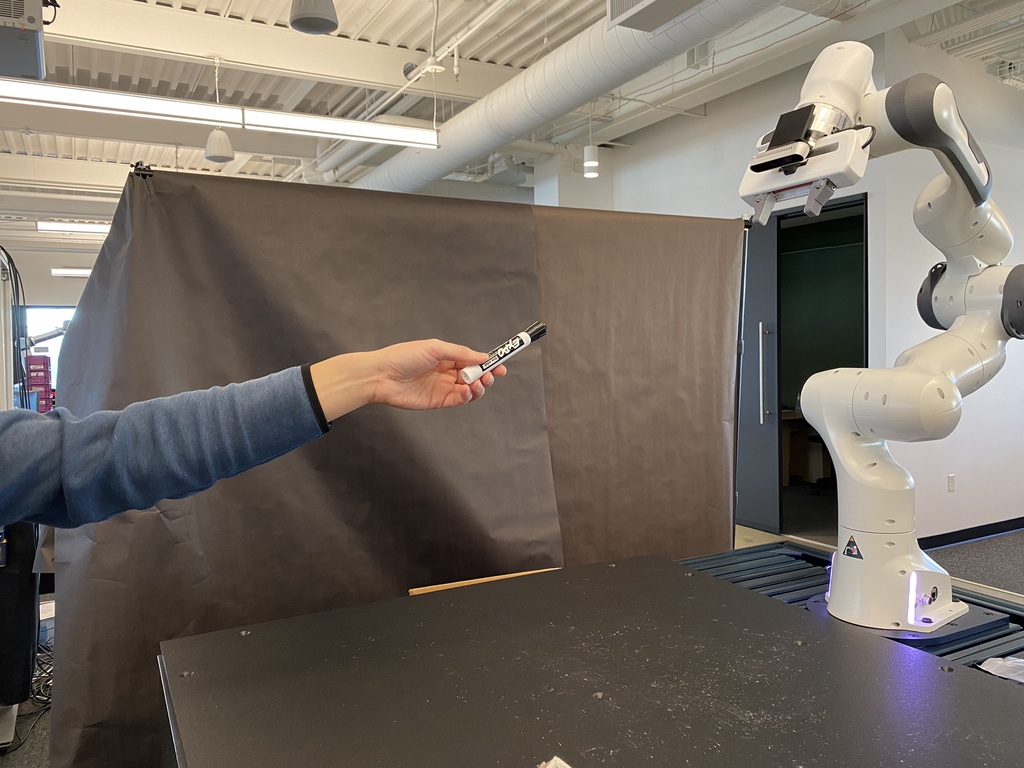}
   \\ \vspace{-1mm}
   pose 1
  \end{minipage}~
  \begin{minipage}{0.320\linewidth}
   \centering
   \includegraphics[width=\linewidth,trim={300 294 410 160},clip]{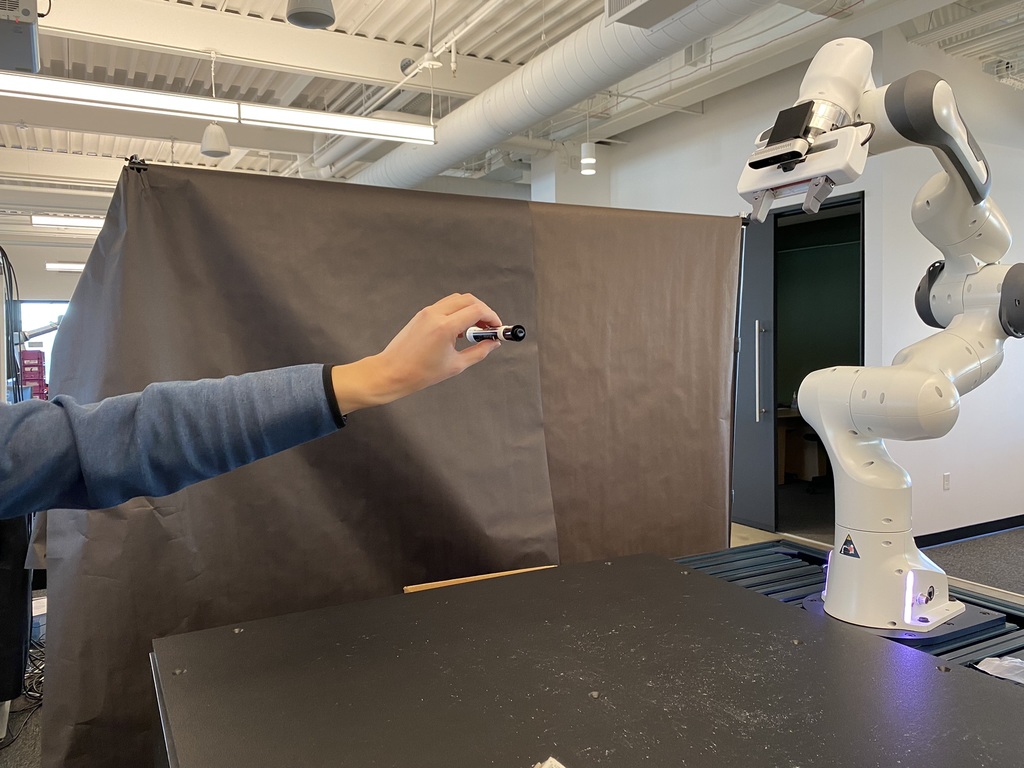}
   \\ \vspace{-1mm}
   pose 2
  \end{minipage}~
  \begin{minipage}{0.320\linewidth}
   \centering
   \includegraphics[width=\linewidth,trim={320 274 390 180},clip]{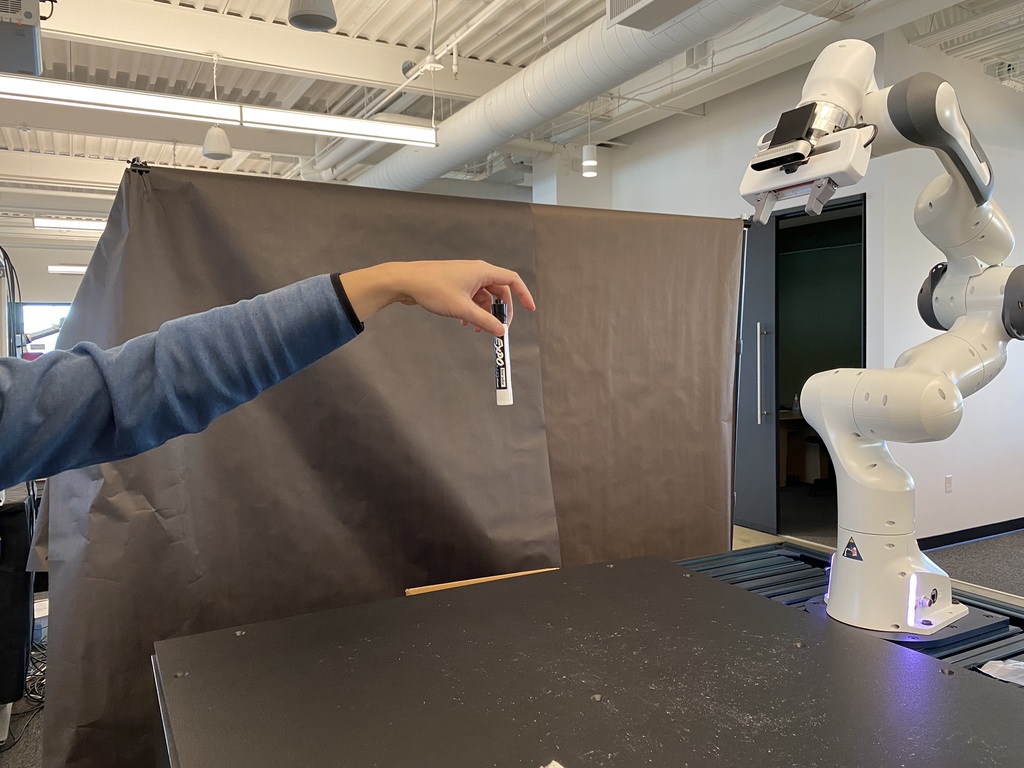}
   \\ \vspace{-1mm}
   pose 3
  \end{minipage}
  \\ \vspace{2mm}
 \end{minipage}~
 \begin{minipage}{0.495\linewidth}
  \centering
  003\_cracker\_box
  \\~\\ \vspace{-3mm}
  \begin{minipage}{0.320\linewidth}
   \centering
   \includegraphics[width=\linewidth,trim={290 244 440 230},clip]{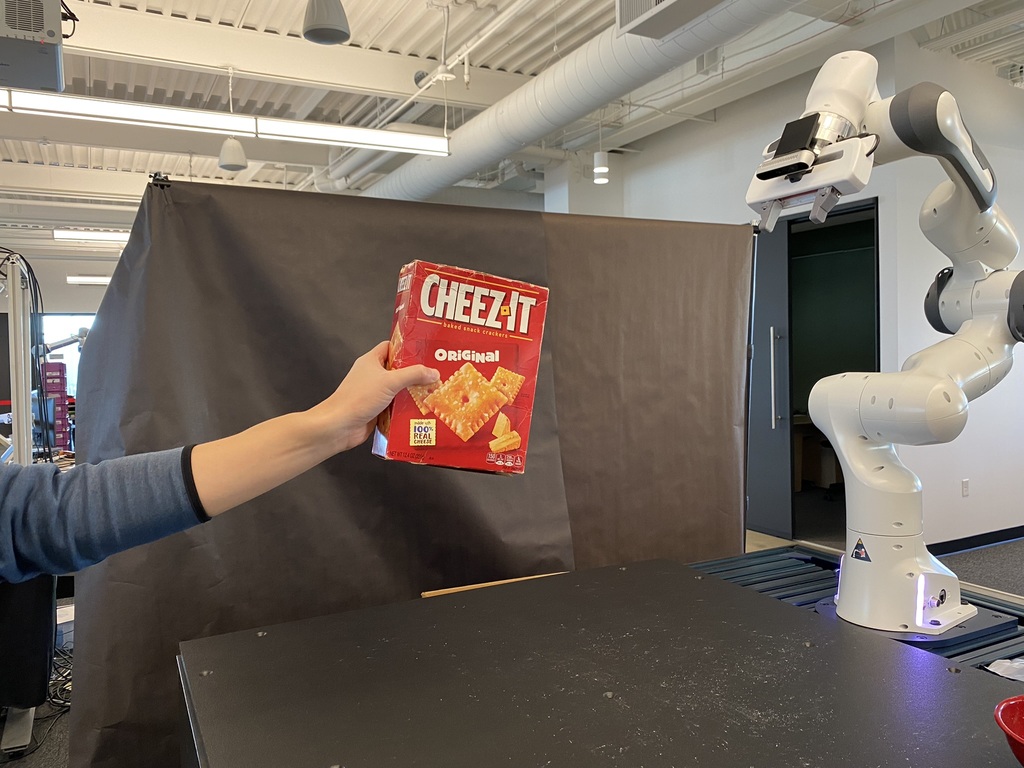}
   \\ \vspace{-1mm}
   pose 1
  \end{minipage}~
  \begin{minipage}{0.320\linewidth}
   \centering
   \includegraphics[width=\linewidth,trim={330 284 400 190},clip]{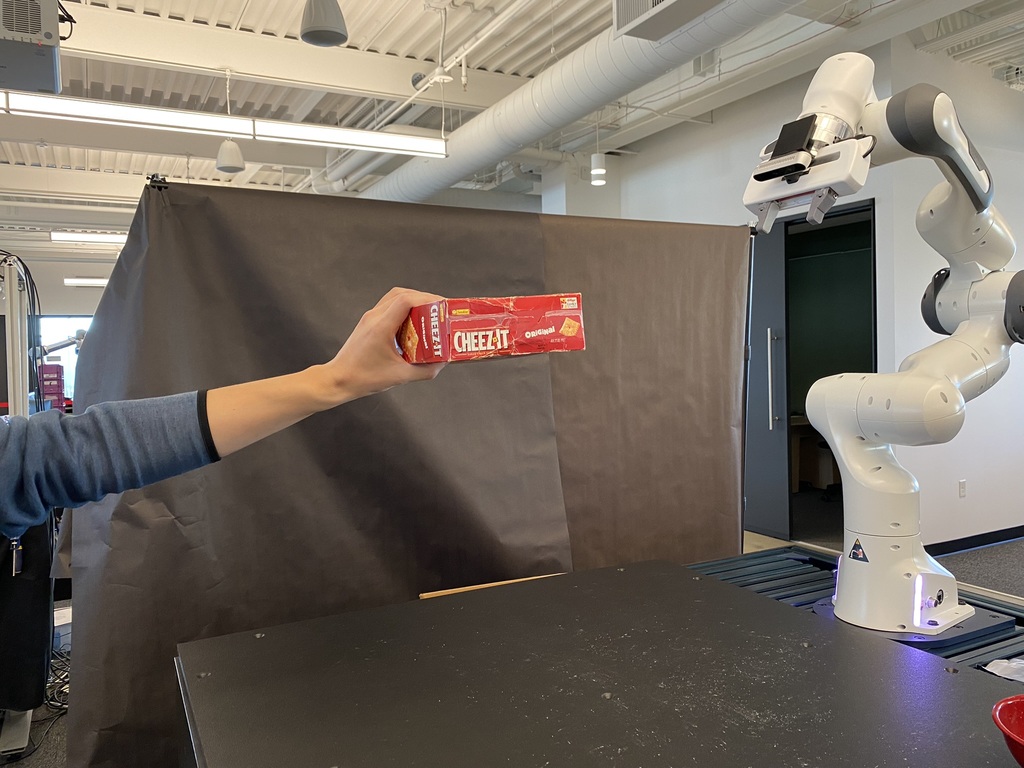}
   \\ \vspace{-1mm}
   pose 2
  \end{minipage}~
  \begin{minipage}{0.320\linewidth}
   \centering
   \includegraphics[width=\linewidth,trim={310 224 420 250},clip]{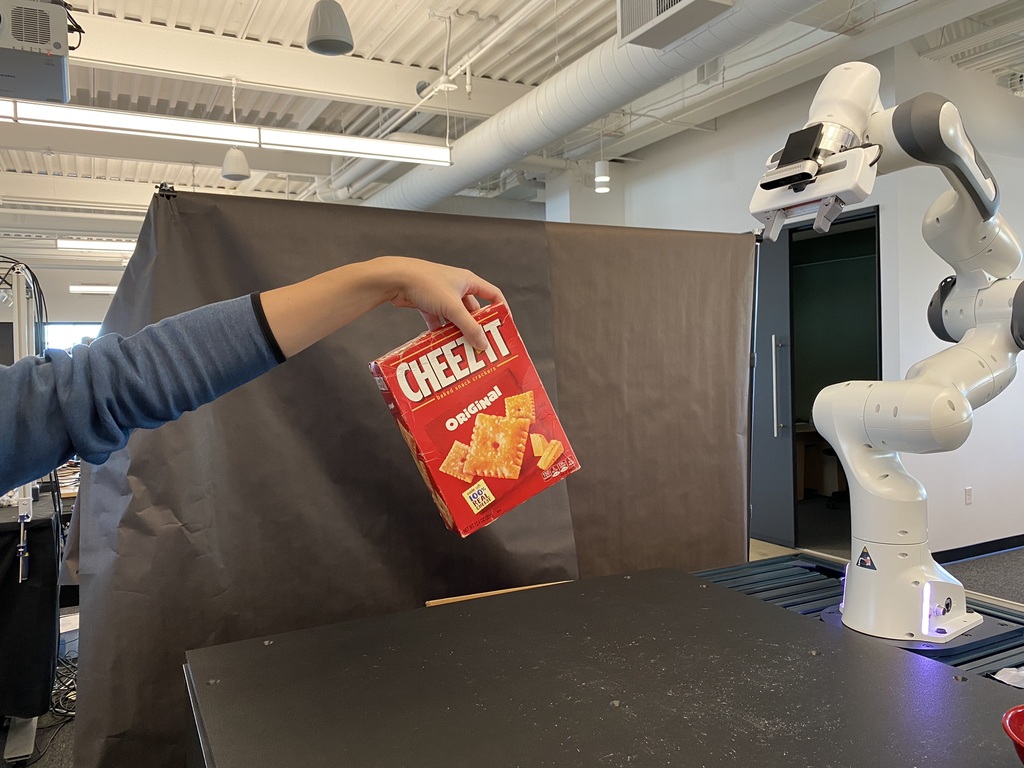}
   \\ \vspace{-1mm}
   pose 3
  \end{minipage}
  \\ \vspace{2mm}
 \end{minipage}
 \\ \vspace{3mm}
 \begin{minipage}{0.495\linewidth}
  \centering
  052\_extra\_large\_clamp
  \\~\\ \vspace{-3mm}
  \begin{minipage}{0.320\linewidth}
   \centering
   \includegraphics[width=\linewidth,trim={340 214 370 240},clip]{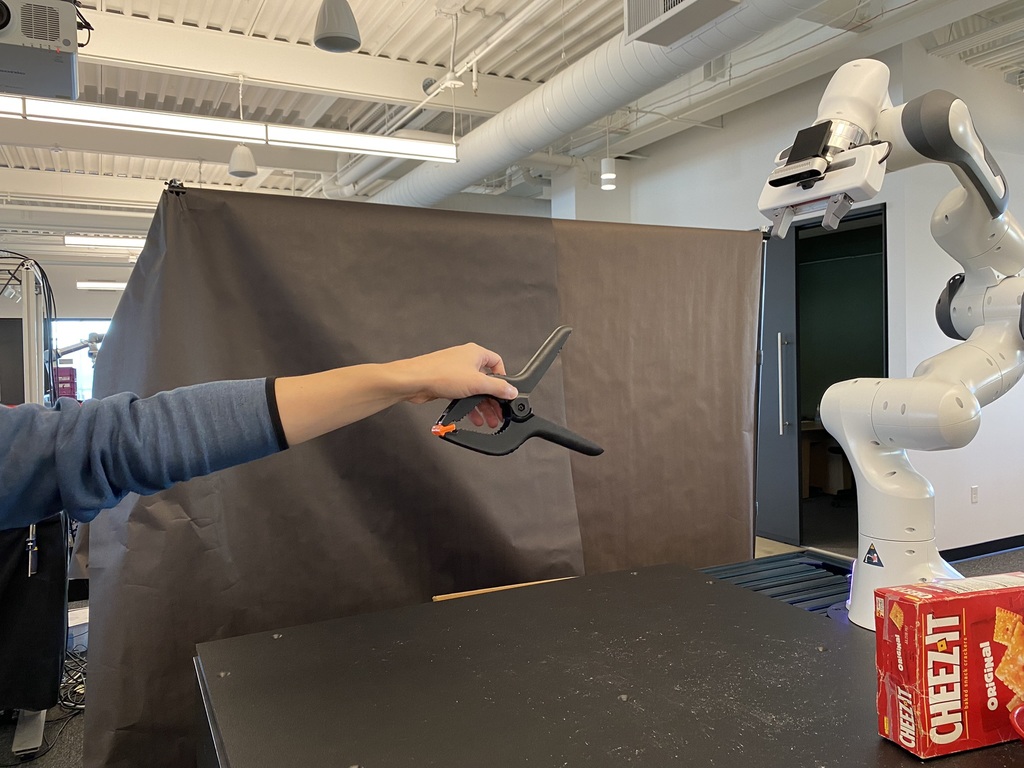}
   \\ \vspace{-1mm}
   pose 1
  \end{minipage}~
  \begin{minipage}{0.320\linewidth}
   \centering
   \includegraphics[width=\linewidth,trim={340 234 370 220},clip]{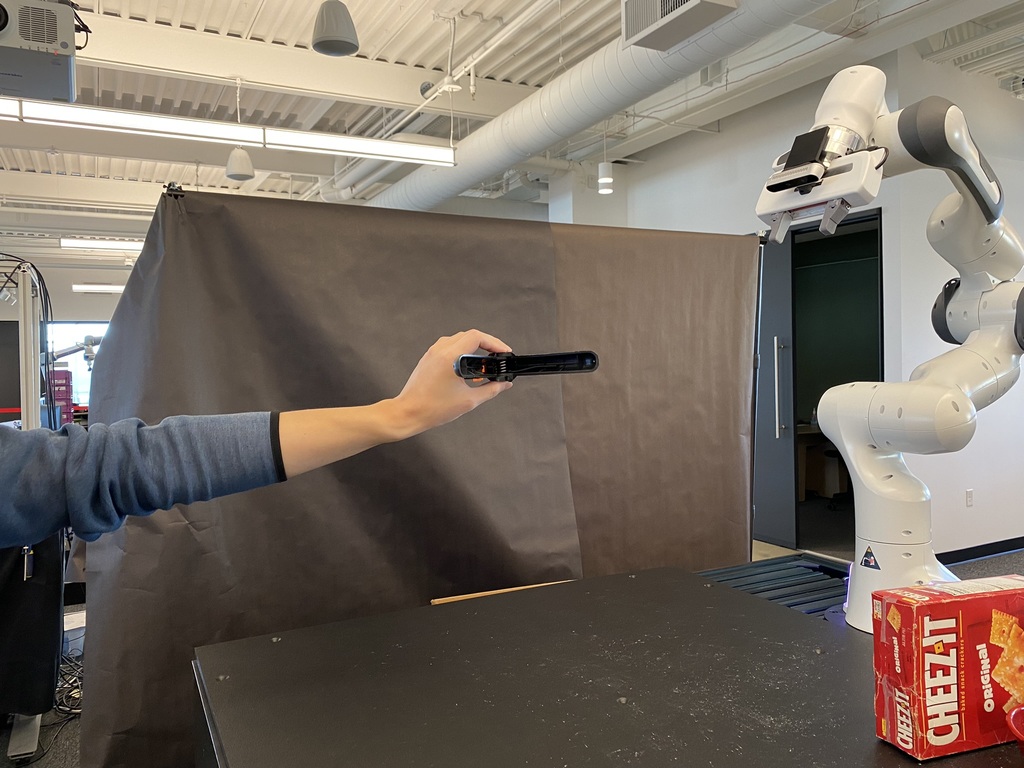}
   \\ \vspace{-1mm}
   pose 2
  \end{minipage}~
  \begin{minipage}{0.320\linewidth}
   \centering
   \includegraphics[width=\linewidth,trim={340 234 370 220},clip]{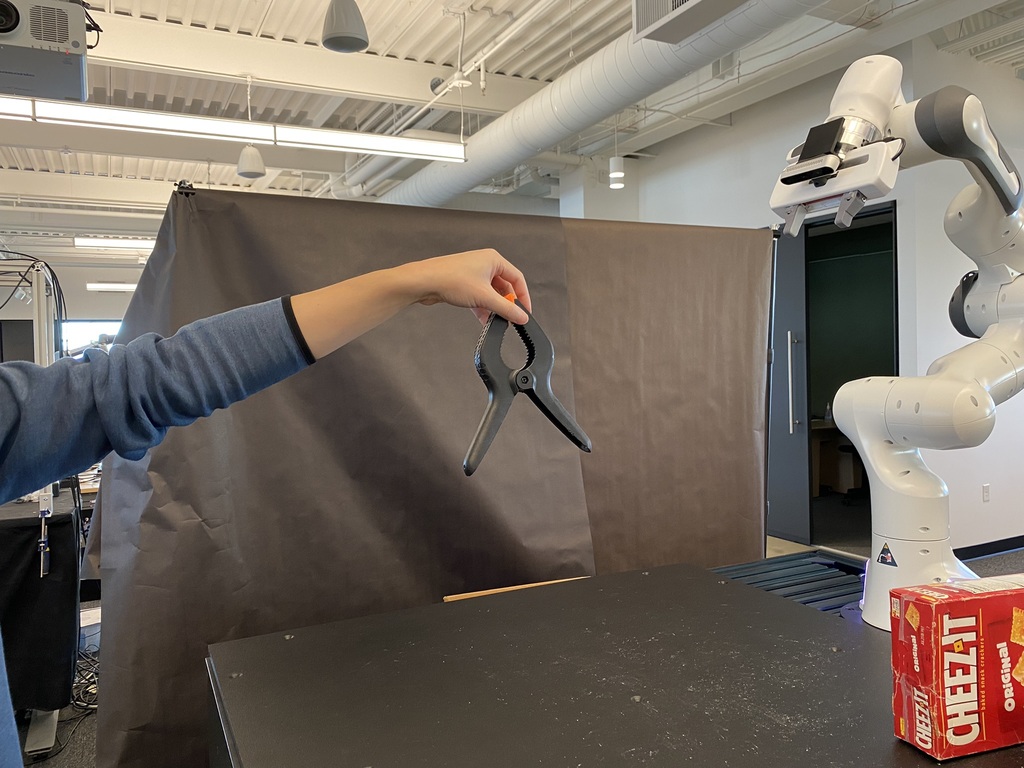}
   \\ \vspace{-1mm}
   pose 3
  \end{minipage}
  \\ \vspace{2mm}
 \end{minipage}~
 \begin{minipage}{0.495\linewidth}
  \centering
  008\_pudding\_box
  \\~\\ \vspace{-3mm}
  \begin{minipage}{0.320\linewidth}
   \centering
   \includegraphics[width=\linewidth,trim={300 234 410 220},clip]{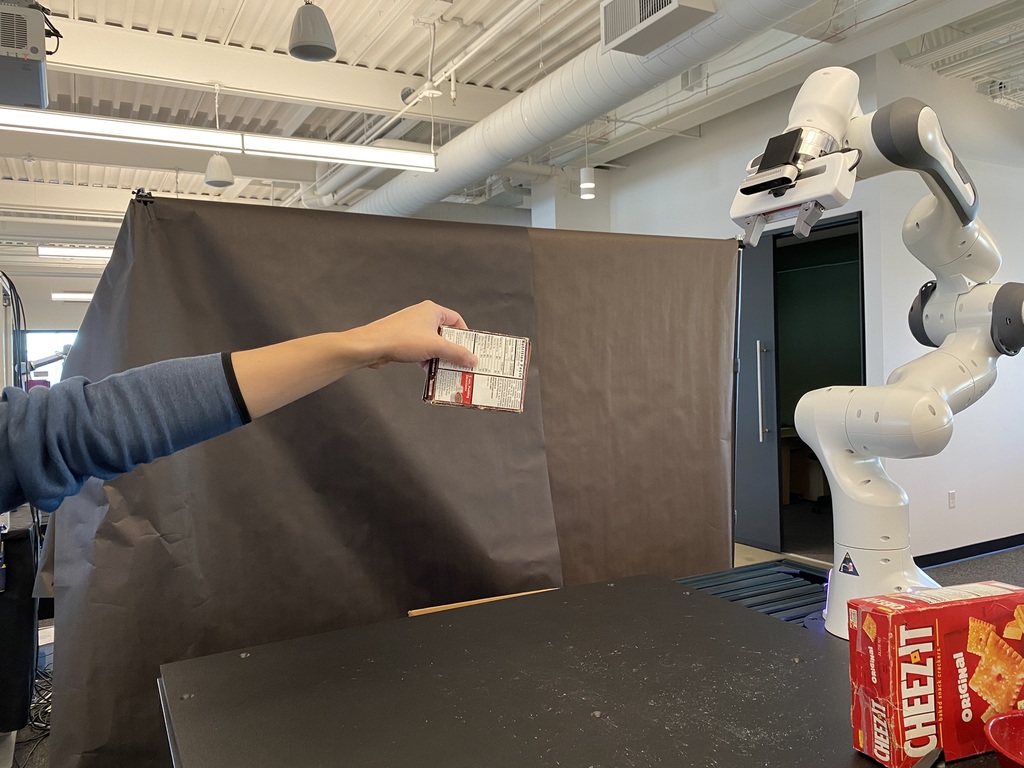}
   \\ \vspace{-1mm}
   pose 1
  \end{minipage}~
  \begin{minipage}{0.320\linewidth}
   \centering
   \includegraphics[width=\linewidth,trim={300 254 410 200},clip]{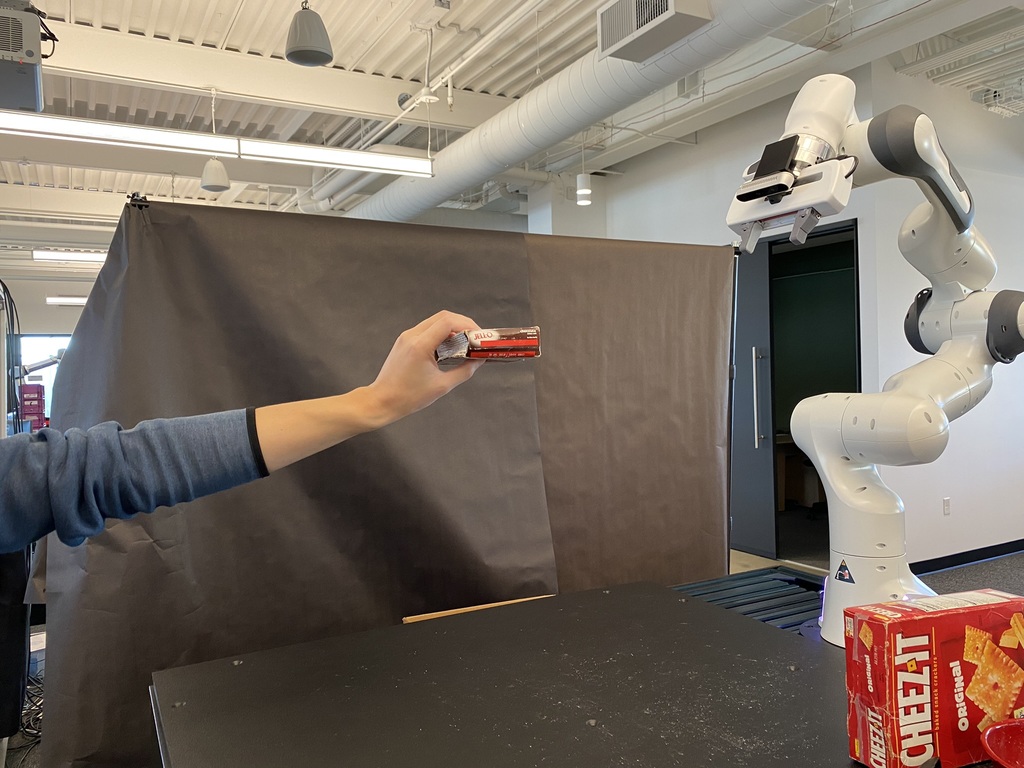}
   \\ \vspace{-1mm}
   pose 2
  \end{minipage}~
  \begin{minipage}{0.320\linewidth}
   \centering
   \includegraphics[width=\linewidth,trim={300 274 410 180},clip]{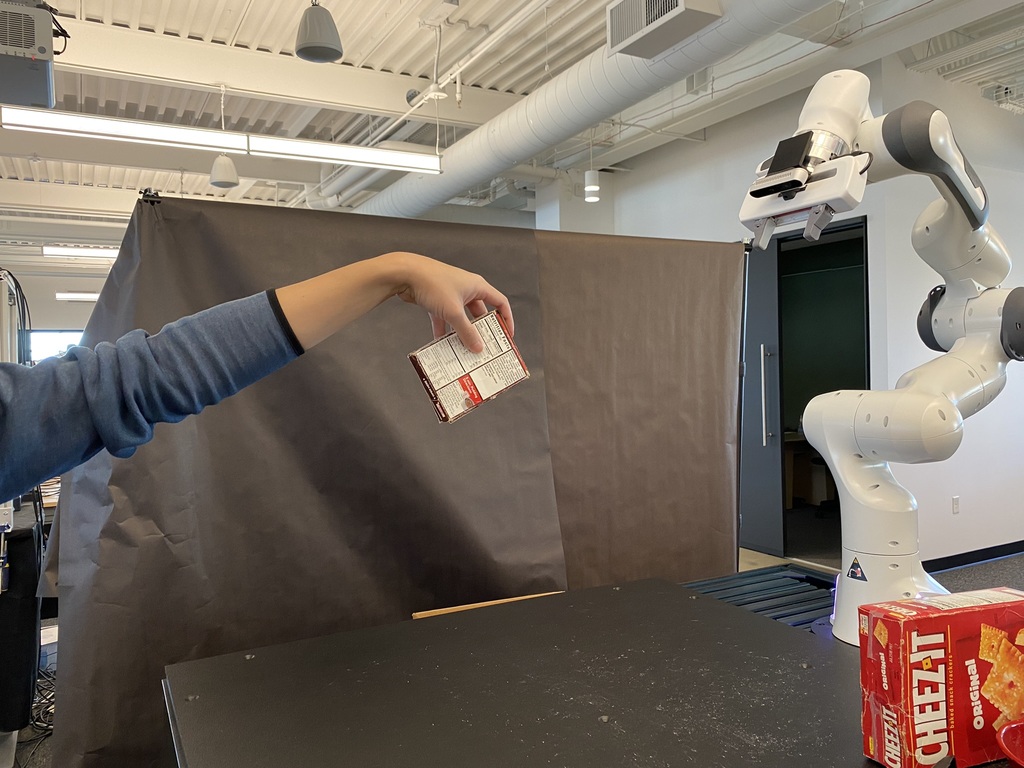}
   \\ \vspace{-1mm}
   pose 3
  \end{minipage}
  \\ \vspace{2mm}
 \end{minipage}
 \\ \vspace{3mm}
 \begin{minipage}{0.495\linewidth}
  \centering
  010\_potted\_meat\_can
  \\~\\ \vspace{-3mm}
  \begin{minipage}{0.320\linewidth}
   \centering
   \includegraphics[width=\linewidth,trim={280 254 430 200},clip]{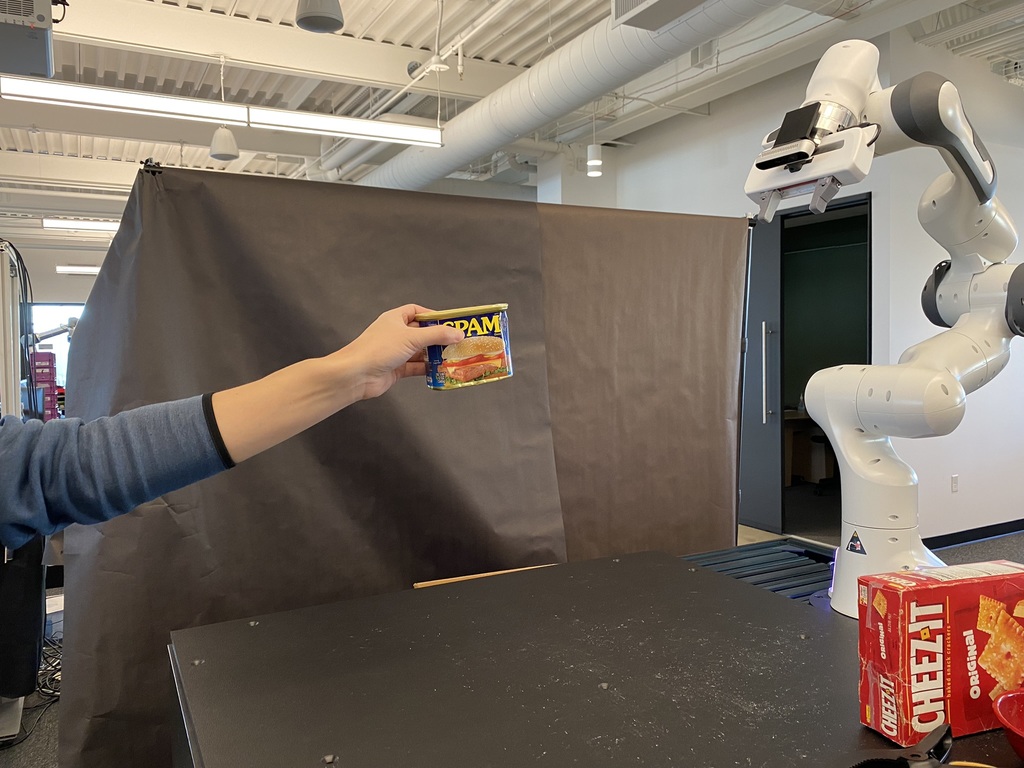}
   \\ \vspace{-1mm}
   pose 1
  \end{minipage}~
  \begin{minipage}{0.320\linewidth}
   \centering
   \includegraphics[width=\linewidth,trim={300 274 410 180},clip]{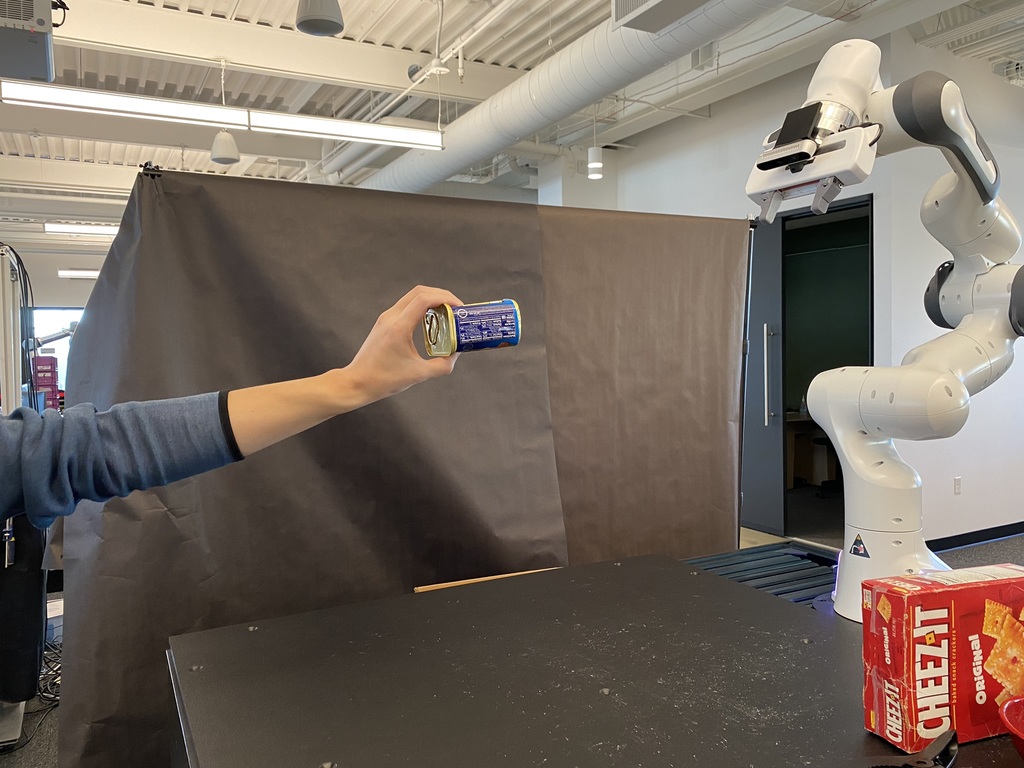}
   \\ \vspace{-1mm}
   pose 2
  \end{minipage}~
  \begin{minipage}{0.320\linewidth}
   \centering
   \includegraphics[width=\linewidth,trim={320 274 390 180},clip]{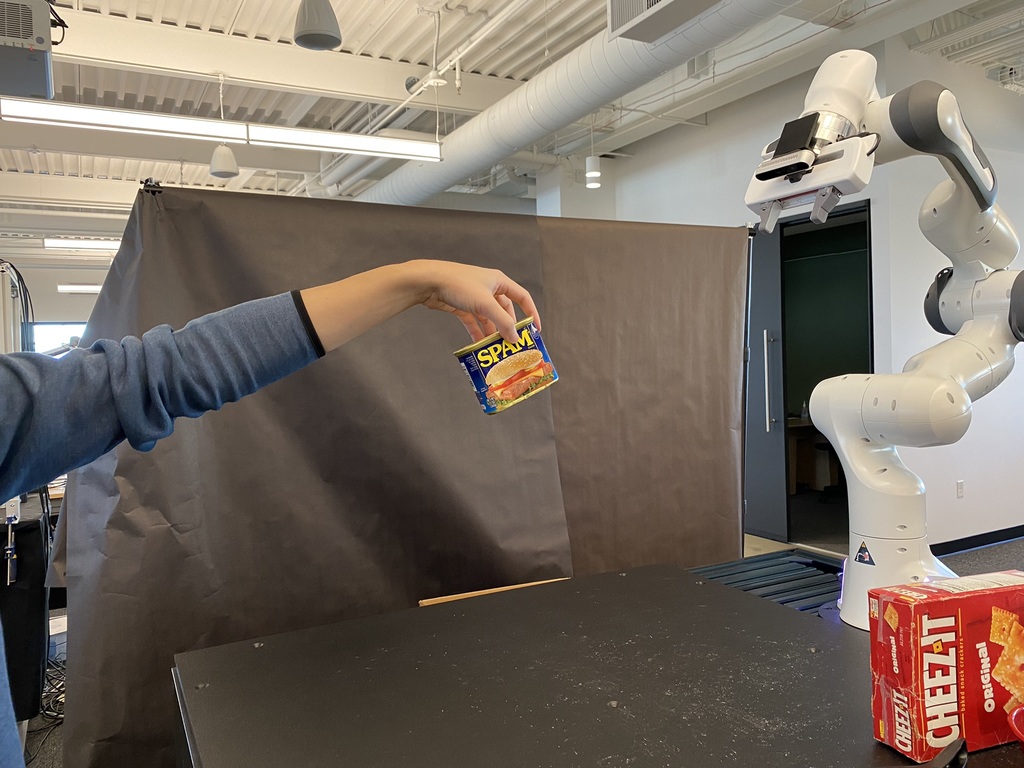}
   \\ \vspace{-1mm}
   pose 3
  \end{minipage}
  \\ \vspace{2mm}
 \end{minipage}~
 \begin{minipage}{0.495\linewidth}
  \centering
  021\_bleach\_cleanser
  \\~\\ \vspace{-3mm}
  \begin{minipage}{0.320\linewidth}
   \centering
   \includegraphics[width=\linewidth,trim={300 254 410 200},clip]{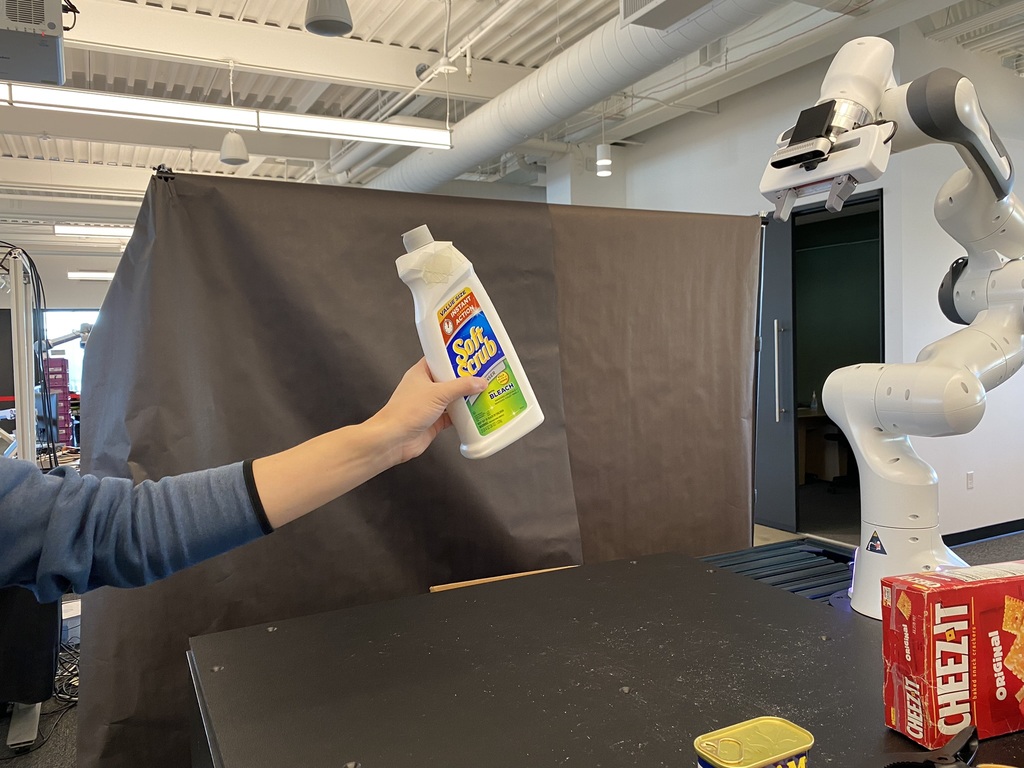}
   \\ \vspace{-1mm}
   pose 1
  \end{minipage}~
  \begin{minipage}{0.320\linewidth}
   \centering
   \includegraphics[width=\linewidth,trim={320 294 390 160},clip]{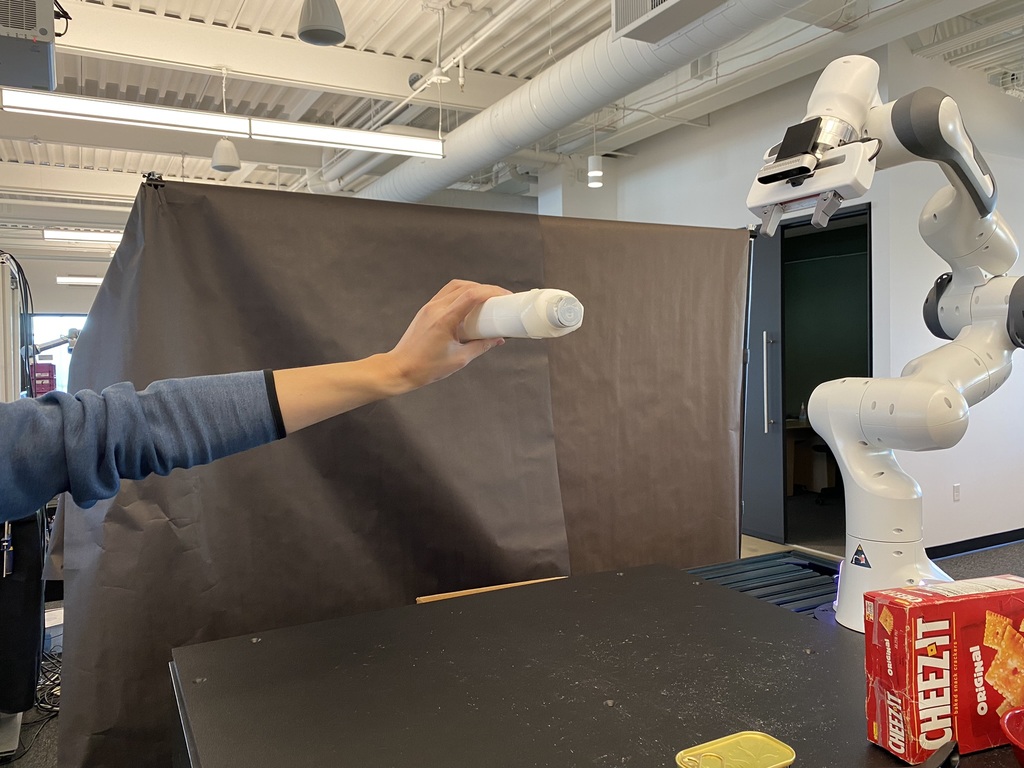}
   \\ \vspace{-1mm}
   pose 2
  \end{minipage}~
  \begin{minipage}{0.320\linewidth}
   \centering
   \includegraphics[width=\linewidth,trim={320 274 390 180},clip]{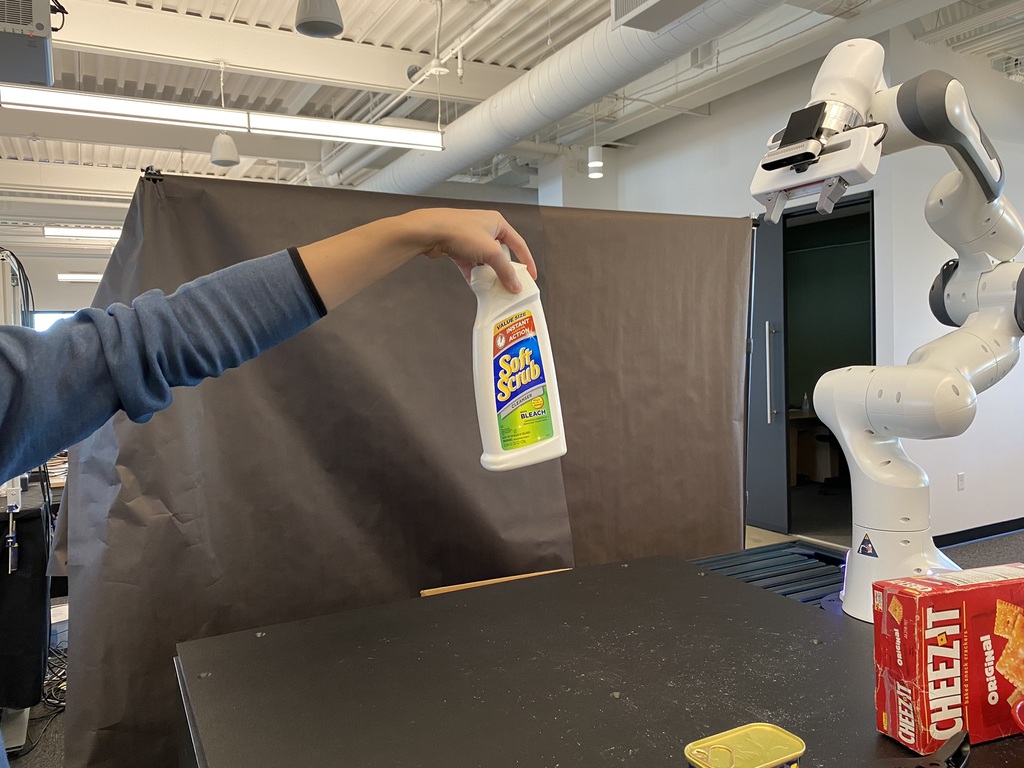}
   \\ \vspace{-1mm}
   pose 3
  \end{minipage}
  \\ \vspace{2mm}
 \end{minipage}
 \caption{\small The instructed handover poses for the \textbf{left hand} in the pilot study. Similar to the right hand poses in \Fig{pilot_right}, we adopt the same 10 objects from the YCB-Video dataset~\cite{xiang:rss2018} and pre-select 3 handover poses per object, totaling 30 handover poses for the left hand.}
 \label{fig:pilot_left}
\end{figure*}

\vspace{-3mm}
\paragraph{Evaluation Protocol}~First, we select the following 10 objects from the YCB-Video dataset~\cite{xiang:rss2018}:
\begin{multicols}{2}
 \begin{itemize}[noitemsep,topsep=0pt,parsep=0pt,partopsep=0pt]
  \item 011\_banana
  \item 037\_scissors
  \item 006\_mustard\_bottle
  \item 024\_bowl
  \item 040\_large\_marker
  \item 003\_cracker\_box
  \item 052\_extra\_large\_clamp
  \item 008\_pudding\_box
  \item 010\_potted\_meat\_can
  \item 021\_bleach\_cleanser
 \end{itemize}
\end{multicols}
\noindent For each object, we select 3 handover poses separately for the right and left hand, totaling 60 handover poses for both hands. The set of handover poses is selected to represent the handover task at different levels of difficulty: for each hand-object combination, we select one common handover pose (``pose 1''), one handover pose with the object held horizontally (``pose 2''), and one handover pose with severe hand occlusion by holding the object from the top (``pose 3''). \cref{fig:pilot} illustrates the setting where a subject holds an object in a controlled handover pose in front of the robot. \cref{fig:pilot_right,fig:pilot_left} show the selected handover poses for the right and left hand respectively.

For each subject, we iterate through the 60 handover poses and evaluate each pose once. A handover is considered failed if (1) the robot pinches (or is about to pinch) the subject's hand (in which case the subject may evade the grasp), (2) the robot drops the object during the handover, or (3) the robot has reached an irrecoverable state (e.g., a locked arm due to joint limits). A handover is successful if the robot retrieves the object from the subject's hand and successfully move it to the drop-off area without incurring any failures. We evaluate the same handover poses on two methods: GA-DDPG~\cite{wang:corl2021b} and ours. Therefore, each subject will perform 120 handover trials in total.

\vspace{-3mm}
\paragraph{Results}~We conduct our pilot study with two subjects. \cref{tab:pilot_subject_1,tab:pilot_subject_2} present the results of subject 1 and 2 respectively. For each pose, we report whether the handover is successful (``succ.''), and if so, the completion time of the handover (``time''), defined as the time span from the robot starts moving to the moment where the gripper lifts the object. For each object, we also report the mean success rate and mean completion time (see the ``mean'' column). For each subject, we present the results separately for the right hand (top), left hand (middle), and overall (bottom).

We observe a gap in the performance between the three selected handover poses. Both methods have a lower performance on ``pose 2'' and ``pose 3'' compared to ``pose 1''. For example, for subject 1, the overall success rate is $2/20$ for ``pose 2'' versus $16/20$ for ``pose 1'' for GA-DDPG~\cite{wang:corl2021b}, and $9/20$ for ``pose 2'' versus $19/20$ for ``pose 1'' for ours. This demonstrates the increase in challenges when the handover is conducted in uncommon poses (``pose 2''), where the robot needs to rotate the end effector, or when the human hand is blocking the robot's closest grasp point (``pose 3''), where the robot needs to diverge to avoid hand collision. However, our method is able to handle these cases better since the model is trained with diverse handover poses and supervision on hand collision avoidance (e.g., for ``pose 3'' on subject 1, an overall success rate of $13/20$ for ours versus $3/20$ for GA-DDPG~\cite{wang:corl2021b}). In addition, our method achieves a higher overall success rate on both subjects (i.e., $41/60$ versus $21/60$ on subject 1 and $41/60$ versus $33/60$ on subject 2), demonstrating its efficacy over GA-DDPG~\cite{wang:corl2021b}. However, our method still fails for $19$ trials on each subject. This can be attributed to either the sim-to-real gaps discussed in \cref{sec:setup}, an inherent failure of the policy on handling these cases, or an interplay of both.

\begin{table*}[t!]
 \centering
 \footnotesize
 \setlength{\tabcolsep}{1.8pt}
 \begin{tabular}{l||C{0.78cm}C{0.78cm}|C{0.78cm}C{0.78cm}|C{0.78cm}C{0.78cm}|C{0.78cm}C{0.78cm}||C{0.78cm}C{0.78cm}|C{0.78cm}C{0.78cm}|C{0.78cm}C{0.78cm}|C{0.78cm}C{0.78cm}}
                           & \multicolumn{16}{c}{subject 1: right hand}                                                                                                                                                                                                                        \\
                           & \multicolumn{8}{c||}{GA-DDPG~\cite{wang:corl2021b}}                                                                             & \multicolumn{8}{c}{Ours}                                                                                                        \\
                           & \multicolumn{2}{c|}{pose 1} & \multicolumn{2}{c|}{pose 2} & \multicolumn{2}{c|}{pose 3} & \multicolumn{2}{c||}{mean}            & \multicolumn{2}{c|}{pose 1} & \multicolumn{2}{c|}{pose 2} & \multicolumn{2}{c|}{pose 3} & \multicolumn{2}{c}{mean}              \\
                           & succ.       & time          & succ.       & time          & succ.       & time          & succ.              & time             & succ.       & time          & succ.       & time          & succ.       & time          & succ.              & time             \\
  \hline
  011\_banana              & \greencheck & 11.313        & \greencheck & 10.790        & \redcross   & --            & ~~2 / ~~3          & \textbf{11.051}  & \greencheck & ~~8.889       & \greencheck & 18.535        & \greencheck & ~~9.292       & \textbf{~~3 / ~~3} & 12.239           \\
  037\_scissors            & \greencheck & ~~9.571       & \redcross   & --            & \redcross   & --            & ~~1 / ~~3          & ~~9.571          & \greencheck & ~~8.863       & \redcross   & --            & \greencheck & 10.216        & \textbf{~~2 / ~~3} & \textbf{~~9.539} \\
  006\_mustard\_bottle     & \redcross   & --            & \redcross   & --            & \redcross   & --            & ~~0 / ~~3          & --               & \greencheck & ~~9.310       & \redcross   & --            & \greencheck & 10.792        & \textbf{~~2 / ~~3} & 10.051           \\
  024\_bowl                & \greencheck & ~~9.719       & \redcross   & --            & \greencheck & 10.760        & \textbf{~~2 / ~~3} & \textbf{10.239}  & \greencheck & 10.634        & \redcross   & --            & \greencheck & 13.333        & \textbf{~~2 / ~~3} & 11.983           \\
  040\_large\_marker       & \redcross   & --            & \redcross   & --            & \redcross   & --            & ~~0 / ~~3          & --               & \greencheck & ~~9.605       & \redcross   & --            & \redcross   & --            & \textbf{~~1 / ~~3} & ~~9.605          \\
  003\_cracker\_box        & \greencheck & ~~9.284       & \redcross   & --            & \greencheck & 10.440        & \textbf{~~2 / ~~3} & \textbf{~~9.862} & \greencheck & 16.782        & \redcross   & --            & \redcross   & --            & ~~1 / ~~3          & 16.782           \\
  052\_extra\_large\_clamp & \redcross   & --            & \redcross   & --            & \redcross   & --            & ~~0 / ~~3          & --               & \greencheck & 10.228        & \greencheck & 19.855        & \greencheck & 10.796        & \textbf{~~3 / ~~3} & 13.626           \\
  008\_pudding\_box        & \greencheck & ~~9.565       & \redcross   & --            & \greencheck & ~~9.095       & ~~2 / ~~3          & \textbf{~~9.330} & \greencheck & 10.583        & \greencheck & 11.833        & \greencheck & 11.387        & \textbf{~~3 / ~~3} & 11.267           \\
  010\_potted\_meat\_can   & \greencheck & ~~9.770       & \redcross   & --            & \redcross   & --            & ~~1 / ~~3          & \textbf{~~9.770} & \greencheck & 13.500        & \redcross   & --            & \greencheck & 10.908        & \textbf{~~2 / ~~3} & 12.204           \\
  021\_bleach\_cleanser    & \greencheck & ~~9.709       & \greencheck & 12.367        & \redcross   & --            & ~~2 / ~~3          & \textbf{11.038}  & \greencheck & 11.016        & \greencheck & 11.572        & \greencheck & 11.539        & \textbf{~~3 / ~~3} & 11.376           \\
  \hline
  total                    & ~~7 / 10    & ~~9.847       & ~~2 / 10    & 11.579        & ~~3 / 10    & 10.098        & 12 / 30            & \textbf{10.199}  & 10 / 10     & 10.941        & ~~4 / 10    & 15.449        & ~~8 / 10    & 11.033        & \textbf{22 / 30}   & 11.794           \\
 \end{tabular}
 \\ \vspace{4mm}
 \begin{tabular}{l||C{0.78cm}C{0.78cm}|C{0.78cm}C{0.78cm}|C{0.78cm}C{0.78cm}|C{0.78cm}C{0.78cm}||C{0.78cm}C{0.78cm}|C{0.78cm}C{0.78cm}|C{0.78cm}C{0.78cm}|C{0.78cm}C{0.78cm}}
                           & \multicolumn{16}{c}{subject 1: left hand}                                                                                                                                                                                                                         \\
                           & \multicolumn{8}{c||}{GA-DDPG~\cite{wang:corl2021b}}                                                                             & \multicolumn{8}{c}{Ours}                                                                                                        \\
                           & \multicolumn{2}{c|}{pose 1} & \multicolumn{2}{c|}{pose 2} & \multicolumn{2}{c|}{pose 3} & \multicolumn{2}{c||}{mean}            & \multicolumn{2}{c|}{pose 1} & \multicolumn{2}{c|}{pose 2} & \multicolumn{2}{c|}{pose 3} & \multicolumn{2}{c}{mean}              \\
                           & succ.       & time          & succ.       & time          & succ.       & time          & succ.              & time             & succ.       & time          & succ.       & time          & succ.       & time          & succ.              & time             \\
  \hline
  011\_banana              & \greencheck & ~~9.790       & \redcross   & --            & \redcross   & --            & ~~1 / ~~3          & ~~9.790          & \greencheck & ~~9.180       & \greencheck & ~~9.192       & \greencheck & ~~8.931       & \textbf{~~3 / ~~3} & \textbf{~~9.101} \\
  037\_scissors            & \greencheck & ~~9.549       & \redcross   & --            & \redcross   & --            & ~~1 / ~~3          & \textbf{~~9.549} & \greencheck & ~~8.973       & \greencheck & ~~9.485       & \greencheck & 10.254        & \textbf{~~3 / ~~3} & ~~9.571          \\
  006\_mustard\_bottle     & \greencheck & 10.135        & \redcross   & --            & \redcross   & --            & \textbf{~~1 / ~~3} & 10.135           & \greencheck & ~~8.499       & \redcross   & --            & \redcross   & --            & \textbf{~~1 / ~~3} & \textbf{~~8.499} \\
  024\_bowl                & \greencheck & 10.062        & \redcross   & --            & \redcross   & --            & ~~1 / ~~3          & \textbf{10.062}  & \greencheck & 14.572        & \redcross   & --            & \greencheck & 10.016        & \textbf{~~2 / ~~3} & 12.294           \\
  040\_large\_marker       & \redcross   & --            & \redcross   & --            & \redcross   & --            & ~~0 / ~~3          & --               & \greencheck & 15.736        & \greencheck & 14.601        & \greencheck & 10.626        & \textbf{~~3 / ~~3} & 13.654           \\
  003\_cracker\_box        & \greencheck & 10.231        & \redcross   & --            & \redcross   & --            & \textbf{~~1 / ~~3} & \textbf{10.231}  & \greencheck & 14.836        & \redcross   & --            & \redcross   & --            & \textbf{~~1 / ~~3} & 14.836           \\
  052\_extra\_large\_clamp & \greencheck & ~~8.513       & \redcross   & --            & \redcross   & --            & \textbf{~~1 / ~~3} & \textbf{~~8.513} & \greencheck & 10.186        & \redcross   & --            & \redcross   & --            & \textbf{~~1 / ~~3} & 10.186           \\
  008\_pudding\_box        & \greencheck & 10.851        & \redcross   & --            & \redcross   & --            & ~~1 / ~~3          & \textbf{10.851}  & \greencheck & 17.965        & \greencheck & 13.015        & \greencheck & 20.966        & \textbf{~~3 / ~~3} & 17.315           \\
  010\_potted\_meat\_can   & \greencheck & ~~9.810       & \redcross   & --            & \redcross   & --            & ~~1 / ~~3          & ~~9.810          & \redcross   & --            & \redcross   & --            & \redcross   & --            & ~~0 / ~~3          & --               \\
  021\_bleach\_cleanser    & \greencheck & 24.872        & \redcross   & --            & \redcross   & --            & ~~1 / ~~3          & 24.872           & \greencheck & 15.028        & \greencheck & ~~9.378       & \redcross   & --            & \textbf{~~2 / ~~3} & \textbf{12.203}  \\
  \hline
  total                    & ~~9 / 10    & 11.535        & ~~0 / 10    & --            & ~~0 / 10    & --            & ~~9 / 30           & \textbf{11.535}  & ~~9 / 10    & 12.775        & ~~5 / 10    & 11.134        & ~~5 / 10    & 12.159        & \textbf{19 / 30}   & 12.181           \\
 \end{tabular}
 \\ \vspace{4mm}
 \begin{tabular}{l||C{0.78cm}C{0.78cm}|C{0.78cm}C{0.78cm}|C{0.78cm}C{0.78cm}|C{0.78cm}C{0.78cm}||C{0.78cm}C{0.78cm}|C{0.78cm}C{0.78cm}|C{0.78cm}C{0.78cm}|C{0.78cm}C{0.78cm}}
                           & \multicolumn{16}{c}{subject 1: overall}                                                                                                                                                                                                                           \\
                           & \multicolumn{8}{c||}{GA-DDPG~\cite{wang:corl2021b}}                                                                             & \multicolumn{8}{c}{Ours}                                                                                                        \\
                           & \multicolumn{2}{c|}{pose 1} & \multicolumn{2}{c|}{pose 2} & \multicolumn{2}{c|}{pose 3} & \multicolumn{2}{c||}{mean}            & \multicolumn{2}{c|}{pose 1} & \multicolumn{2}{c|}{pose 2} & \multicolumn{2}{c|}{pose 3} & \multicolumn{2}{c}{mean}              \\
                           & succ.       & time          & succ.       & time          & succ.       & time          & succ.              & time             & succ.       & time          & succ.       & time          & succ.       & time          & succ.              & time             \\
  \hline
  011\_banana              & ~~2 / ~~2   & 10.551        & ~~1 / ~~2   & 10.282        & ~~0 / ~~2   & --            & ~~3 / ~~6          & \textbf{10.631}  & ~~2 / ~~2   & ~~9.034       & ~~2 / ~~2   & 13.864        & ~~2 / ~~2   & ~~9.111       & \textbf{~~6 / ~~6} & 10.670           \\
  037\_scissors            & ~~2 / ~~2   & ~~9.560       & ~~0 / ~~2   & --            & ~~0 / ~~2   & --            & ~~2 / ~~6          & ~~9.560          & ~~2 / ~~2   & ~~8.918       & ~~1 / ~~2   & 10.292        & ~~2 / ~~2   & 10.235        & \textbf{~~5 / ~~6} & \textbf{~~9.558} \\
  006\_mustard\_bottle     & ~~1 / ~~2   & 10.064        & ~~0 / ~~2   & --            & ~~0 / ~~2   & --            & ~~1 / ~~6          & 10.135           & ~~2 / ~~2   & ~~8.904       & ~~0 / ~~2   & --            & ~~1 / ~~2   & 10.038        & \textbf{~~3 / ~~6} & \textbf{~~9.534} \\
  024\_bowl                & ~~2 / ~~2   & ~~9.890       & ~~0 / ~~2   & --            & ~~1 / ~~2   & 10.076        & ~~3 / ~~6          & \textbf{10.180}  & ~~2 / ~~2   & 12.603        & ~~0 / ~~2   & --            & ~~2 / ~~2   & 11.675        & \textbf{~~4 / ~~6} & 12.139           \\
  040\_large\_marker       & ~~0 / ~~2   & --            & ~~0 / ~~2   & --            & ~~0 / ~~2   & --            & ~~0 / ~~6          & --               & ~~2 / ~~2   & 12.670        & ~~1 / ~~2   & 14.268        & ~~1 / ~~2   & 10.481        & \textbf{~~4 / ~~6} & 12.642           \\
  003\_cracker\_box        & ~~2 / ~~2   & ~~9.757       & ~~0 / ~~2   & --            & ~~1 / ~~2   & 10.021        & \textbf{~~3 / ~~6} & \textbf{~~9.985} & ~~2 / ~~2   & 15.809        & ~~0 / ~~2   & --            & ~~0 / ~~2   & --            & \textbf{~~3 / ~~6} & 15.809           \\
  052\_extra\_large\_clamp & ~~1 / ~~2   & ~~8.874       & ~~0 / ~~2   & --            & ~~0 / ~~2   & --            & ~~1 / ~~6          & \textbf{~~8.513} & ~~2 / ~~2   & 10.207        & ~~1 / ~~2   & 15.824        & ~~1 / ~~2   & 16.119        & \textbf{~~4 / ~~6} & 12.766           \\
  008\_pudding\_box        & ~~2 / ~~2   & 10.208        & ~~0 / ~~2   & --            & ~~1 / ~~2   & ~~9.579       & ~~3 / ~~6          & \textbf{~~9.837} & ~~2 / ~~2   & 14.274        & ~~2 / ~~2   & 12.424        & ~~2 / ~~2   & 16.176        & \textbf{~~6 / ~~6} & 14.291           \\
  010\_potted\_meat\_can   & ~~2 / ~~2   & ~~9.790       & ~~0 / ~~2   & --            & ~~0 / ~~2   & --            & \textbf{~~2 / ~~6} & \textbf{~~9.790} & ~~1 / ~~2   & 14.458        & ~~0 / ~~2   & --            & ~~1 / ~~2   & 12.596        & \textbf{~~2 / ~~6} & 12.204           \\
  021\_bleach\_cleanser    & ~~2 / ~~2   & 17.291        & ~~1 / ~~2   & 11.867        & ~~0 / ~~2   & --            & ~~3 / ~~6          & 15.650           & ~~2 / ~~2   & 13.022        & ~~2 / ~~2   & 10.475        & ~~1 / ~~2   & 16.574        & \textbf{~~5 / ~~6} & \textbf{11.707}  \\
  \hline
  total                    & 16 / 20     & 10.797        & ~~2 / 20    & 11.579        & ~~3 / 20    & 10.098        & 21 / 60            & \textbf{10.771}  & 19 / 20     & 11.810        & ~~9 / 20    & 13.052        & 13 / 20     & 11.466        & \textbf{41 / 60}   & 11.973           \\
 \end{tabular}
 \caption{\small The pilot study results of \textbf{subject 1}. We present the results separately for the right-hand handovers (top), the left-hand handovers (middle), and overall (bottom). We report both the success rate (``succ.'') and the completion time (``time''). Our method outperforms GA-DDPG~\cite{wang:corl2021b} in the success rate.}
 \label{tab:pilot_subject_1}
\end{table*}
\begin{table*}[t!]
 \centering
 \footnotesize
 \setlength{\tabcolsep}{1.8pt}
 \begin{tabular}{l||C{0.78cm}C{0.78cm}|C{0.78cm}C{0.78cm}|C{0.78cm}C{0.78cm}|C{0.78cm}C{0.78cm}||C{0.78cm}C{0.78cm}|C{0.78cm}C{0.78cm}|C{0.78cm}C{0.78cm}|C{0.78cm}C{0.78cm}}
                           & \multicolumn{16}{c}{subject 2: right hand}                                                                                                                                                                                                                         \\
                           & \multicolumn{8}{c||}{GA-DDPG~\cite{wang:corl2021b}}                                                                             & \multicolumn{8}{c}{Ours}                                                                                                        \\
                           & \multicolumn{2}{c|}{pose 1} & \multicolumn{2}{c|}{pose 2} & \multicolumn{2}{c|}{pose 3} & \multicolumn{2}{c||}{mean}            & \multicolumn{2}{c|}{pose 1} & \multicolumn{2}{c|}{pose 2} & \multicolumn{2}{c|}{pose 3} & \multicolumn{2}{c}{mean}              \\
                           & succ.       & time          & succ.       & time          & succ.       & time          & succ.              & time             & succ.       & time          & succ.       & time          & succ.       & time          & succ.              & time             \\
  \hline
  011\_banana              & \greencheck & 13.599        & \greencheck & 15.559        & \greencheck & ~~9.333       & \textbf{~~3 / ~~3} & 12.830           & \greencheck & 11.995        & \redcross   & --            & \greencheck & ~~8.046       & ~~2 / ~~3          & \textbf{10.021}  \\
  037\_scissors            & \greencheck & 10.609        & \redcross   & --            & \greencheck & ~~9.769       & \textbf{~~2 / ~~3} & 10.189           & \greencheck & ~~8.343       & \redcross   & --            & \greencheck & ~~8.583       & \textbf{~~2 / ~~3} & \textbf{~~8.463} \\
  006\_mustard\_bottle     & \greencheck & ~~9.070       & \redcross   & --            & \redcross   & --            & ~~1 / ~~3          & ~~9.070          & \greencheck & ~~8.787       & \redcross   & --            & \greencheck & ~~9.186       & \textbf{~~2 / ~~3} & \textbf{~~8.987} \\
  024\_bowl                & \redcross   & --            & \redcross   & --            & \greencheck & ~~9.085       & \textbf{~~1 / ~~3} & \textbf{~~9.085} & \redcross   & --            & \redcross   & --            & \greencheck & ~~9.506       & \textbf{~~1 / ~~3} & ~~9.506          \\
  040\_large\_marker       & \greencheck & 10.762        & \redcross   & --            & \greencheck & ~~8.748       & \textbf{~~2 / ~~3} & \textbf{~~9.755} & \greencheck & ~~9.035       & \redcross   & --            & \greencheck & 17.609        & \textbf{~~2 / ~~3} & 13.322           \\
  003\_cracker\_box        & \redcross   & --            & \redcross   & --            & \redcross   & --            & ~~0 / ~~3          & --               & \greencheck & 11.629        & \redcross   & --            & \greencheck & 11.563        & \textbf{~~2 / ~~3} & 11.596           \\
  052\_extra\_large\_clamp & \greencheck & ~~9.319       & \redcross   & --            & \greencheck & ~~9.156       & \textbf{~~2 / ~~3} & \textbf{~~9.237} & \greencheck & ~~9.788       & \greencheck & 17.289        & \redcross   & --            & \textbf{~~2 / ~~3} & 13.539           \\
  008\_pudding\_box        & \greencheck & 10.402        & \redcross   & --            & \greencheck & ~~8.838       & \textbf{~~2 / ~~3} & \textbf{~~9.620} & \greencheck & ~~8.803       & \redcross   & --            & \greencheck & 11.061        & \textbf{~~2 / ~~3} & ~~9.932          \\
  010\_potted\_meat\_can   & \redcross   & --            & \redcross   & --            & \greencheck & 12.886        & ~~1 / ~~3          & 12.886           & \greencheck & ~~8.540       & \redcross   & --            & \greencheck & ~~8.610       & \textbf{~~2 / ~~3} & \textbf{~~8.575} \\
  021\_bleach\_cleanser    & \greencheck & 10.571        & \greencheck & 10.223        & \redcross   & --            & \textbf{~~2 / ~~3} & 10.397           & \greencheck & ~~8.241       & \redcross   & --            & \greencheck & 11.706        & \textbf{~~2 / ~~3} & \textbf{~~9.974} \\
  \hline
  total                    & ~~7 / 10    & 10.619        & ~~2 / 10    & 12.891        & ~~7 / 10    & ~~9.688       & 16 / 30            & 10.495           & ~~9 / 10    & ~~9.462       & ~~1 / 10    & 17.289        & ~~9 / 10    & 10.652        & \textbf{19 / 30}   & \textbf{10.438}  \\
 \end{tabular}
 \\ \vspace{4mm}
 \begin{tabular}{l||C{0.78cm}C{0.78cm}|C{0.78cm}C{0.78cm}|C{0.78cm}C{0.78cm}|C{0.78cm}C{0.78cm}||C{0.78cm}C{0.78cm}|C{0.78cm}C{0.78cm}|C{0.78cm}C{0.78cm}|C{0.78cm}C{0.78cm}}
                           & \multicolumn{16}{c}{subject 2: left hand}                                                                                                                                                                                                                         \\
                           & \multicolumn{8}{c||}{GA-DDPG~\cite{wang:corl2021b}}                                                                             & \multicolumn{8}{c}{Ours}                                                                                                        \\
                           & \multicolumn{2}{c|}{pose 1} & \multicolumn{2}{c|}{pose 2} & \multicolumn{2}{c|}{pose 3} & \multicolumn{2}{c||}{mean}            & \multicolumn{2}{c|}{pose 1} & \multicolumn{2}{c|}{pose 2} & \multicolumn{2}{c|}{pose 3} & \multicolumn{2}{c}{mean}              \\
                           & succ.       & time          & succ.       & time          & succ.       & time          & succ.              & time             & succ.       & time          & succ.       & time          & succ.       & time          & succ.              & time             \\
  \hline
  011\_banana              & \greencheck & ~~8.709       & \greencheck & ~~9.186       & \greencheck & ~~8.570       & \textbf{~~3 / ~~3} & \textbf{~~8.821} & \greencheck & ~~9.096       & \greencheck & 10.447        & \greencheck & ~~7.965       & \textbf{~~3 / ~~3} & ~~9.169          \\
  037\_scissors            & \greencheck & ~~9.849       & \redcross   & --            & \redcross   & --            & ~~1 / ~~3          & ~~9.849          & \greencheck & ~~8.587       & \greencheck & 10.570        & \greencheck & ~~8.370       & \textbf{~~3 / ~~3} & \textbf{~~9.176} \\
  006\_mustard\_bottle     & \redcross   & --            & \redcross   & --            & \greencheck & 11.332        & ~~1 / ~~3          & 11.332           & \greencheck & ~~8.078       & \redcross   & --            & \greencheck & ~~9.703       & \textbf{~~2 / ~~3} & \textbf{~~8.891} \\
  024\_bowl                & \greencheck & ~~9.183       & \redcross   & --            & \greencheck & ~~9.214       & \textbf{~~2 / ~~3} & \textbf{~~9.199} & \redcross   & --            & \greencheck & ~~9.958       & \greencheck & 17.879        & \textbf{~~2 / ~~3} & 13.918           \\
  040\_large\_marker       & \greencheck & 10.651        & \redcross   & --            & \greencheck & ~~9.881       & ~~2 / ~~3          & \textbf{10.266}  & \greencheck & ~~9.804       & \greencheck & 14.292        & \greencheck & ~~9.466       & \textbf{~~3 / ~~3} & 11.187           \\
  003\_cracker\_box        & \redcross   & --            & \redcross   & --            & \redcross   & --            & \textbf{~~0 / ~~3} & --               & \redcross   & --            & \redcross   & --            & \redcross   & --            & \textbf{~~0 / ~~3} & --               \\
  052\_extra\_large\_clamp & \greencheck & 19.748        & \greencheck & ~~9.566       & \greencheck & 10.204        & \textbf{~~3 / ~~3} & 13.173           & \greencheck & 19.253        & \greencheck & ~~9.632       & \greencheck & ~~9.552       & \textbf{~~3 / ~~3} & \textbf{12.813}  \\
  008\_pudding\_box        & \greencheck & ~~9.794       & \redcross   & --            & \greencheck & ~~9.236       & \textbf{~~2 / ~~3} & ~~9.515          & \greencheck & ~~8.532       & \redcross   & --            & \greencheck & ~~8.590       & \textbf{~~2 / ~~3} & \textbf{~~8.561} \\
  010\_potted\_meat\_can   & \greencheck & ~~9.353       & \redcross   & --            & \greencheck & ~~9.240       & \textbf{~~2 / ~~3} & ~~9.296          & \greencheck & ~~8.344       & \redcross   & --            & \greencheck & ~~9.277       & \textbf{~~2 / ~~3} & \textbf{~~8.810} \\
  021\_bleach\_cleanser    & \greencheck & 10.140        & \redcross   & --            & \redcross   & --            & ~~1 / ~~3          & 10.140           & \greencheck & ~~9.301       & \redcross   & --            & \greencheck & 10.288        & \textbf{~~2 / ~~3} & \textbf{~~9.795} \\
  \hline
  total                    & ~~8 / 10    & 10.928        & ~~2 / 10    & ~~9.376       & ~~7 / 10    & ~~9.668       & 17 / 30            & \textbf{10.227}  & ~~8 / 10    & 10.125        & ~~5 / 10    & 10.980        & ~~9 / 10    & 10.121        & \textbf{22 / 30}   & 10.318           \\
 \end{tabular}
 \\ \vspace{4mm}
 \begin{tabular}{l||C{0.78cm}C{0.78cm}|C{0.78cm}C{0.78cm}|C{0.78cm}C{0.78cm}|C{0.78cm}C{0.78cm}||C{0.78cm}C{0.78cm}|C{0.78cm}C{0.78cm}|C{0.78cm}C{0.78cm}|C{0.78cm}C{0.78cm}}
                           & \multicolumn{16}{c}{subject 2: overall}                                                                                                                                                                                                                           \\
                           & \multicolumn{8}{c||}{GA-DDPG~\cite{wang:corl2021b}}                                                                             & \multicolumn{8}{c}{Ours}                                                                                                        \\
                           & \multicolumn{2}{c|}{pose 1} & \multicolumn{2}{c|}{pose 2} & \multicolumn{2}{c|}{pose 3} & \multicolumn{2}{c||}{mean}            & \multicolumn{2}{c|}{pose 1} & \multicolumn{2}{c|}{pose 2} & \multicolumn{2}{c|}{pose 3} & \multicolumn{2}{c}{mean}              \\
                           & succ.       & time          & succ.       & time          & succ.       & time          & succ.              & time             & succ.       & time          & succ.       & time          & succ.       & time          & succ.              & time             \\
  \hline
  011\_banana              & ~~2 / ~~2   & 11.154        & ~~2 / ~~2   & 12.372        & ~~2 / ~~2   & ~~8.951       & \textbf{~~6 / ~~6} & 10.826           & ~~2 / ~~2   & 10.546        & ~~1 / ~~2   & 10.408        & ~~2 / ~~2   & ~~8.006       & ~~5 / ~~6          & \textbf{~~9.510} \\
  037\_scissors            & ~~2 / ~~2   & 10.229        & ~~0 / ~~2   & --            & ~~1 / ~~2   & ~~9.243       & ~~3 / ~~6          & 10.076           & ~~2 / ~~2   & ~~8.465       & ~~1 / ~~2   & ~~9.638       & ~~2 / ~~2   & ~~8.477       & \textbf{~~5 / ~~6} & \textbf{~~8.891} \\
  006\_mustard\_bottle     & ~~1 / ~~2   & ~~9.476       & ~~0 / ~~2   & --            & ~~1 / ~~2   & 10.479        & ~~2 / ~~6          & 10.201           & ~~2 / ~~2   & ~~8.433       & ~~0 / ~~2   & --            & ~~2 / ~~2   & ~~9.445       & \textbf{~~4 / ~~6} & \textbf{~~8.939} \\
  024\_bowl                & ~~1 / ~~2   & 11.101        & ~~0 / ~~2   & --            & ~~2 / ~~2   & ~~9.149       & \textbf{~~3 / ~~6} & \textbf{~~9.161} & ~~0 / ~~2   & --            & ~~1 / ~~2   & ~~9.720       & ~~2 / ~~2   & 13.692        & \textbf{~~3 / ~~6} & 12.447           \\
  040\_large\_marker       & ~~2 / ~~2   & 10.707        & ~~0 / ~~2   & --            & ~~2 / ~~2   & ~~9.314       & ~~4 / ~~6          & \textbf{10.010}  & ~~2 / ~~2   & ~~9.419       & ~~1 / ~~2   & 15.882        & ~~2 / ~~2   & 13.537        & \textbf{~~5 / ~~6} & 12.041           \\
  003\_cracker\_box        & ~~0 / ~~2   & --            & ~~0 / ~~2   & --            & ~~0 / ~~2   & --            & ~~0 / ~~6          & --               & ~~1 / ~~2   & 11.925        & ~~0 / ~~2   & --            & ~~1 / ~~2   & 12.989        & \textbf{~~2 / ~~6} & 11.596           \\
  052\_extra\_large\_clamp & ~~2 / ~~2   & 14.533        & ~~1 / ~~2   & ~~9.562       & ~~2 / ~~2   & ~~9.680       & \textbf{~~5 / ~~6} & \textbf{11.599}  & ~~2 / ~~2   & 14.521        & ~~2 / ~~2   & 13.460        & ~~1 / ~~2   & 36.778        & \textbf{~~5 / ~~6} & 13.103           \\
  008\_pudding\_box        & ~~2 / ~~2   & 10.098        & ~~0 / ~~2   & --            & ~~2 / ~~2   & ~~9.037       & \textbf{~~4 / ~~6} & ~~9.567          & ~~2 / ~~2   & ~~8.668       & ~~0 / ~~2   & --            & ~~2 / ~~2   & ~~9.825       & \textbf{~~4 / ~~6} & \textbf{~~9.247} \\
  010\_potted\_meat\_can   & ~~1 / ~~2   & ~~9.919       & ~~0 / ~~2   & --            & ~~2 / ~~2   & 11.063        & ~~3 / ~~6          & 10.493           & ~~2 / ~~2   & ~~8.442       & ~~0 / ~~2   & --            & ~~2 / ~~2   & ~~8.943       & \textbf{~~4 / ~~6} & \textbf{~~8.693} \\
  021\_bleach\_cleanser    & ~~2 / ~~2   & 10.355        & ~~1 / ~~2   & 12.233        & ~~0 / ~~2   & --            & ~~3 / ~~6          & 10.311           & ~~2 / ~~2   & ~~8.771       & ~~0 / ~~2   & --            & ~~2 / ~~2   & 10.997        & \textbf{~~4 / ~~6} & \textbf{~~9.884} \\
  \hline
  total                    & 15 / 20     & 10.784        & ~~4 / 20    & 11.134        & 14 / 20     & ~~9.678       & 33 / 60            & \textbf{10.357}  & 17 / 20     & ~~9.774       & ~~6 / 20    & 12.031        & 18 / 20     & 10.387        & \textbf{41 / 60}   & 10.373           \\
 \end{tabular} 
 \caption{\small The pilot study results of \textbf{subject 2}. We present the results separately for the right-hand handovers (top), the left-hand handovers (middle), and overall (bottom). We report both the success rate (``succ.'') and the completion time (``time''). Our method outperforms GA-DDPG~\cite{wang:corl2021b} in the success rate.}
 \label{tab:pilot_subject_2}
\end{table*}

\subsection{User Evaluation}

\begin{figure*}[t!]
 \centering
 \includegraphics[width=0.191\linewidth]{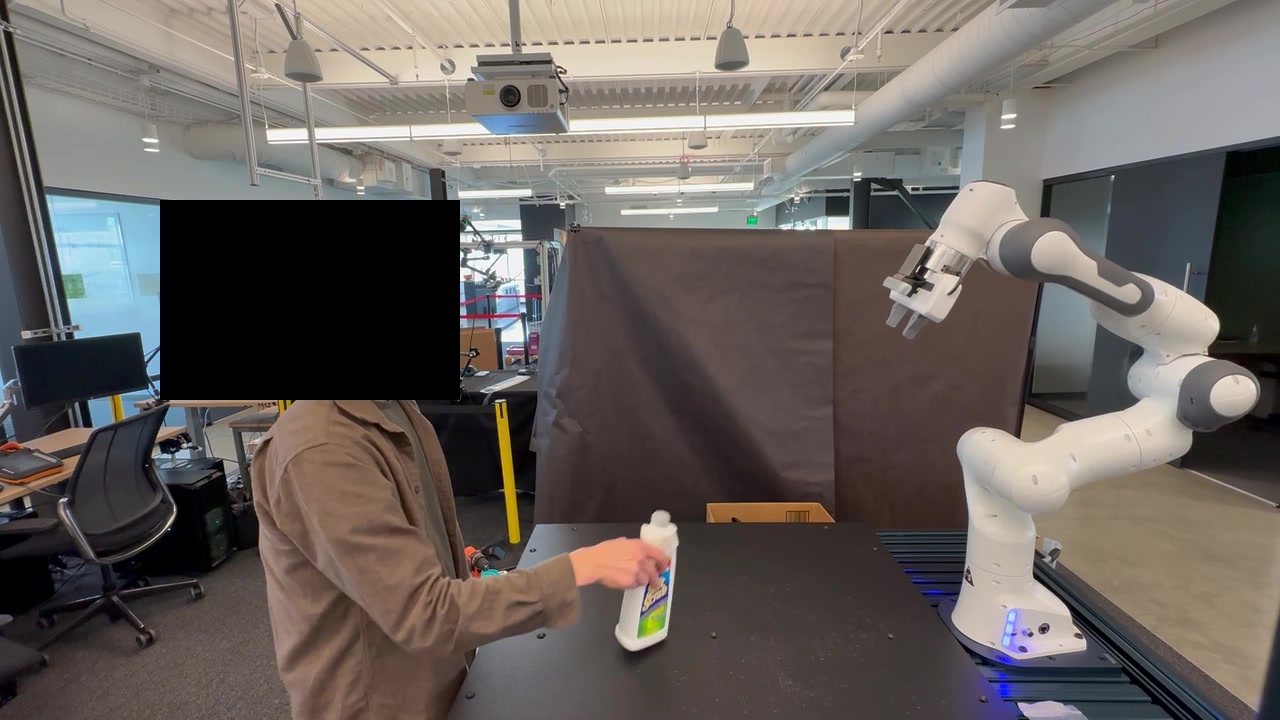}~
 \includegraphics[width=0.191\linewidth]{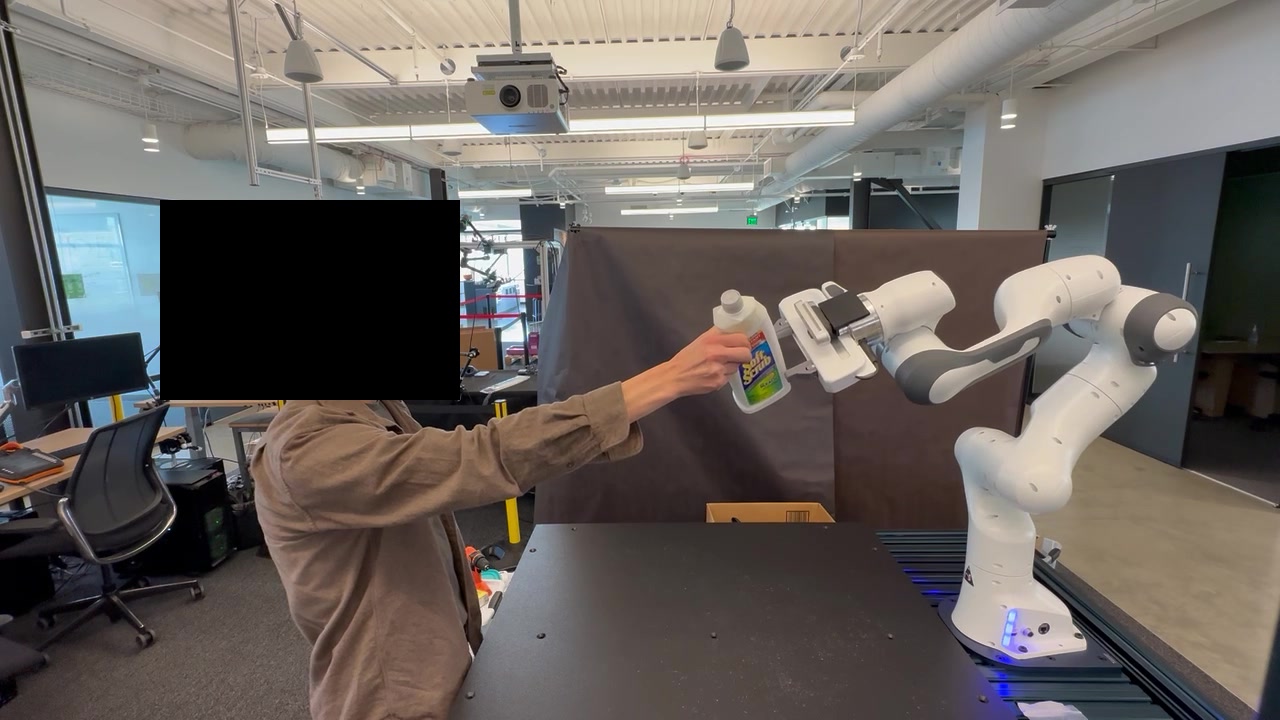}~
 \includegraphics[width=0.191\linewidth]{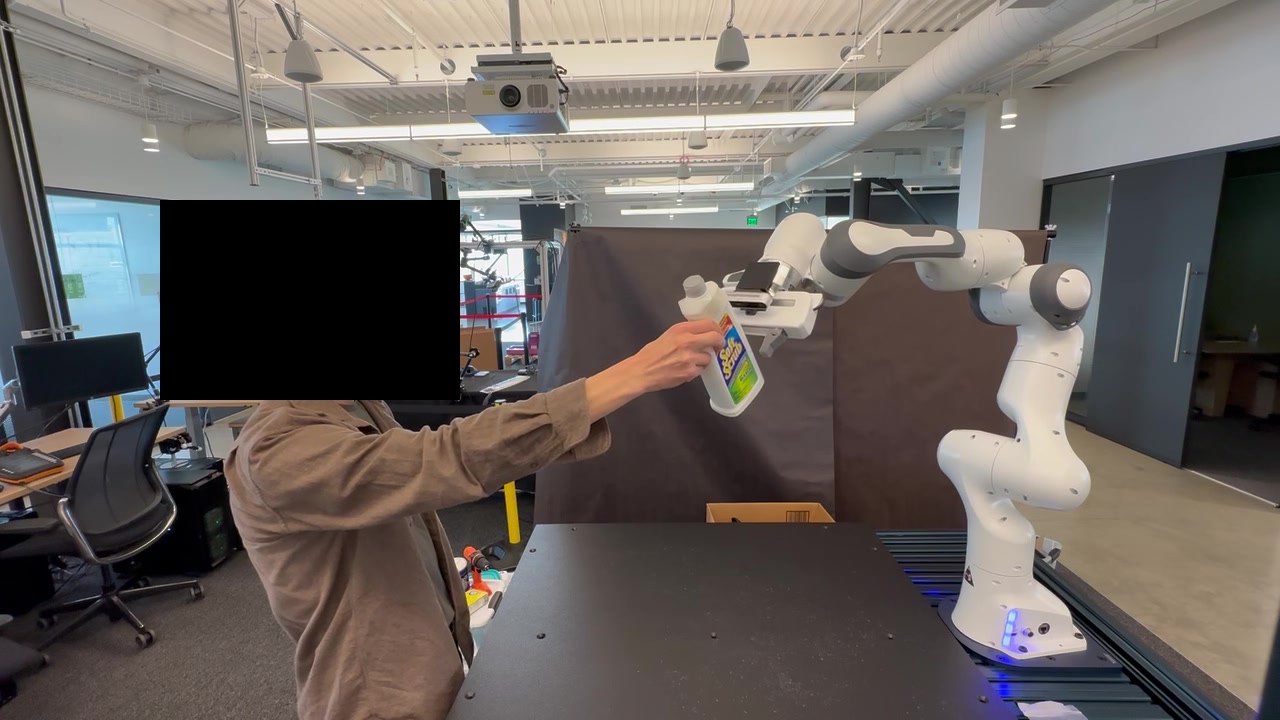}~
 \includegraphics[width=0.191\linewidth]{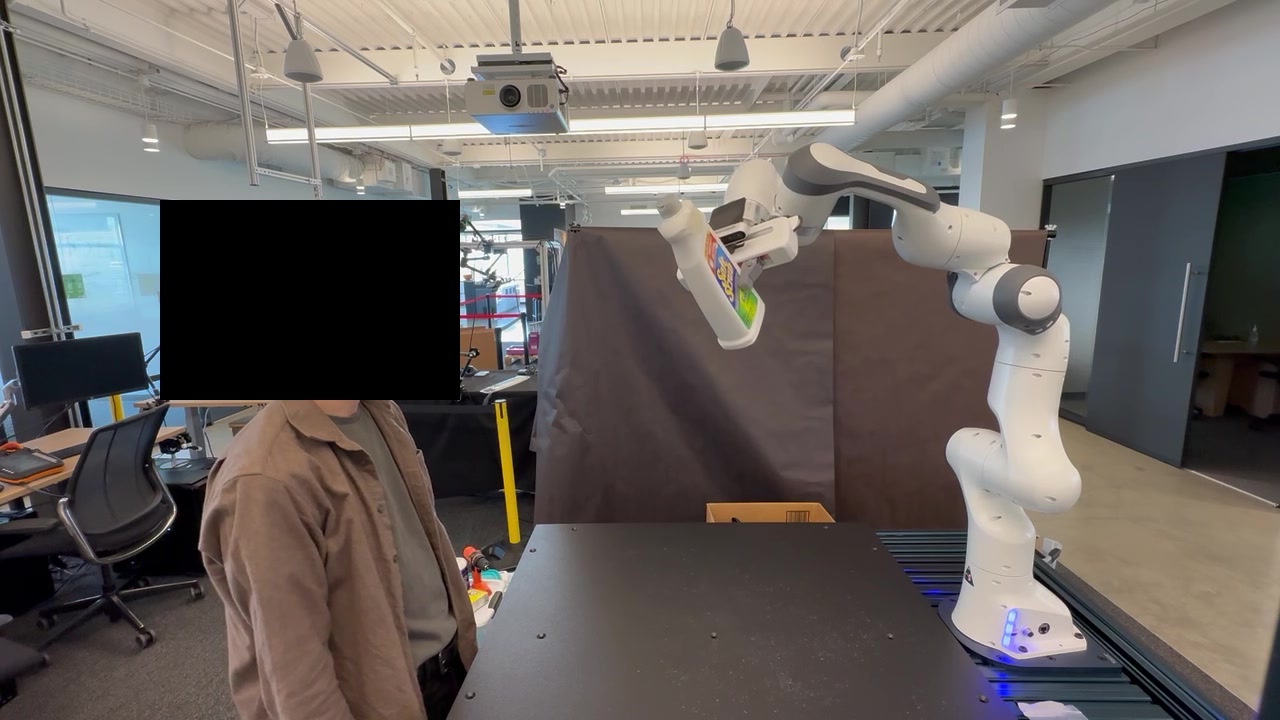}~
 \includegraphics[width=0.191\linewidth]{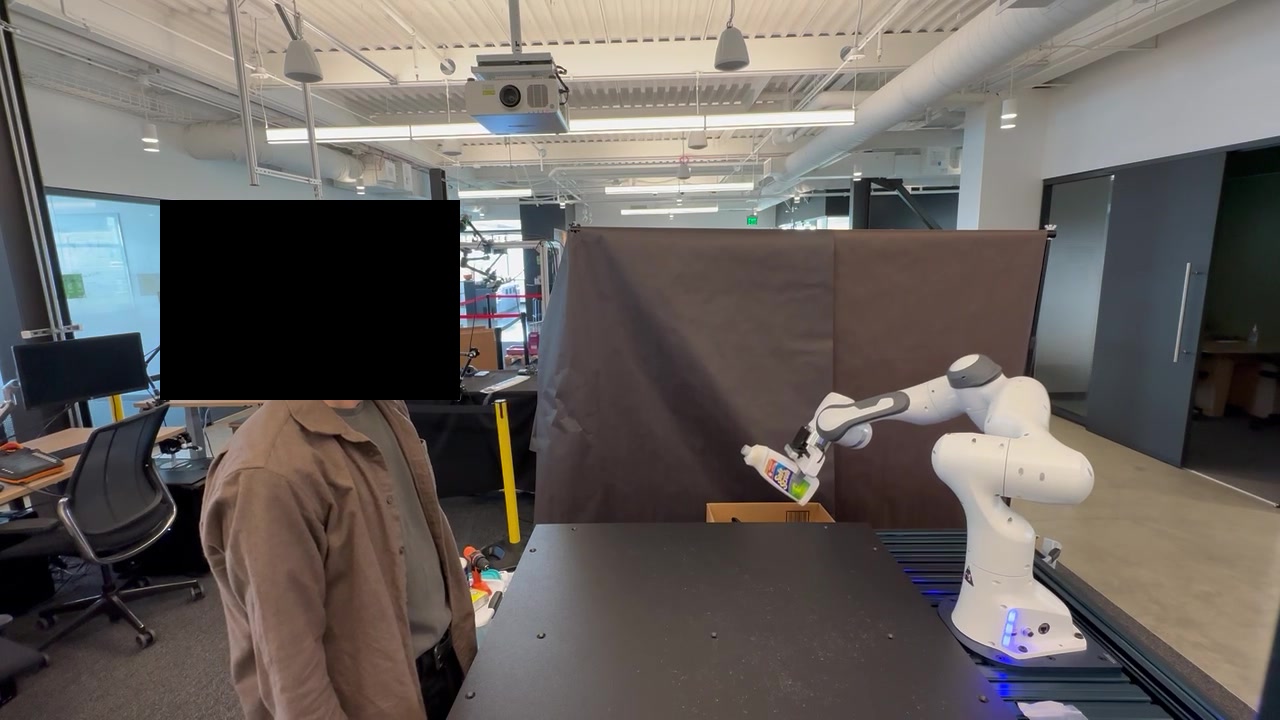}
 \\ \vspace{1mm}
 \includegraphics[width=0.191\linewidth]{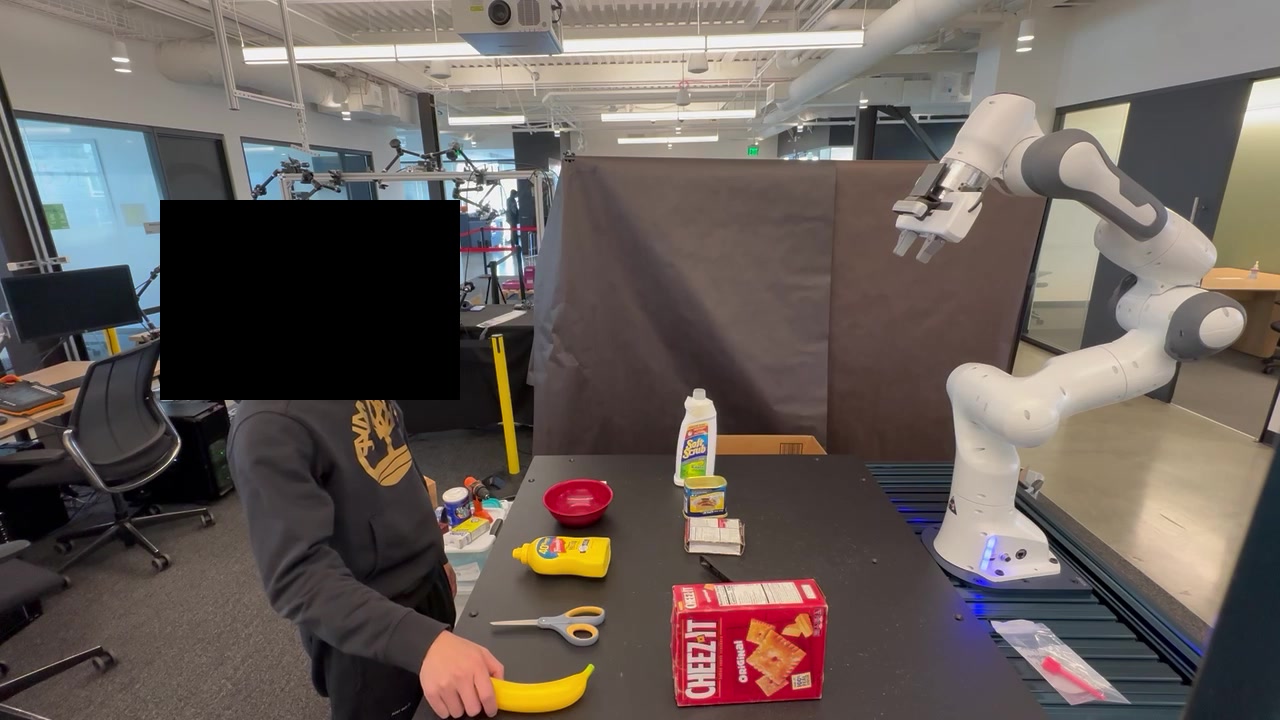}~
 \includegraphics[width=0.191\linewidth]{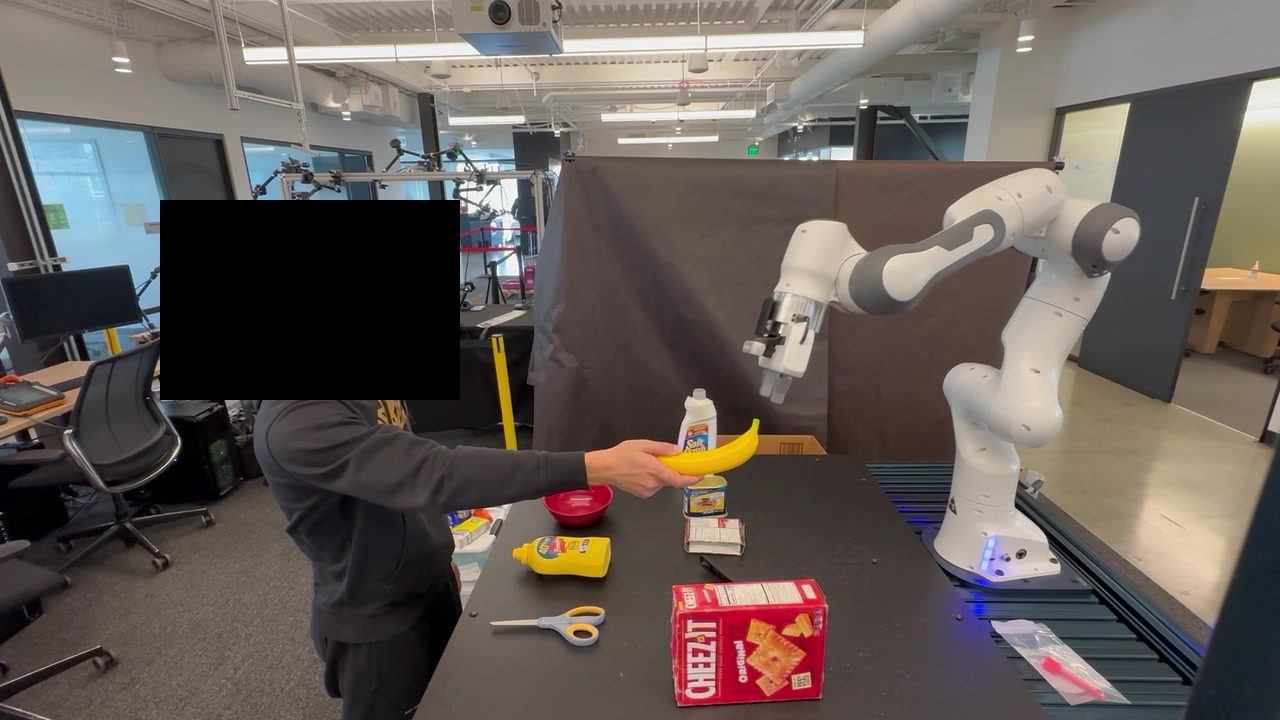}~
 \includegraphics[width=0.191\linewidth]{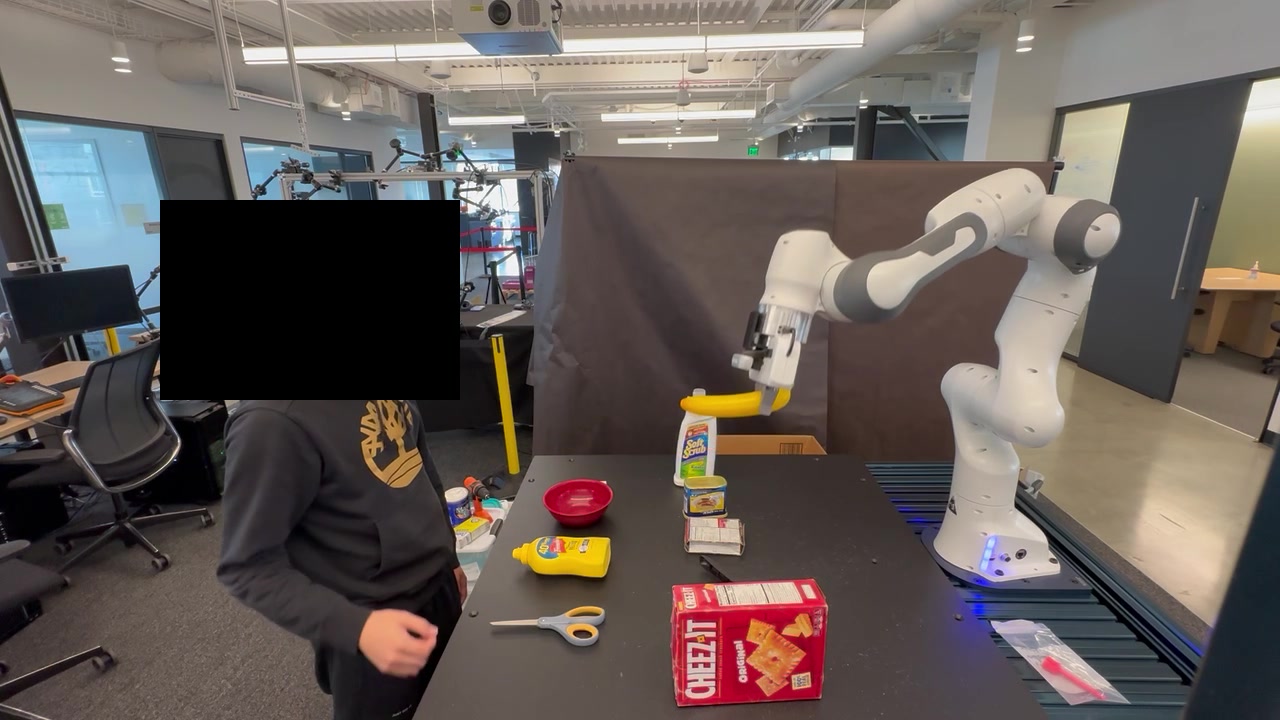}~
 \includegraphics[width=0.191\linewidth]{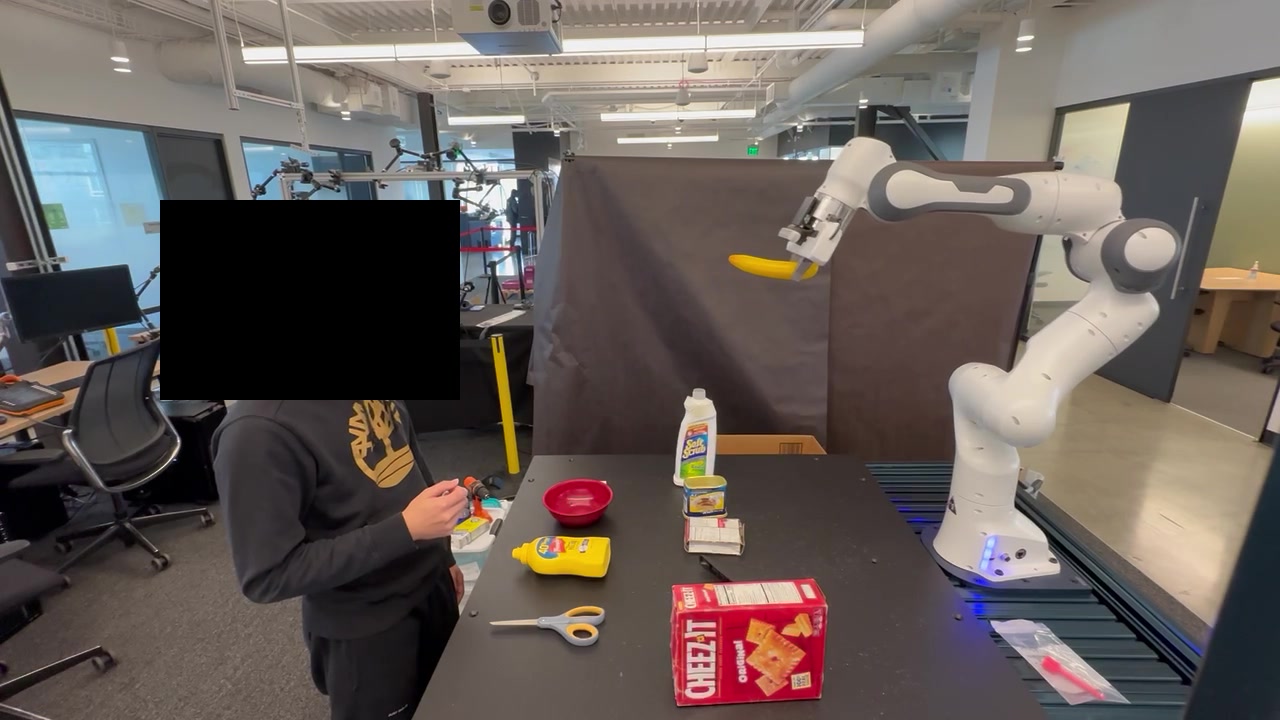}~
 \includegraphics[width=0.191\linewidth]{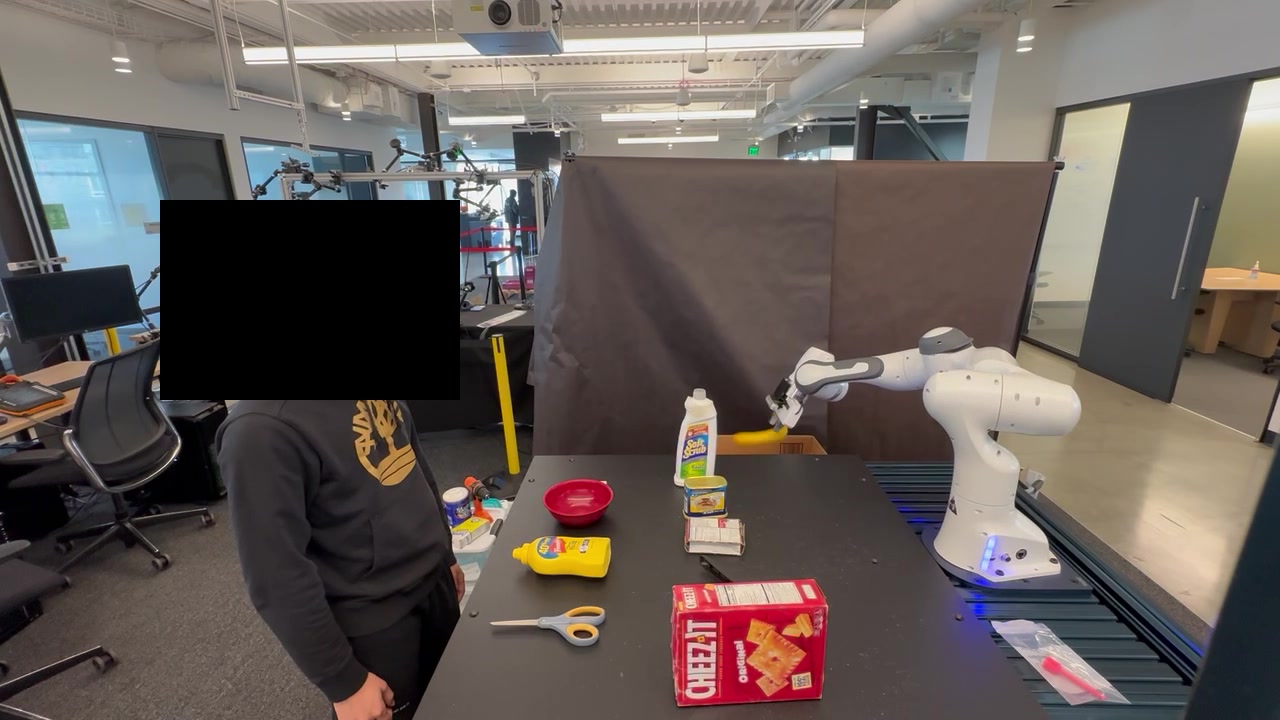}
 \\ \vspace{1mm}
 \includegraphics[width=0.191\linewidth]{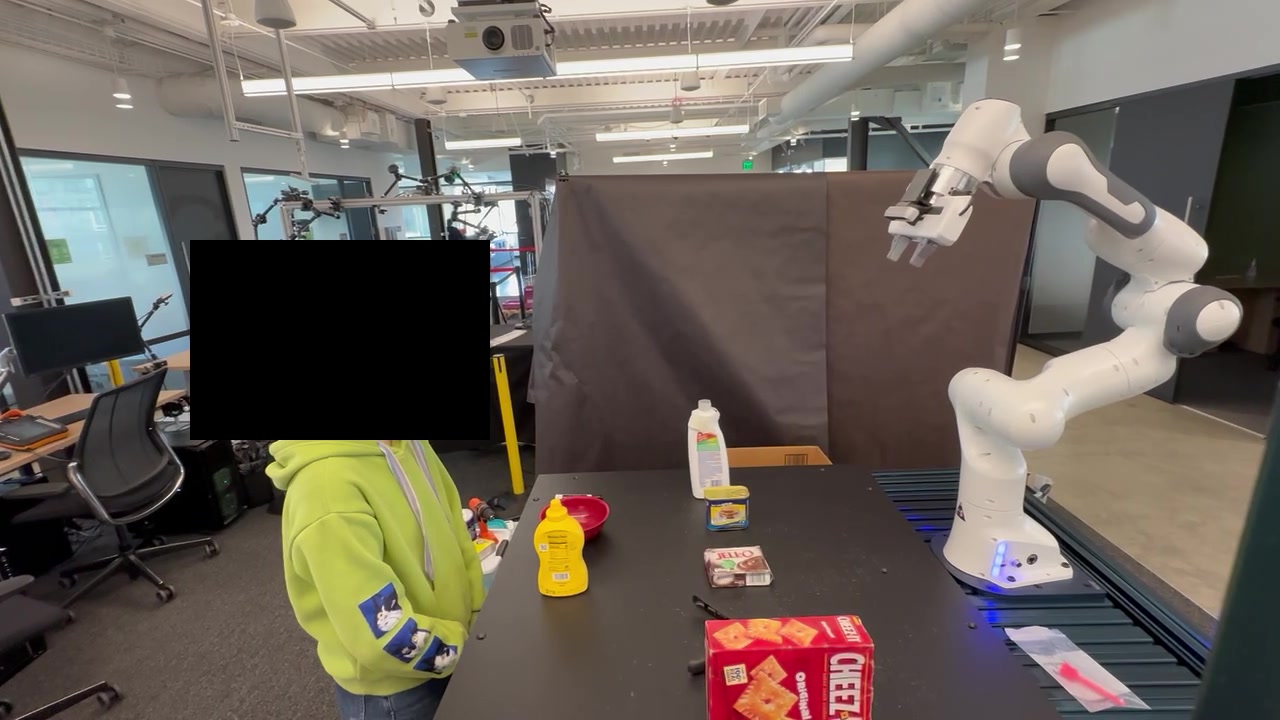}~
 \includegraphics[width=0.191\linewidth]{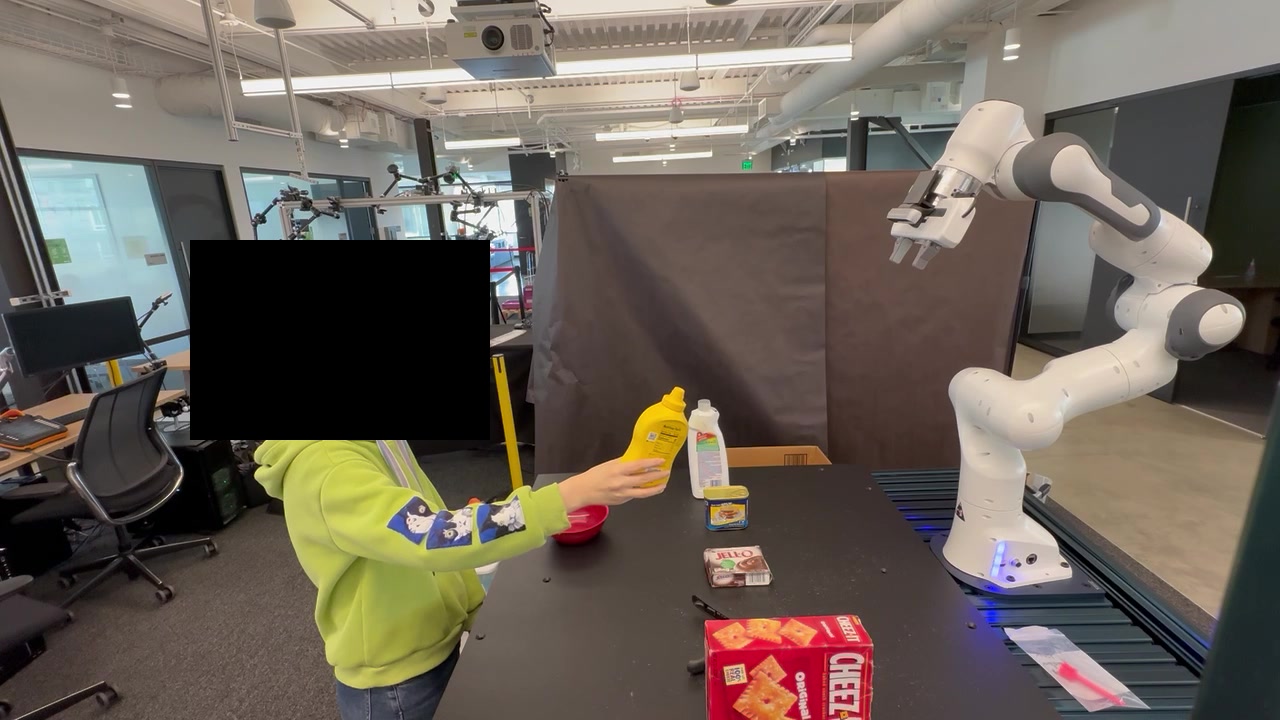}~
 \includegraphics[width=0.191\linewidth]{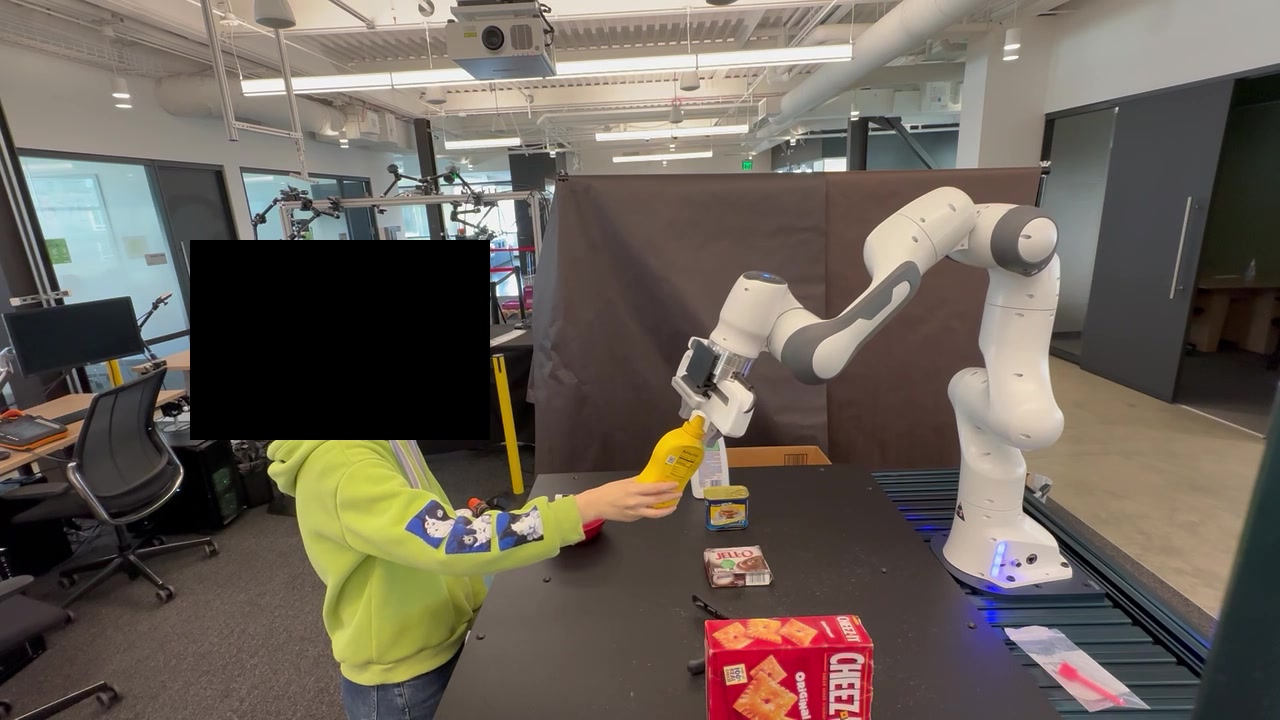}~
 \includegraphics[width=0.191\linewidth]{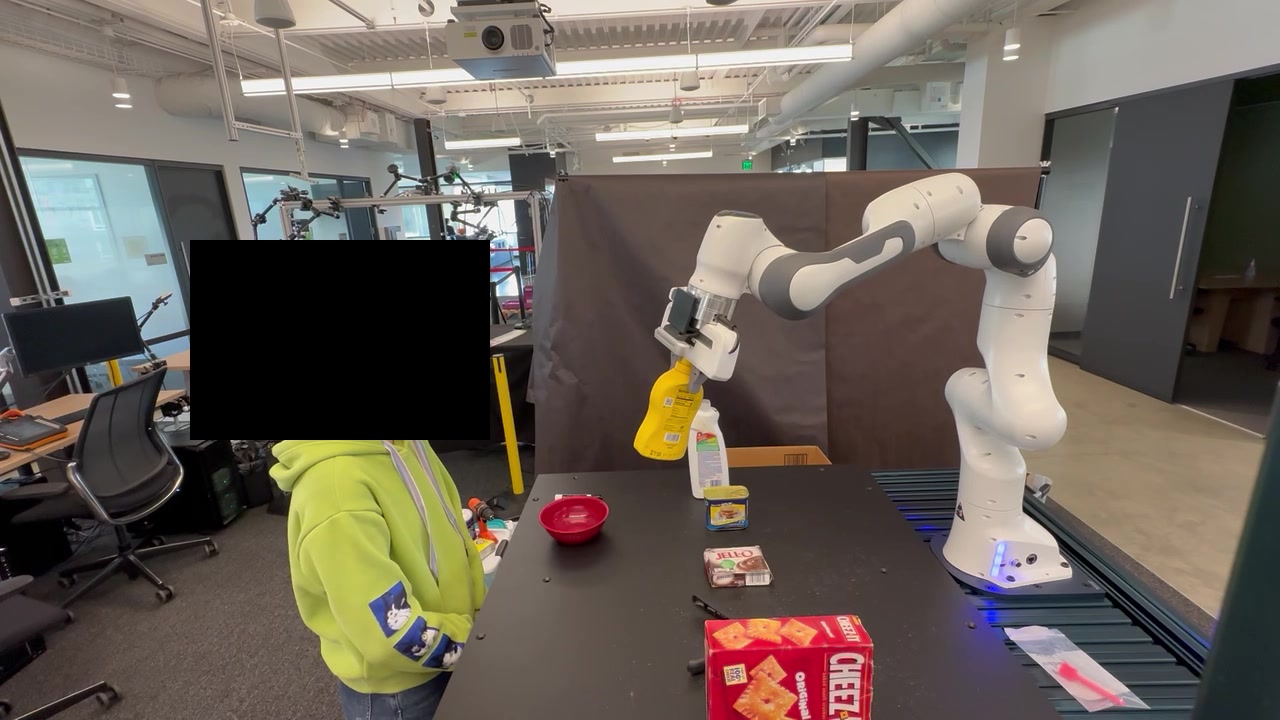}~
 \includegraphics[width=0.191\linewidth]{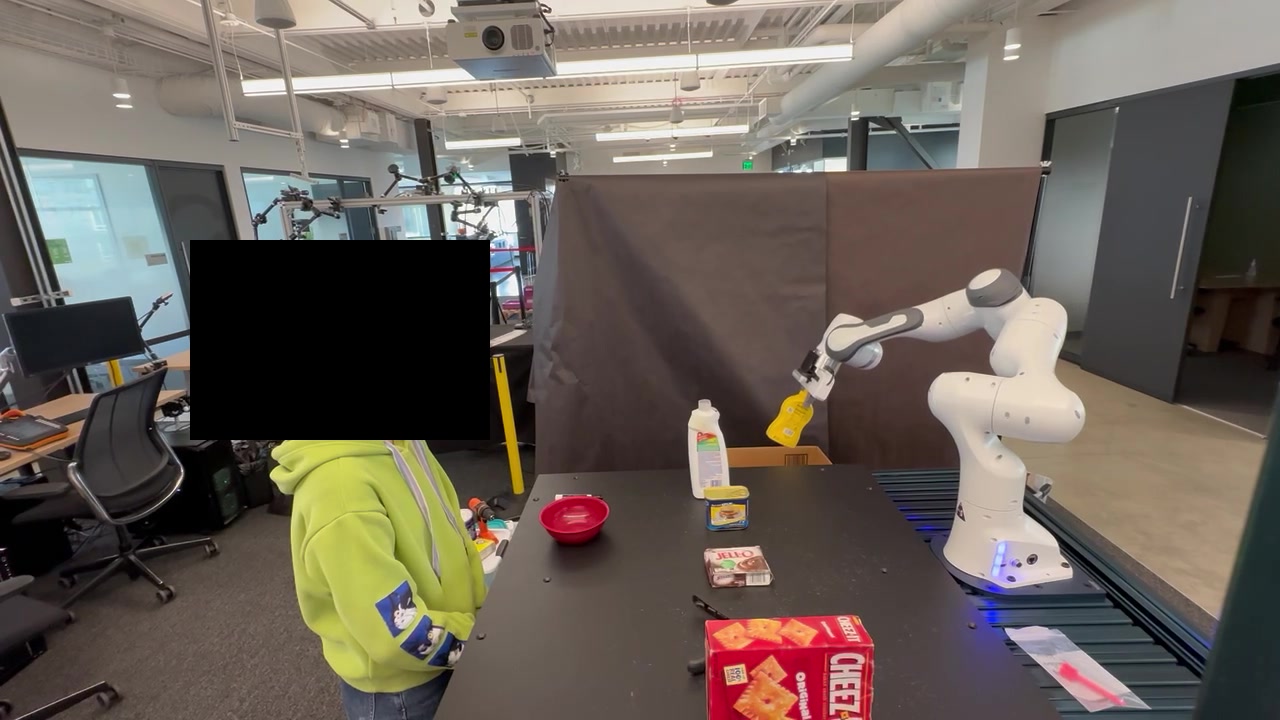}
 \\ \vspace{1mm}
 \includegraphics[width=0.191\linewidth]{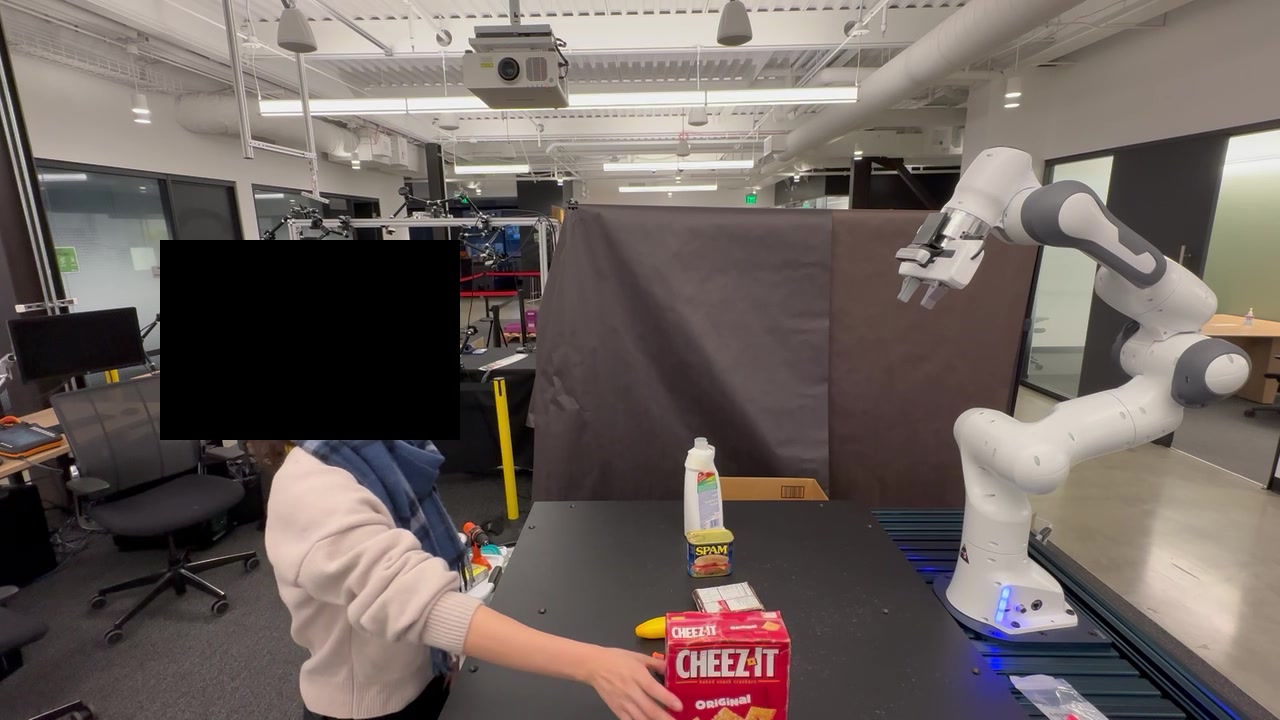}~
 \includegraphics[width=0.191\linewidth]{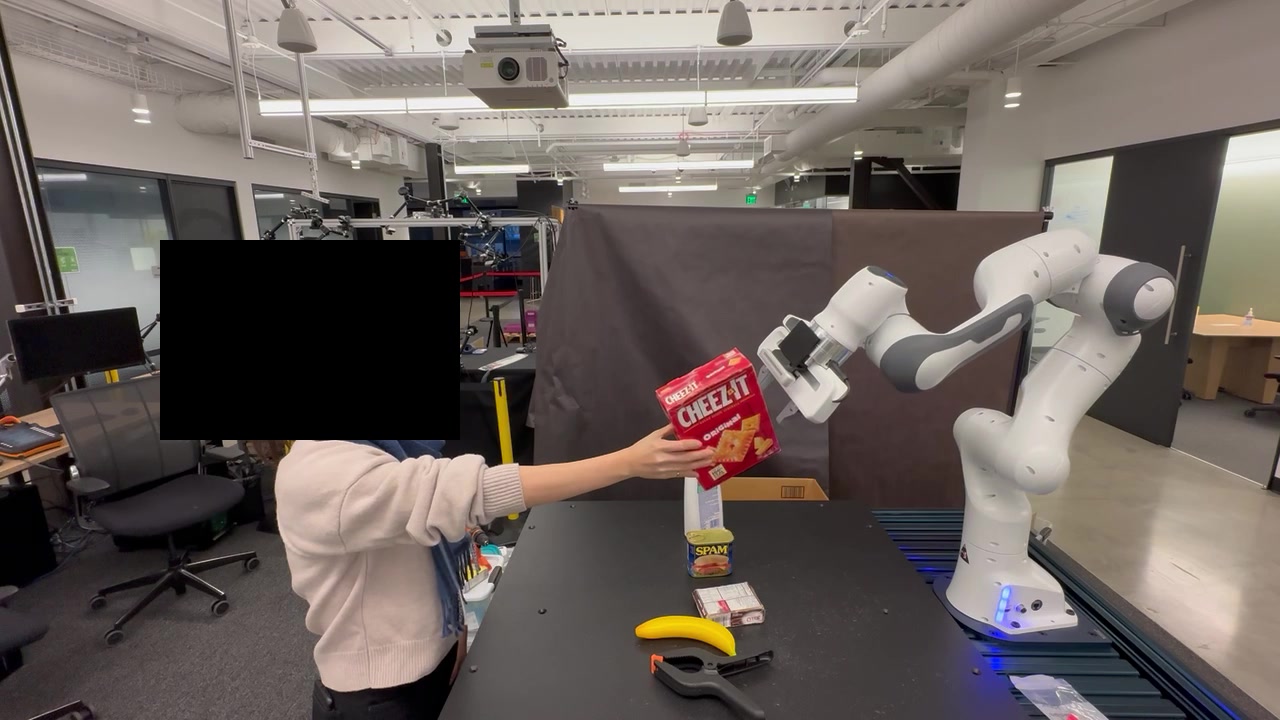}~
 \includegraphics[width=0.191\linewidth]{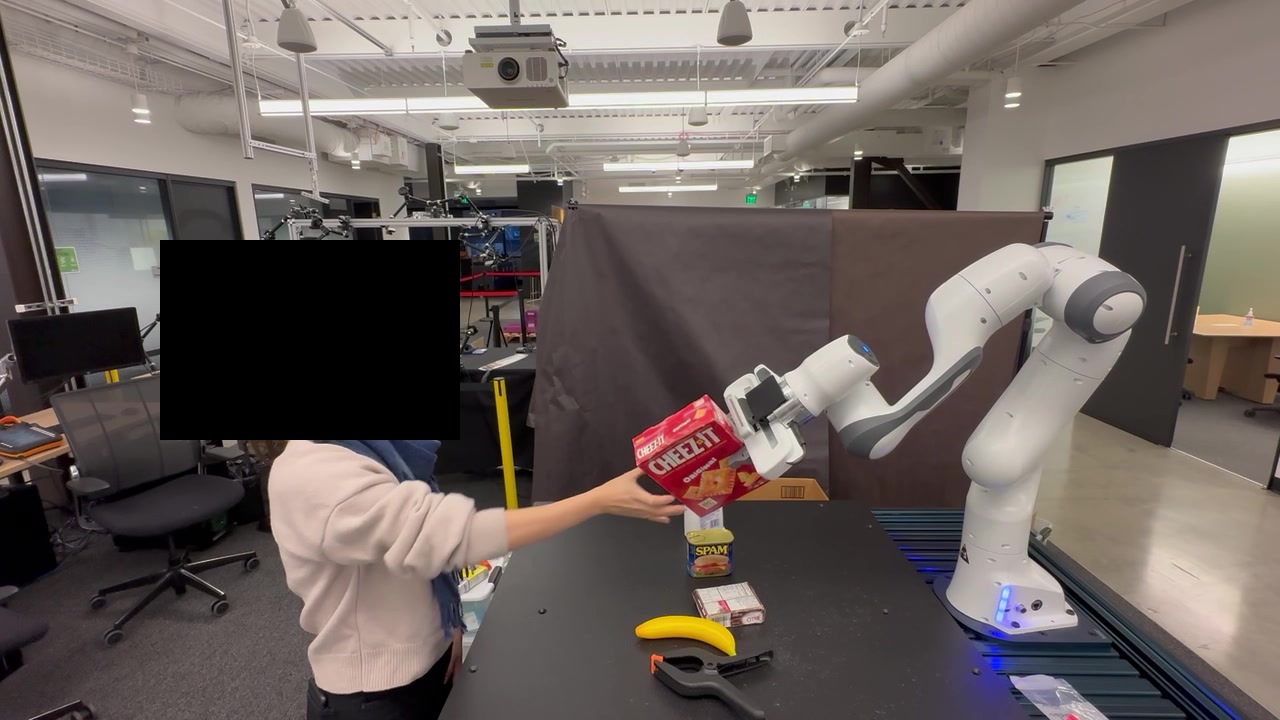}~
 \includegraphics[width=0.191\linewidth]{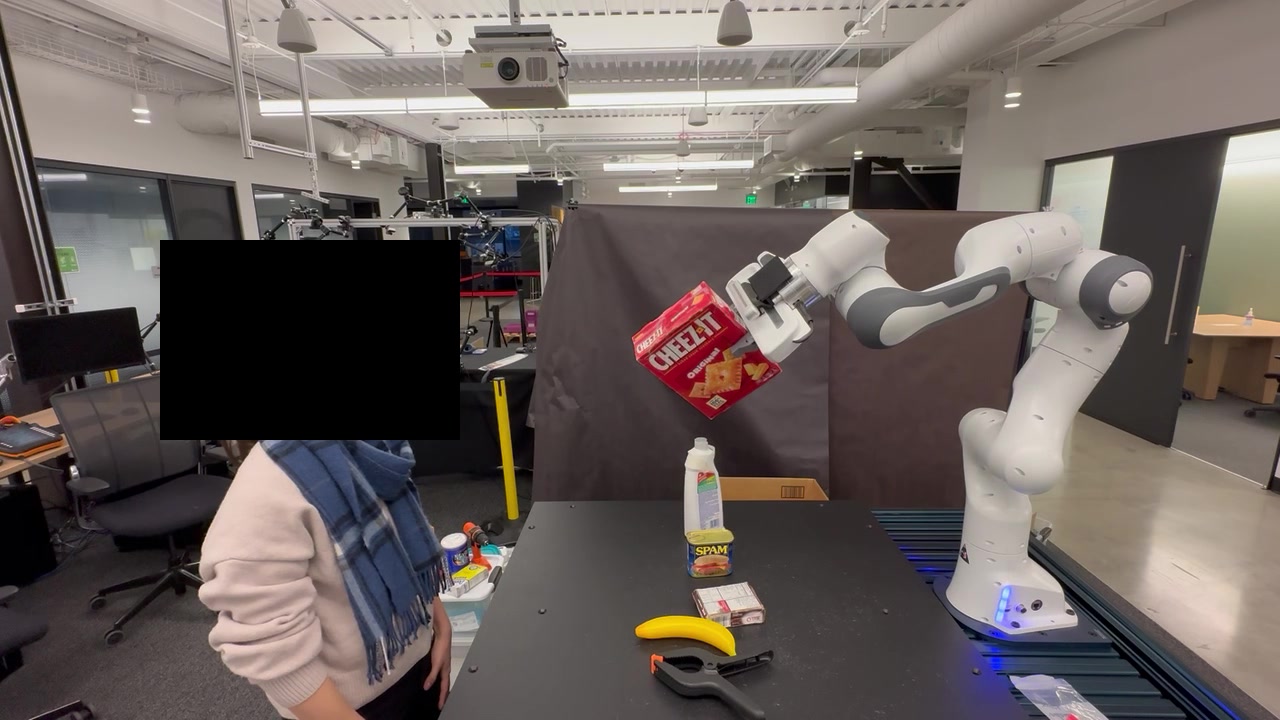}~
 \includegraphics[width=0.191\linewidth]{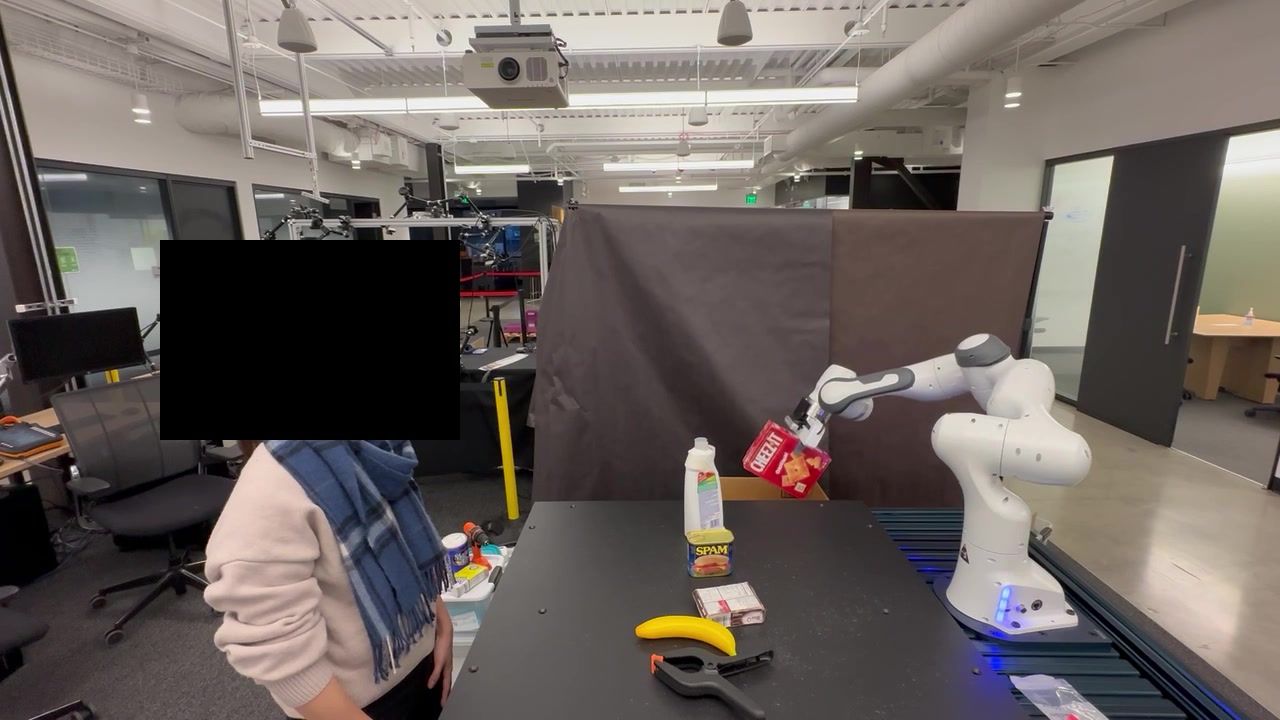}
 \\ \vspace{1mm}
 \includegraphics[width=0.191\linewidth]{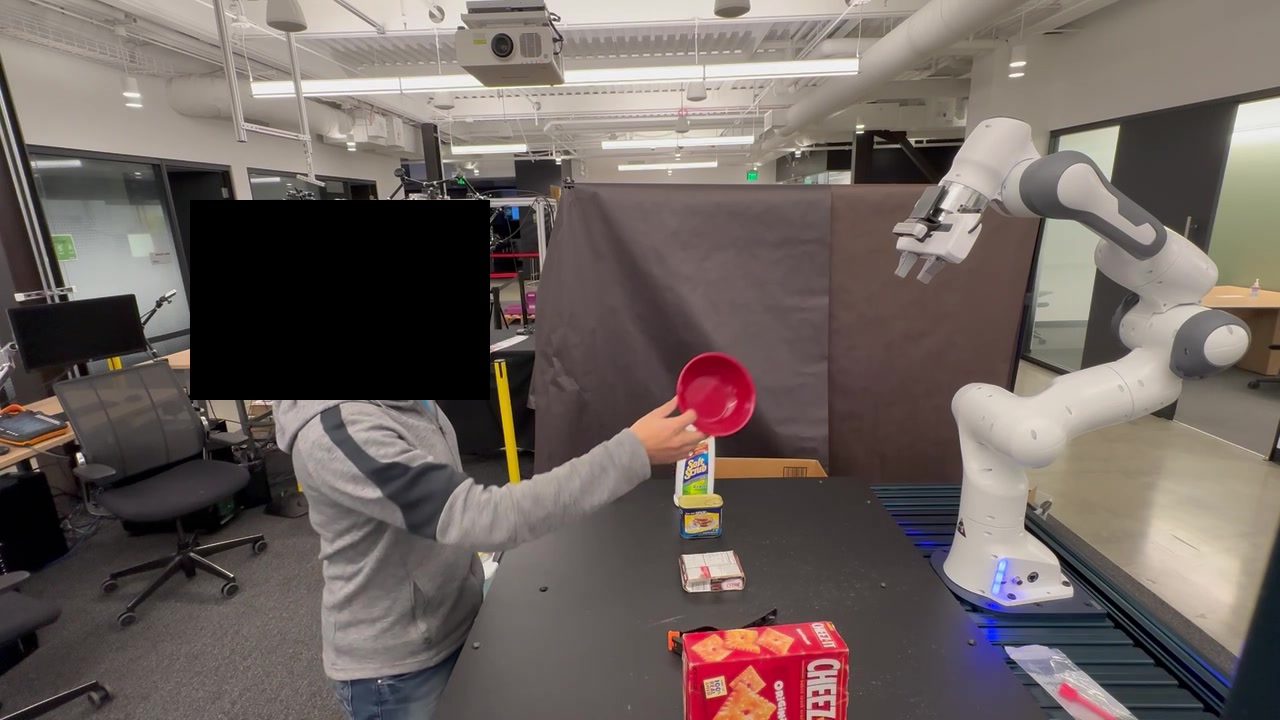}~
 \includegraphics[width=0.191\linewidth]{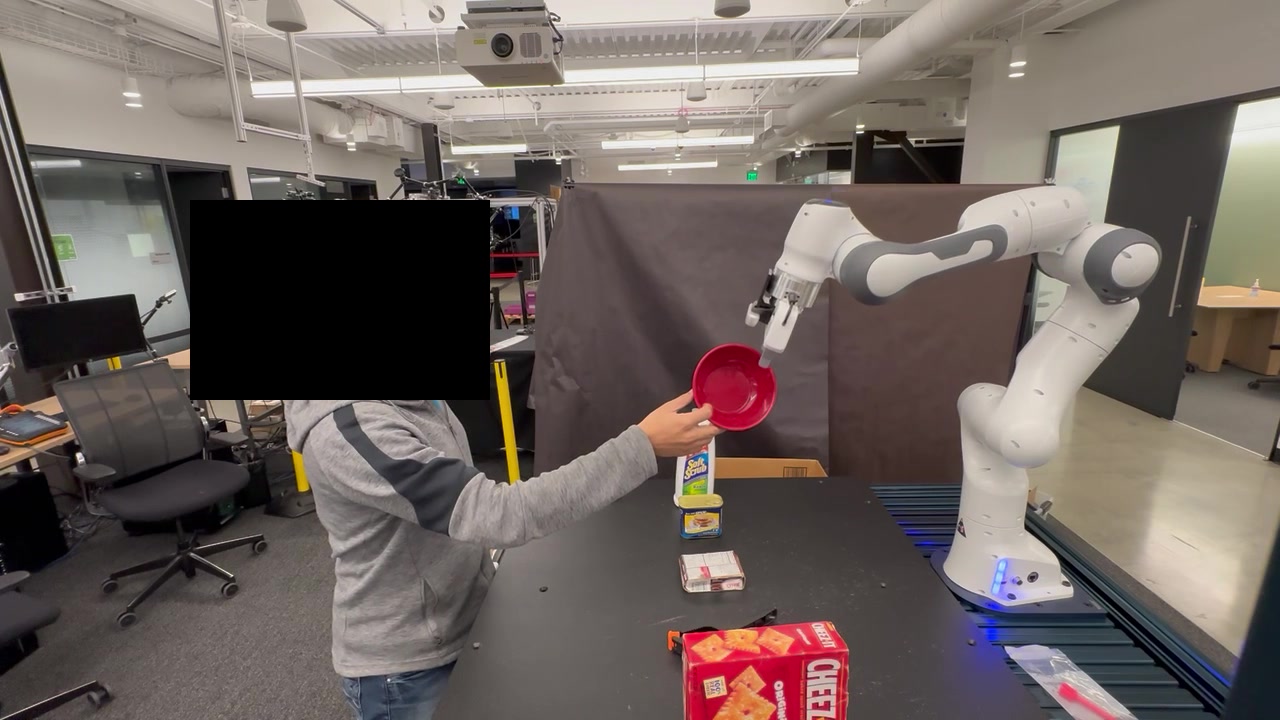}~
 \includegraphics[width=0.191\linewidth]{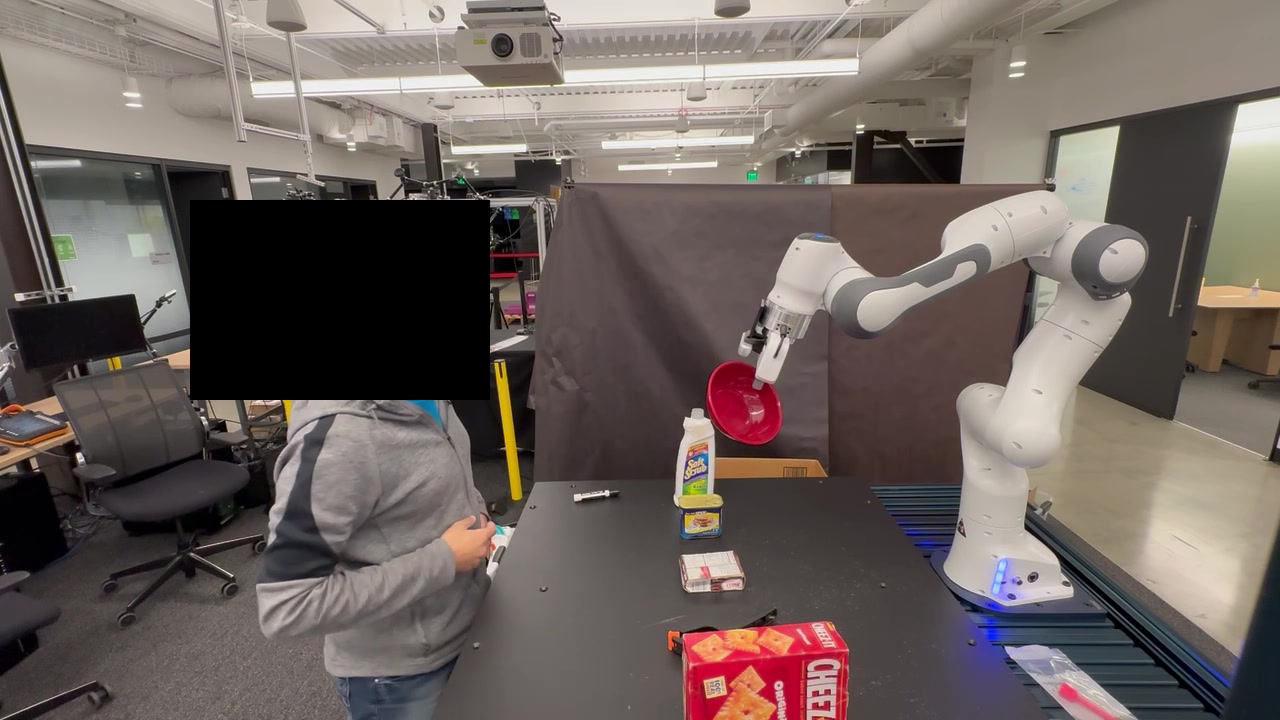}~
 \includegraphics[width=0.191\linewidth]{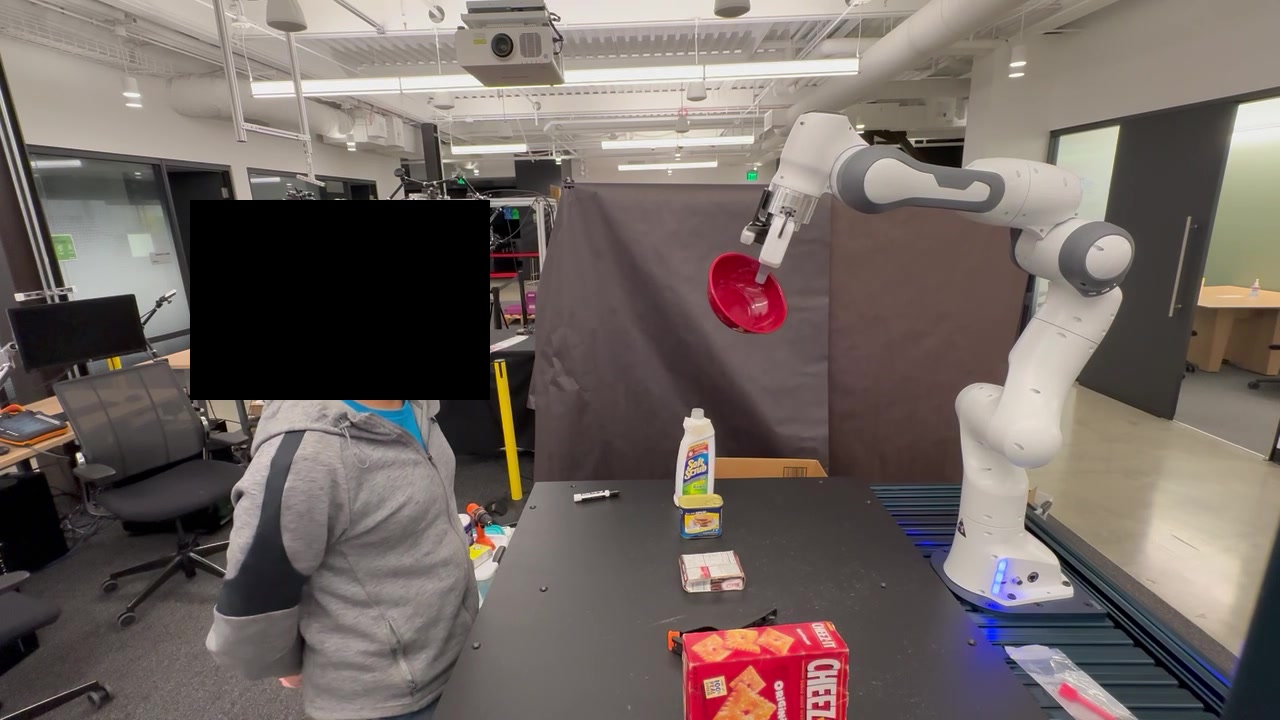}~
 \includegraphics[width=0.191\linewidth]{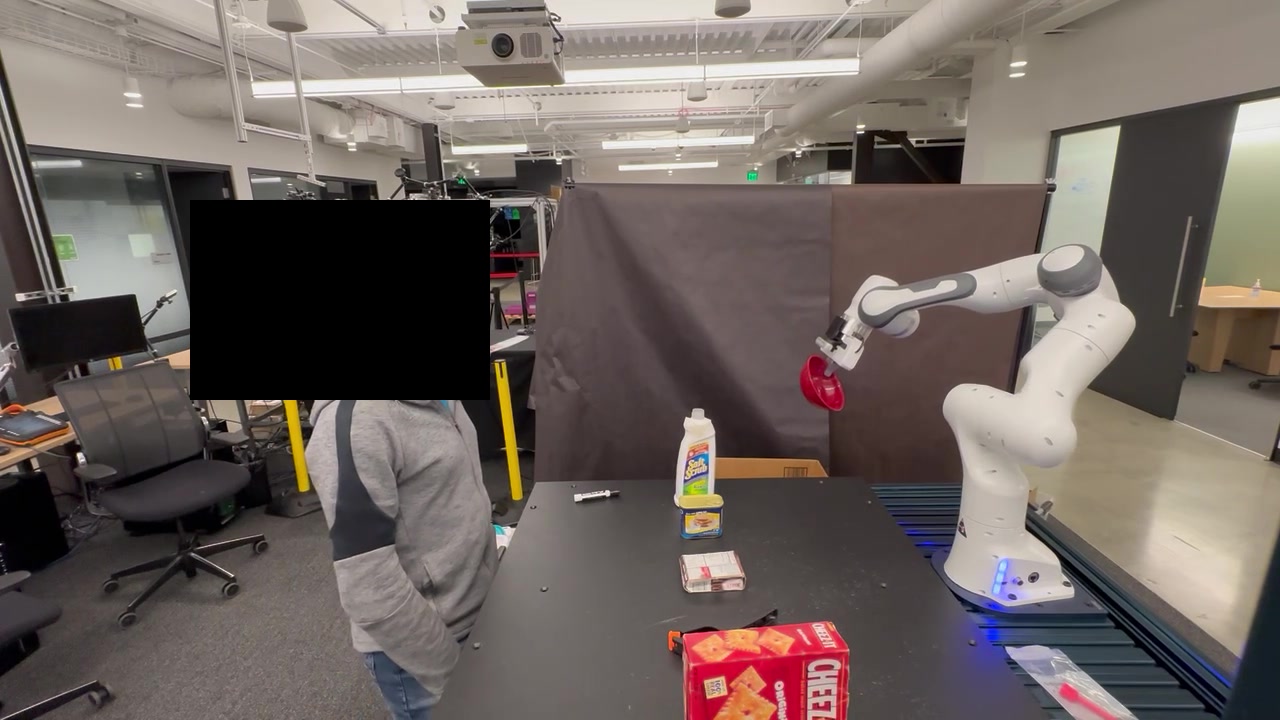}
 \\ \vspace{1mm}
 \includegraphics[width=0.191\linewidth]{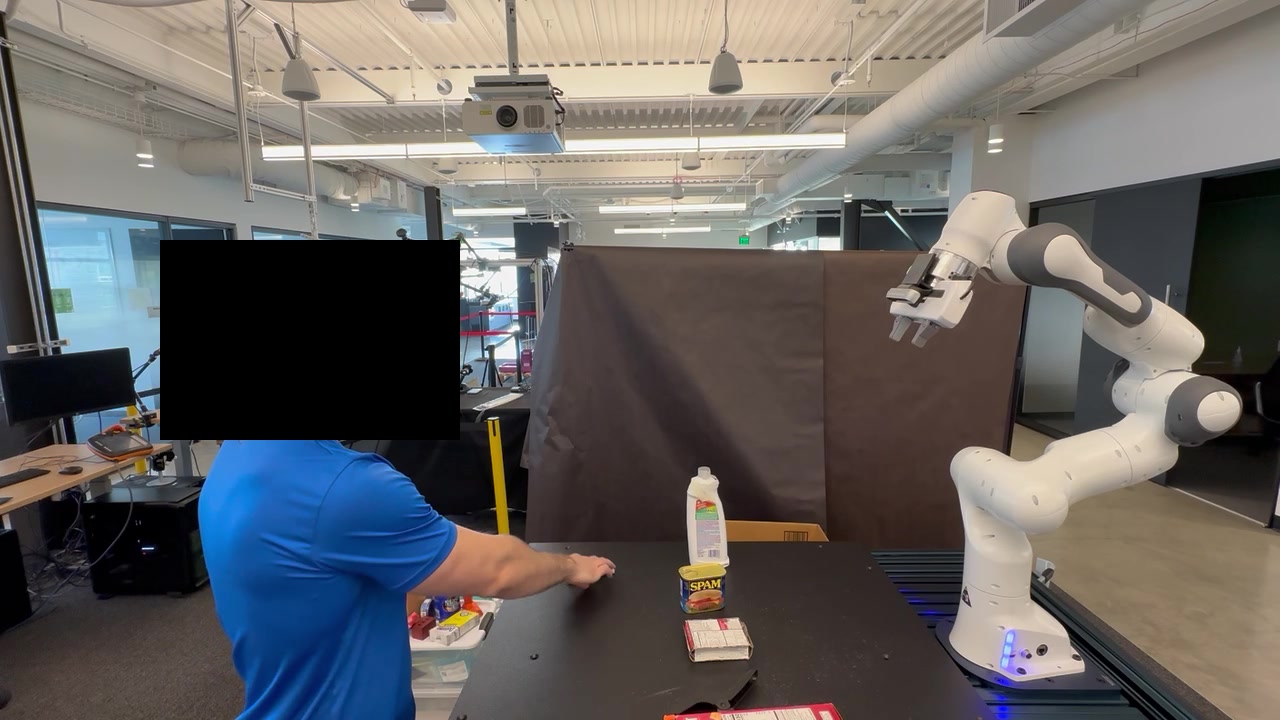}~
 \includegraphics[width=0.191\linewidth]{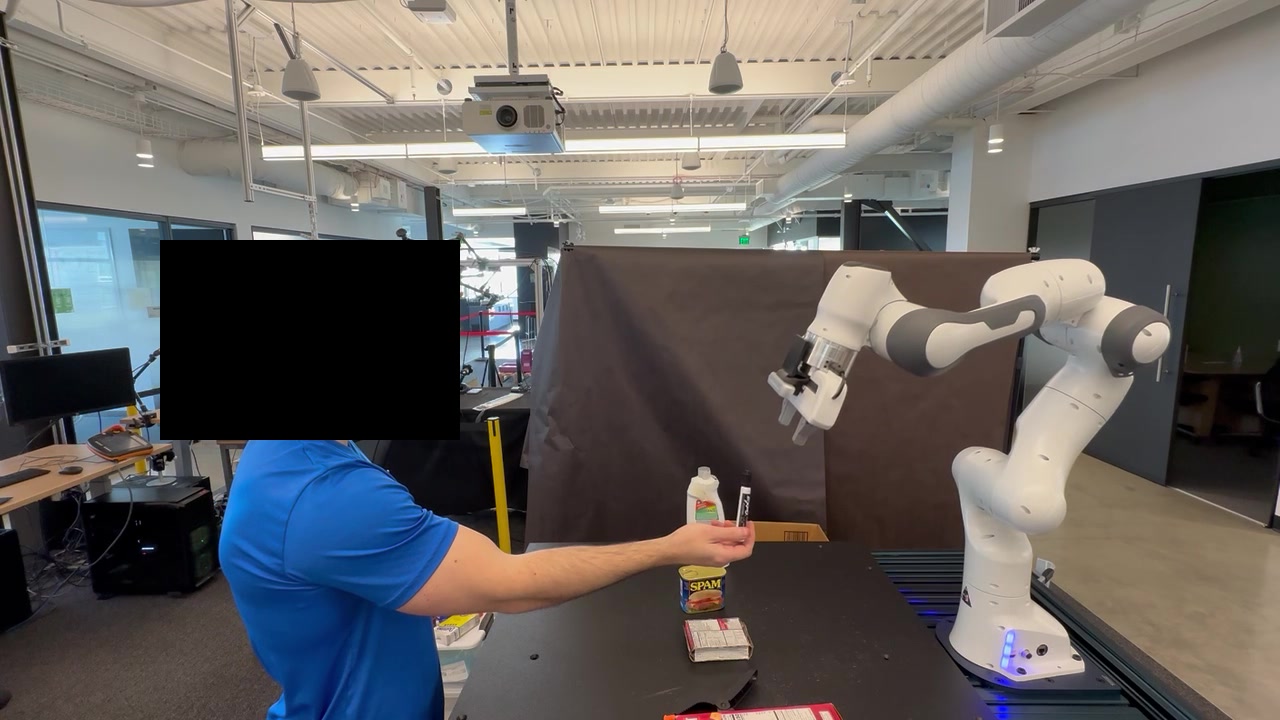}~
 \includegraphics[width=0.191\linewidth]{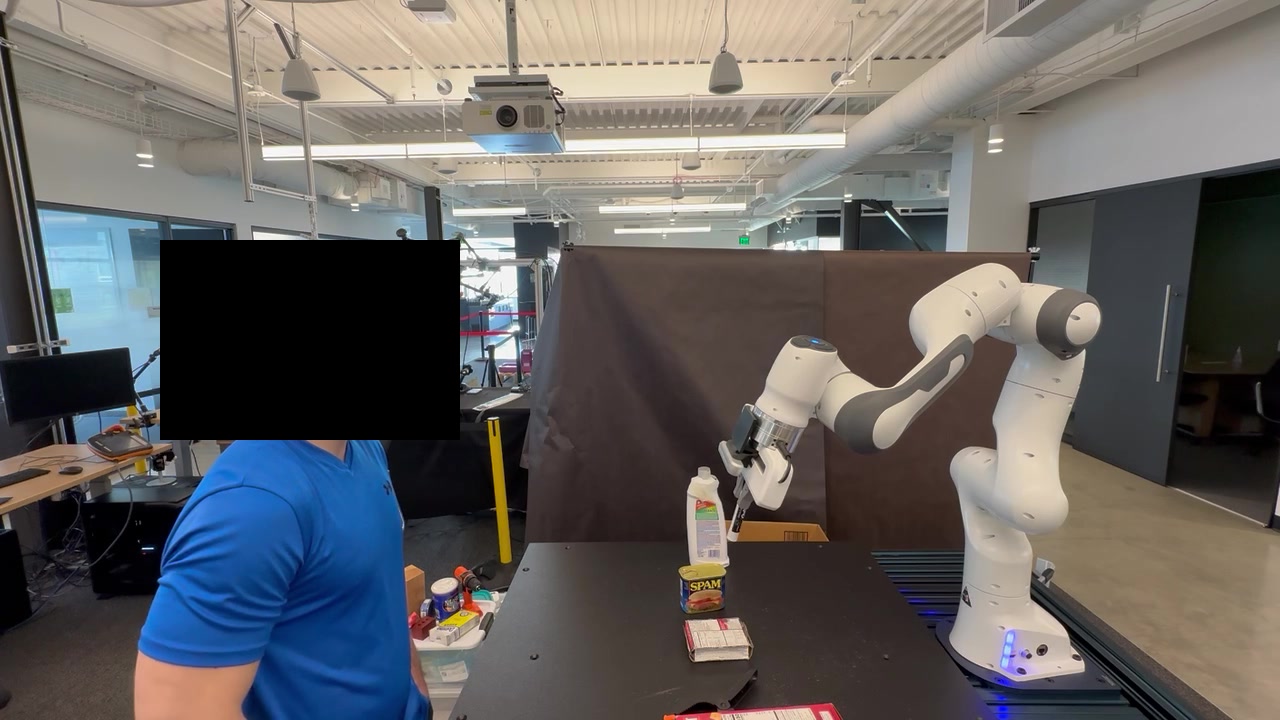}~
 \includegraphics[width=0.191\linewidth]{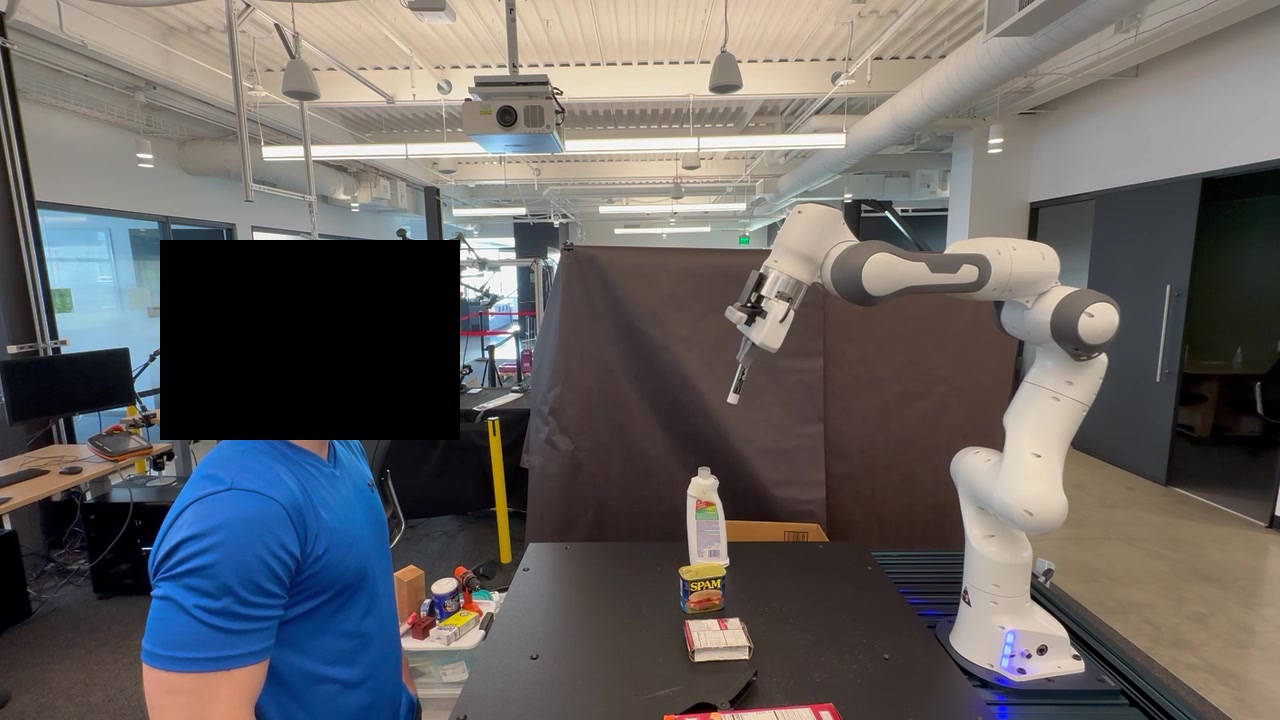}~
 \includegraphics[width=0.191\linewidth]{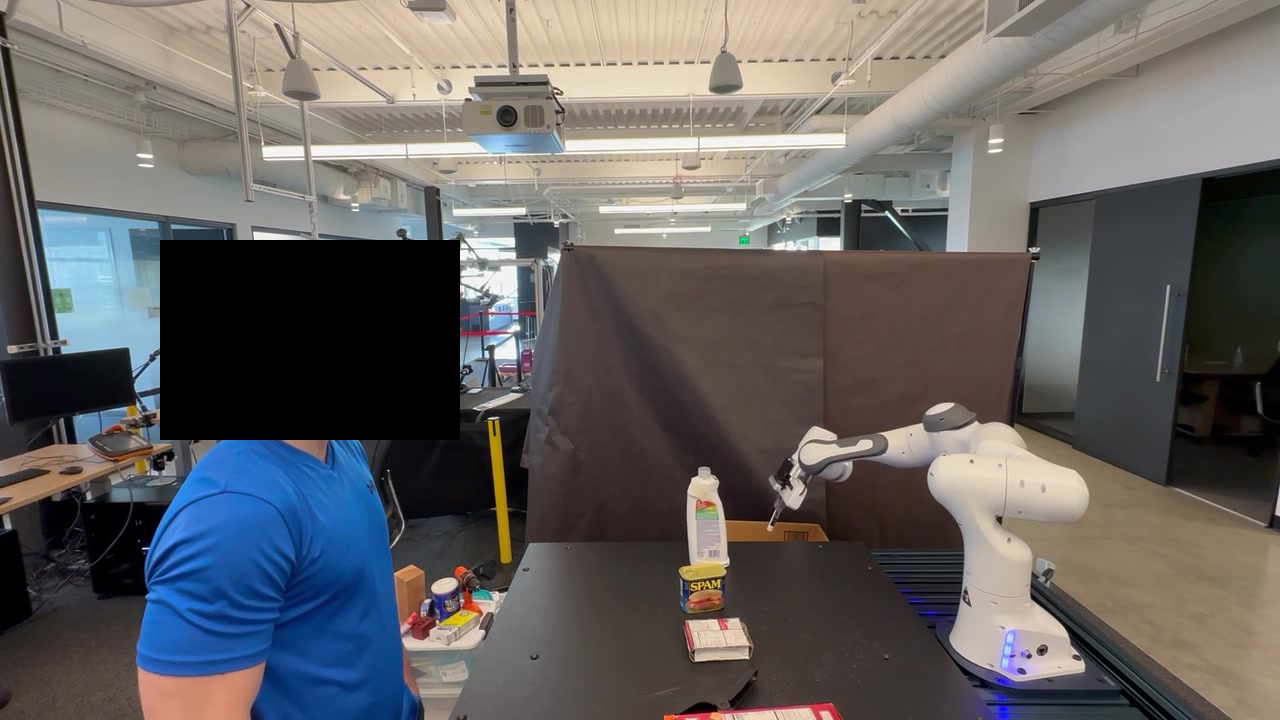}
 \caption{\small We conduct a user evaluation with 6 users by allowing the users to perform handovers freely. The images depict sequences (from left to right) of different users handing over a variety of objects to the robot.}
  \vspace{-2mm}
 \label{fig:user}
\end{figure*}

In contrast to the standardized evaluation in the pilot study, the goal of our user evaluation is to collect feedback from lay users with their own handover preferences. Therefore, we do not constrain the users on how they hand over objects. Instead, we let them carry out in ways which they feel most comfortable (\cref{fig:user}). Rather than objective metrics, we collect subjective feedback from the users via a questionnaire.

\vspace{-3mm}
\paragraph{Evaluation Protocol}~We adopt the same 10 objects from the pilot study, and ask each user to hand over each object once with their right hand. We instruct the users to hand over objects ``in any way they like''. We compare the two methods (i.e., GA-DDPG~\cite{wang:corl2021b} and ours) by repeating the same process, i.e., we instruct the user to hand over the 10 objects to one system first, followed by to the other system. We counterbalance the order of the two systems throughout the user evaluation to avoid bias. During their experiments, the users are asked to fill out a questionnaire with Likert-scale and open-ended questions to provide feedback after they interact with each system.

\begin{figure*}[t!]
 \centering
 \includegraphics[width=\linewidth]{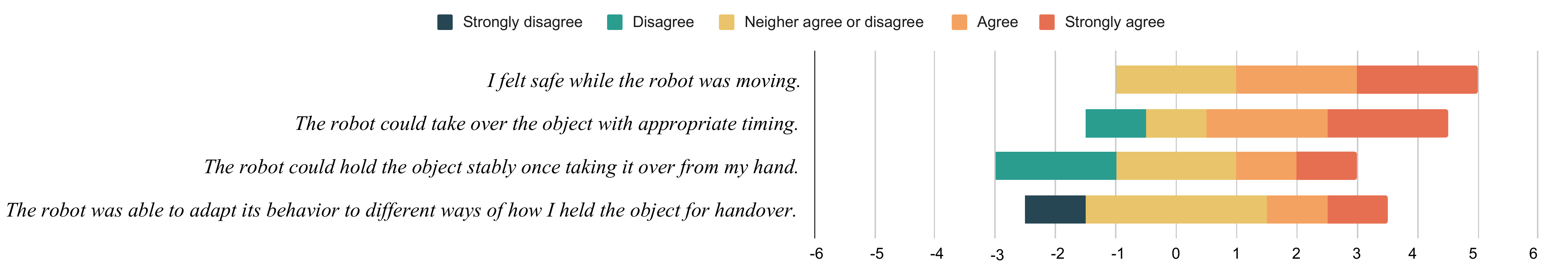}
 \includegraphics[width=\linewidth]{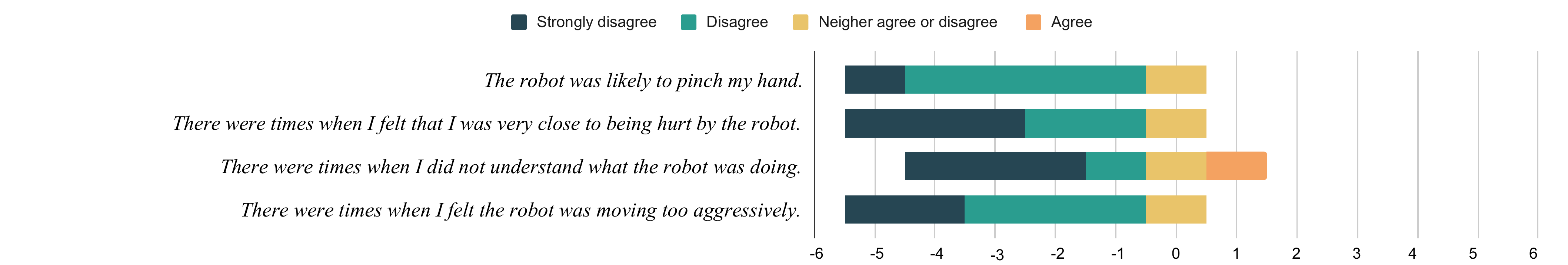}
 \caption{\small User's ranking with each statement for \textbf{GA-DDPG~\cite{wang:corl2021b}} in the user evaluation. Each color denotes a different a degree of agreement. The length of the bar denotes the count of the users. For each bar, the center count of ``Neither agree or disagree'' is aligned with 0 in the horizontal axis. In the top figure, a higher agreement score (orange) indicates a better performance, while in the bottom figure, a lower agreement score (green) indicates a better performance.}
 \label{fig:user_gaddpg}
\end{figure*}

\begin{figure*}[t!]
 \centering
 \includegraphics[width=0.9\linewidth]{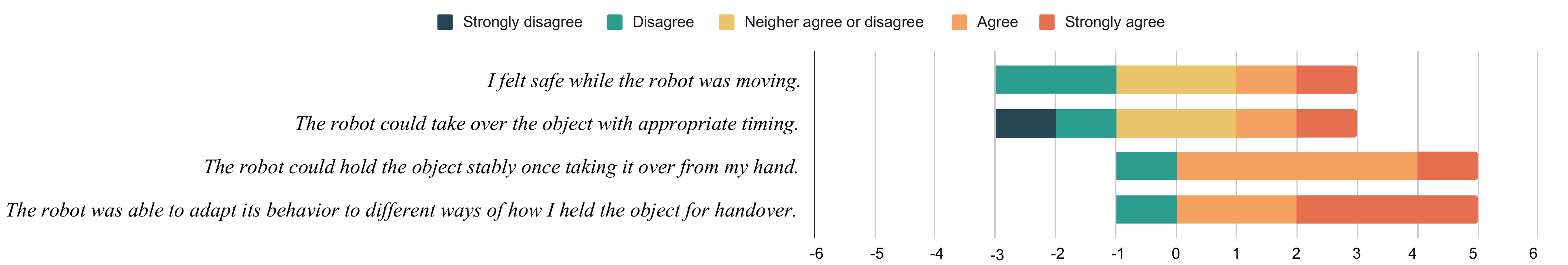}
 \includegraphics[width=0.9\linewidth]{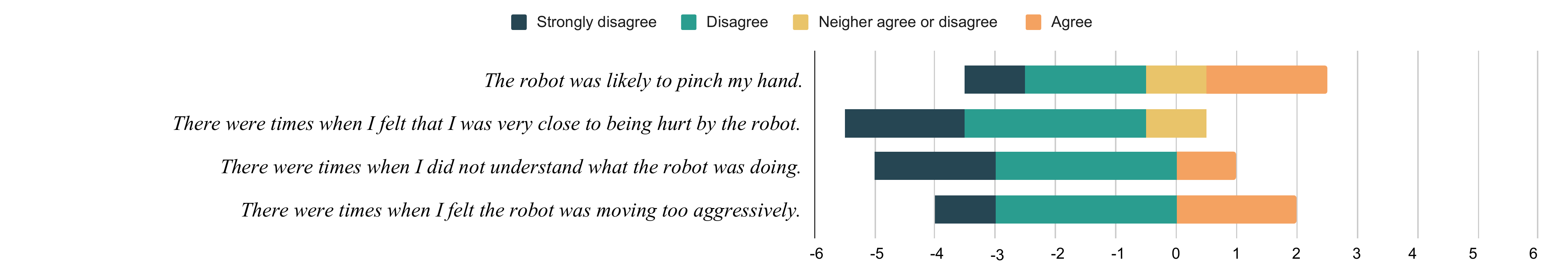}
 \caption{\small User's ranking with each statement for \textbf{our method} in the user evaluation. Each color denotes a different a degree of agreement. The length of the bar denotes the count of the users. For each bar, the center count of ``Neither agree or disagree'' is aligned with 0 in the horizontal axis. In the top figure, a higher agreement score (orange) indicates a better performance, while in the bottom figure, a lower agreement score (green) indicates a better performance.}
 \label{fig:user_ours}
\end{figure*}

\vspace{-3mm}
\paragraph{Results}~We conduct our user evaluation with 6 users (\cref{fig:user}). The evaluation results are presented in \cref{fig:user_gaddpg} for GA-DDPG~\cite{wang:corl2021b} and \cref{fig:user_ours} for our method. Each figure shows the user's ranking with the statements queried in the questionnaire. For each statement, a user can rank their agreement level with one of the five options: ``Strongly disagree'' (1), ``Disagree'' (2), ``Neither agree or disagree'' (3), ``Agree'' (4), and ``Strongly agree'' (5) (see the color codes in \cref{fig:user_gaddpg,fig:user_ours}). The length of each color bar denotes the count of the users. For each method, the statements are further grouped into two subfigures, where a higher agreement score indicates a better performance (top), and a lower agreement score indicates a better performance (bottom).

Overall, our method receives higher agreement scores over GA-DDPG~\cite{wang:corl2021b} for the statements ``\textit{The robot could hold the object stably once taking it over from my hand.}'' (i.e., (5,4,4,4,4,3) versus (5,4,3,3,3,2)) and ``\textit{The robot was able to adapt its behavior to different ways of how I held the object for handover.}'' (i.e., (5,5,5,4,4,3) versus (5,4,3,3,3,2)). This is congruent with our simulation evaluation results that our method can grasp objects more robustly by finding good pre-grasp poses around the object. \update{This was also reflected in participants' comments. One said our method ``\textit{tends to explore more diverse grasp}", ``\textit{was much better at aligning the grasp}" and ``\textit{adjusts behavior for different objects in different poses}" when compared with GA-DDPG.  One pointed out that sometimes GA-DDPG ``\textit{grasped from the tip of the object}". 
The interpretability of the robot's motion was also acknowledged by their comments, \eg, it ``\textit{[was] safe and interpretable at all times}" and ``\textit{felt like we understood each other}".}
Surprisingly, the users favor GA-DDPG~\cite{wang:corl2021b} more when it comes to safety related metrics, e.g., for the statement ``\textit{I felt safe while the robot was moving.}'' ((5,4,3,3,2,2) for ours versus (5,5,4,4,3,3) for GA-DDPG~\cite{wang:corl2021b}) and ``\textit{The robot was likely to pinch my hand.}'' ((1,2,2,3,4,4) for ours versus (1,2,2,2,2,3) for GA-DDPG~\cite{wang:corl2021b}). This can be attributed to GA-DDPG's tendency to grasp from the grasp points closest to the robot, and hence it often keeps a safe distance from the human hand. For our method, several users felt the robot hand pushing too much during grasping. One said it was ``\textit{flexible in grasp selection, but may be too close to my finger}''. Another said ``\textit{the forward movement ... put the gripper fairly close to me}''. This can potentially be addressed by incorporating force feedback in the grasping motion as well as taking gripper hand distance into account during training. The majority of participants agreed that the timing of our method is more appropriate, commenting the ``\textit{handover time was pretty seamless}" and ``\textit{didn't have to wait too long}".

\update{Although the main objective in the user study was to let users interact freely with the system in a non-standardized manner, we additionally evaluate the user study quantitatively. We report the success rate and approach time (\ie, from the robot starting to move to grasp completion). Our method still compares favorably to GA-DDPG with a higher success rate ($88.9\%$ \vs $80.0\%$) and a shorter approach time ($6.40\pm2.27$s \vs $7.48\pm2.64$s). The better timing was noted by the majority of participants, who commented that the ``\textit{handover time was pretty seamless}" and ``\textit{didn't have to wait too long}". Interestingly, we observed in our user study that natural H2R handovers are less susceptible to grasping failures, since the human partner would often help by agilely adjusting the object pose in the last mile to ensure a successful grasp.}




\begin{figure*}[t!]
 \centering
 \includegraphics[width=0.21\linewidth]{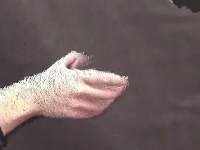}~
 \includegraphics[width=0.21\linewidth]{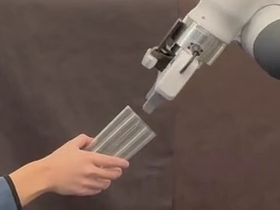}
 \includegraphics[width=0.21\linewidth]{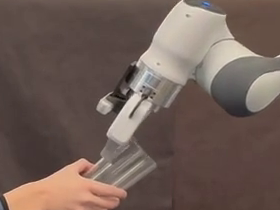}
 \includegraphics[width=0.21\linewidth]{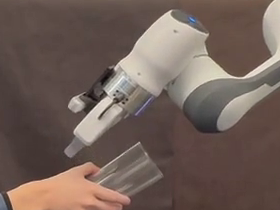}
 \\ \vspace{1mm}
 \includegraphics[width=0.21\linewidth]{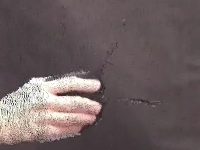}~
 \includegraphics[width=0.21\linewidth]{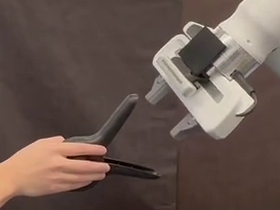}
 \includegraphics[width=0.21\linewidth]{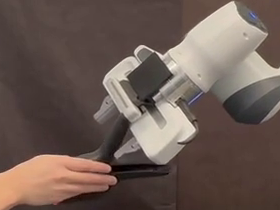}
 \includegraphics[width=0.21\linewidth]{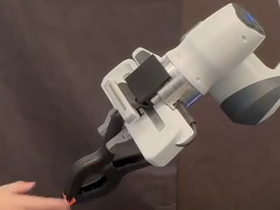}
 \\

 \caption{\textbf{Real World Failures}. \textnormal{Left:} Missing/sparse point cloud of transparent/dark objects in real world perception. \textnormal{Right:} Handover policy behavior.}

 \label{fig:real_failure}
\end{figure*}


\section{Limitations and Future Work}
\label{app_lim}

We will now discuss failure cases of our method and exciting directions for future work. Some failures occur with smaller objects, where the human hand often encloses large parts of the object. For the robot to find grasps where the gripper does not touch the hand at all in such cases is extremely difficult, especially when only having access to point cloud input. The grasp prediction task that decides when to switch from approaching to grasping is quite challenging, because a small change in end-effector pose can already cause a handover to fail. We sometimes find that the grasp prediction triggers the grasp too early or in an instance where the object will eventually drop. Since the grasp prediction network is trained offline, it may be improved by finetuning in online fashion with experiences from policy rollouts. Furthermore, we investigate \textit{sensor-challenging} objects in real world transfers. Our depth sensor is vulnerable to transparent or dark objects, which may lead to failures of the policy (\Fig{real_failure}). Improving the vision pipeline to detect such objects reliably \cite{yen:2022icra} could be a viable direction.

We find that most human trajectories in HandoverSim have roughly the same length. A future direction can therefore include exploring a wider variety of human behaviors. \update{For example, in a realistic, interactive setting the human may be constantly moving, and the robot should only take objects from the human once it wants to hand them over. Anticipating the intent and future states of the human could provide a more natural system. We also noticed that humans start adapting to the robot once they learn how it behaves. Therefore, introducing a multi-agent training scheme where both the simulated human and the robot are trained jointly \cite{langerak2022marlui} is interesting. Another direction could include making the RL exploration more efficient, as it currently still requires long training times, e.g., by leveraging state-agnostic priors \cite{bagatella2022sfp} or using an intrinsic reward to incentivize contacts \cite{vulin2021}. Lastly, our framework can potentially serve as a real world application framework to evaluate vision pipelines, e.g., for testing hand and object pose estimation estimation pipelines \cite{ziani2022tempclr,tse2022collaborative} or segmentation models \cite{fan2021digit}.}




\end{document}